\documentclass[11pt]{article}
\usepackage[utf8]{inputenc}
\usepackage[margin=1in]{geometry}

\usepackage{amsmath,amsfonts,amsthm,amssymb,bm, verbatim,dsfont,mathtools}
\usepackage{booktabs,multirow} % for professional tables
\usepackage{hyperref}
\usepackage[capitalize,noabbrev]{cleveref}
\theoremstyle{plain}
\newtheorem{theorem}{Theorem}[section]
\newtheorem{proposition}[theorem]{Proposition}
\newtheorem{lemma}[theorem]{Lemma}
\newtheorem{corollary}[theorem]{Corollary}
\theoremstyle{definition}

\newtheorem{assumption}[theorem]{Assumption}
\theoremstyle{remark}
\newtheorem{remark}[theorem]{Remark}
\usepackage[textsize=tiny]{todonotes}

\usepackage[toc,page]{appendix}
\usepackage{url}

\usepackage{color,graphicx,appendix}
\usepackage{subcaption}
\usepackage{etoolbox}
\usepackage{tikz}
\usepackage{xr,xspace}
\usepackage{todonotes}
\usepackage{paralist}
\usepackage{pgfplots}
\usepackage{algorithm}% http://ctan.org/pkg/algorithms
\usepackage{algorithmic}% http://ctan.org/pkg/algorithms
\makeatletter
\newcommand{\argmax}{\mathop{\arg\max}}

\usepackage{xspace,prettyref}

\newcommand{\iiddistr}{{\stackrel{\text{\iid}}{\sim}}}

\newcommand{\reals}{{\mathbb{R}}}

\newcommand{\diff}{{\rm d}}
\newcommand{\red}{\color{red}}

\newcommand{\Expect}{\mathbb{E}}
\newcommand{\expect}[1]{\mathbb{E}\left[ #1 \right]}

\newcommand{\Prob}{\mathbb{P}}

\newcommand{\var}{\mathsf{var}}

\newcommand\indep{\protect\mathpalette{\protect\independenT}{\perp}}
\def\independenT#1#2{\mathrel{\rlap{$#1#2$}\mkern2mu{#1#2}}}
\newcommand{\Bern}{{\rm Bern}}
\newcommand{\Binom}{{\rm Binom}}

\newcommand{\iid}{i.i.d.\xspace}
\newrefformat{eq}{(\ref{#1})}
\newrefformat{chap}{Chapter~\ref{#1}}
\newrefformat{sec}{Section~\ref{#1}}
\newrefformat{alg}{Algorithm~\ref{#1}}
\newrefformat{fig}{Fig.~\ref{#1}}
\newrefformat{tab}{Table~\ref{#1}}
\newrefformat{rmk}{Remark~\ref{#1}}
\newrefformat{clm}{Claim~\ref{#1}}
\newrefformat{def}{Definition~\ref{#1}}
\newrefformat{cor}{Corollary~\ref{#1}}
\newrefformat{lmm}{Lemma~\ref{#1}}
\newrefformat{prop}{Proposition~\ref{#1}}
\newrefformat{app}{Appendix~\ref{#1}}
\newrefformat{hyp}{Hypothesis~\ref{#1}}
\newrefformat{thm}{Theorem~\ref{#1}}
\newrefformat{ass}{Assumption~\ref{#1}}

\newcommand{\pth}[1]{\left( #1 \right)}
\newcommand{\qth}[1]{\left[ #1 \right]}
\newcommand{\sth}[1]{\left\{ #1 \right\}}

\newcommand{\abth}[1]{\left | #1 \right |}

\newcommand{\norm}[1]{\left\|{#1} \right\|_2}
\newcommand{\Norm}[1]{\|{#1} \|_2}

\newcommand{\iprod}[2]{\left \langle #1, #2 \right\rangle}

\newcommand{\indc}[1]{{\mathbf{1}_{\left\{{#1}\right\}}}}

\newcommand{\Floor}[1]{\lfloor {#1} \rfloor}

\newcommand{\calA}{{\mathcal{A}}}

\newcommand{\calC}{{\mathcal{C}}}
\newcommand{\calD}{{\mathcal{D}}}
\newcommand{\calE}{{\mathcal{E}}}
\newcommand{\calF}{{\mathcal{F}}}
\newcommand{\calG}{{\mathcal{G}}}

\newcommand{\calN}{{\mathcal{N}}}

\newcommand{\calP}{{\mathcal{P}}}

\newcommand{\calS}{{\mathcal{S}}}

\renewcommand{\hat}{\widehat}
\renewcommand{\tilde}{\widetilde}

\begin{document}
\title{Federated Learning in the Presence of Adversarial Client Unavailability: Minimax Rates\thanks{L.~Su and M.~Xiang are supported by the ARO Grant W911NF-23-2-0014. J.~Xu is supported by the NSF Grants CCF-1856424 and CCF-2144593. P.~Yang is supported by the NSFC Grant 12101353 and Tsinghua University Initiative Scientific
Research Program.}
}
% Achieving Fundamental Limits in the Presence of Adversarial Client Unavailability

\author{
Lili Su \\
Electrical and Computer Engineering \\
Northeastern University  
%{l.su@northeastern.edu}
 \and
 Ming Xiang \\
 Electrical and Computer Engineering \\
 Northeastern University 
\and 
Jiaming Xu\\
The Fuqua School of Business\\
Duke University %\\
%{jiaming.xu868@duke.edu}
\and 
Pengkun Yang \\
Center for Statistical Science \\ 
Tsinghua University %\\
}

\maketitle

\begin{abstract}
Federated learning is a decentralized machine learning framework that enables collaborative model training without revealing raw data. Due to the diverse hardware and software limitations, a client may not always be available for the computation requests from the parameter server.
An emerging line of research is devoted to tackling arbitrary client unavailability. However, existing work still imposes structural assumptions on the unavailability patterns, impeding their applicability in challenging scenarios wherein the unavailability patterns are beyond the control of the parameter server. Moreover, in harsh environments like battlefields, adversaries can selectively and adaptively silence specific clients.
In this paper, we relax the structural 
assumptions and consider adversarial client unavailability.  
To quantify the degrees of client unavailability, we use the notion of {\em $\epsilon$-adversary dropout fraction}.    
We show that simple variants of FedAvg or FedProx, albeit completely agnostic to $\epsilon$, converge to an estimation error on the order of $\epsilon (G^2 + \sigma^2)$ for non-convex global objectives and $\epsilon(G^2 + \sigma^2)/\mu^2$ for $\mu$-strongly convex global objectives, where $G$ is a heterogeneity parameter and $\sigma^2$ is the noise level. 
Conversely, we prove that any algorithm 
has to suffer an estimation error of at least 
$\epsilon (G^2 + \sigma^2)/8$ and $\epsilon(G^2 + \sigma^2)/(8\mu^2)$ 
for non-convex global objectives and $\mu$-strongly convex global objectives. 
Furthermore, the convergence speeds of the FedAvg or FedProx variants
are $O(1/\sqrt{T})$ for non-convex objectives and $O(1/T)$ for strongly-convex objectives, both of which are the best possible for any
first-order method that only has access to 
noisy gradients. Our proofs build upon a   
tight analysis of the selection bias that arises from adversarial client unavailability yet persists in the entire learning process. 
\end{abstract}

\section{Introduction}
\label{sec: motivation} 
Federated learning is a decentralized machine learning framework wherein the parameter server and the clients collaboratively train machine learning models without having the clients 
%clients coordinate with a parameter server 
%in %the model training without 
disclose local data \cite{mcmahan2017communication,kairouz2021advances}. 
% A parameter server coordinates with the client devices to collaborate in the model training, resulting in a reduced computation burden on the cloud.  
% %Since clients
% %directly participate in the model training, FL also benefits from reduced computation burden at the cloud. 
% However, the massively distributed nature of the FL and the limited availability of clients \cite{mcmahan2017communication,kairouz2021advances,Li2020} may incur significant performance degradation.%bring significant challenges to the actual deployment of federated learning.  
% In particular, 
Cross-device federated learning is often massive in scale and the availability of a client may be constrained by the diverse hardware and software limitations, making full client participation %either computationally expensive or 
impractical~\cite{mcmahan2017communication}. 
% Consequently, the parameter server aggregates the updates from a small client subset only. %, either passively or actively.  
% %selecting some clients to participate \cite{mcmahan2017communication,cho2022towards}  or by passively waiting for the few fastest responses \cite{Li2020,philippenko2020bidirectional,kairouz2021advances}. In particular, 
Most existing work either assumes that client unavailability follows benign distributions  \cite{mcmahan2017communication,Li2020,ruan2021towards,philippenko2020bidirectional} or that the parameter server can arbitrarily recruit participants~\cite{cho2022towards}, i.e., a client must adhere to the computation requests from the parameter server.  % will accept the computation request. 
%However, these assumptions rarely hold in practice unless the training environments are under good control. % \cite{mcmahan2017communication}.  
% 
% 
There is a recent surge of interest in studying %federated learning under 
arbitrary client unavailability~\cite{gu2021fast,wang2022,yan2020distributed,yang2022anarchic}.  Despite that they collectively % These papers 
laid a solid foundation in this direction, % and exciting progress has been made, 
this line of work still imposes structural assumptions on unavailability patterns, impeding their applicability in challenging scenarios wherein the unavailability patterns are beyond the control of the parameter server. This is particularly relevant in non-controllable real-world environments wherein the availability of a client depends, in a complicated way, on multiple time-varying factors that %are beyond the control of the parameter server 
such as external interruptions and hardware/software status \cite{mcmahan2017communication}.
For instance, when the clients are mobile devices, a client may fail to respond to a computation request if disconnected from WiFi or if the communication connection is blocked by surrounding buildings or moving obstacles. 
A client may also suddenly abort federated learning due to factors such as low battery. % and changes of prioritized tasks.  
Moreover, in harsh environments like battlefields, adversaries can selectively and adaptively silence specific clients.

In this paper, we relax those structural assumptions and 
%consider a general form of arbitrary client unavailability -- termed as {\em adversarial client unavailability}. 
consider adversarial client unavailability.  
Specifically, in each round, first the parameter server randomly %selects %recruits
samples $K$ clients, and then 
a system adversary, based on all the information up to now, adaptively selects a subset of sampled clients to be non-responsive.  To quantify the degrees of client unavailability, we use the notion of {\em $\epsilon$-adversary dropout fraction}.    
It is worth noting that our adversarial client unavailability model can be viewed as a special case of Byzantine attacks \cite{lynch1996distributed,bonomi2019approximate}. 
Nevertheless, existing Byzantine-resilient results do not apply to our problem. See Section \ref{sec: related work} for details.

We show that simple variants of the standard FedAvg or FedProx algorithms, albeit %completely 
agnostic to the degree of client unavailability $\epsilon$, enjoy strong performance guarantees. 
We further validate our theories
through numerical experiments on synthetic and real-world datasets. 
Specifically, we study a canonical setup in which the goal is to minimize $F(\theta)\triangleq \sum_{i=1}^M w_i F_i(\theta)$, where
$w_i$ is the weight and $F_i$ is the local objective of client $i\in [M]$.
%\footnote{The true gradient is too expensive to compute. }
The local data is non-IID yet satisfies the standard $(B,G)$-heterogeneity condition % in 
\cite{kairouz2021advances} (formally described in~\prettyref{ass:BG}). 
Only noisy stochastic gradients are available. 
Our main theoretical results can be summarized in the following informal theorem. 

\begin{theorem}[Informal]
\label{thm:informal}
% When $F$ is non-convex, let $\calR(\hat{\theta}):=\|\nabla F(\hat{\theta})\|^2$;
% when $F$ is $\mu$-strongly convex, let $\calR(\hat{\theta}):=\mu^2 \| \hat{\theta}-\theta^*\|_2^2$,
% where $\theta^* = \arg\min F(\theta)$. 
Let $\sigma$ be the average noise level of the stochastic gradients. 
For $\sqrt{\epsilon} B \le 0.1,$
\begin{itemize}
    \item when $F$ is non-convex:
\begin{align*}
\frac{\epsilon(G^2+\sigma^2)}{8} \le \inf_{\hat\theta}\sup_{\calA}\sup_{F_1, \cdots, F_M}
\mathbb{E}[\|\nabla F(\hat{\theta})\|^2]  \le 8 \epsilon(G^2+\sigma^2);
\end{align*}
\item when $F$ is $\mu$-strongly convex:
\begin{align*}
\frac{\epsilon \pth{G^2+\sigma^2}}{8\mu^2}  
\le \inf_{\hat\theta}\sup_{\calA}\sup_{F_1, \cdots, F_M}
\mathbb{E}[\| \hat{\theta}-\theta^*\|_2^2] \le  \frac{8 \epsilon (G^2+\sigma^2)}{\mu^2}, 
\end{align*} 
\end{itemize}
where $\sup_{F_1, \cdots, F_M}$ is taken over all local objectives that collectively satisfy the $(B,G)$-heterogeneity condition, $\calA$ is all adversarial client unavailability that is subject to $\epsilon$-adversary dropout fraction, and $\inf_{\hat\theta}$ is taken over all algorithms $\hat{\theta}$ that have access to noisy local gradients only. % computed by the available clients at each communication round.  
\end{theorem}
% 
%\begin{itemize}
Importantly, the lower and upper bounds match each other up to a universal constant factor. 
\begin{itemize} 
\item The upper bounds are proved in \prettyref{sec:convergence} by analyzing variants of FedAvg or FedProx. We also characterize their convergence speeds to be $O(1/\sqrt{T})$ 
%for non-convex functions
and $O(1/T)$, respectively.  
%for strongly convex functions. 
%. For  non-convex functions, we show that $\expect{\|\nabla F( \hat \theta_t)\|^2}$
%$\calR(\hat{\theta}_t)$
%the estimation errors 
%approaches $4 \epsilon\pth{G^2+\sigma^2}$ at a speed of $O(1/\sqrt{t})$  (with general form in Theorem \ref{thm:nonconvex} and explicit speeds in Corollary \ref{cor:convergence-rate}). For $\mu$-strongly convex functions, we show that $\expect{\| \hat{\theta}_t-\theta^*\|_2^2}$ approaches $4\epsilon\pth{G^2+\sigma^2}/\mu^2$ at a speed of  $O(1/t)$ (Theorem \ref{thm:strongly_convex}). %, where $T$ is the number of communication rounds. 
These convergence speeds are on par with the centralized settings~\cite{ghadimi2013stochastic,nemirovski2009robust}, and are the best possible for any first-order method that has only access to noisy gradients~\cite{arjevani2022lower,gu2021fast}.
% 

%\end{itemize}

\item %We remark that 
The threshold $0.1$ of the assumption  $\sqrt{\epsilon} B \le 0.1$ is chosen for the ease of presentation. Our results continue to hold when $\sqrt{\epsilon} B \le c_0$ as long as $c_0<1$ yet at the expense of inflating the estimation error upper bounds by a large constant factor. The assumption $\sqrt{\epsilon} B <1$ is to some extent necessary. This is because as $B$ increases,  %gets larger, 
the $F_i$'s become more dissimilar, resulting in a reduced ability to tolerate adversarial dropouts. 
%Hence fewer adversarial dropouts can be tolerated. 
% See~\prettyref{rmk:eps-B} 
% %in Appendix~\ref{app: proof of lower bound} 
% for a more detailed discussion.

\item The lower bounds are shown in Section \ref{sec: challenges and limits} (Theorem \ref{thm:nonconvex-lb-dropout-G}) for all algorithms (encompassing randomized and non-federated learning algorithms) and all $\epsilon \in (0,1). $ 
\end{itemize}
%Interestingly, among the two heterogeneity parameters $B$ and $G$, only $G$ appears in the lower and upper bounds. 
%Our analysis reveals that %the heterogeneity parameter 
%$B$ instead influences the convergence speed of FedAvg or FedProx. 

\section{Related work}
\label{sec: related work}
\subsection{Partial client participation.}
Most literature on partial client participation considers random client unavailability ~\cite{kairouz2021advances,mcmahan2017communication,Li2020,yuan2022,ruan2021towards} with the implicit assumption that every selected client will respond to the computation requests from the parameter server. 
In parallel, the analysis on fastest responsive clients  \cite{Li2020,philippenko2020bidirectional,kairouz2021advances,wang2022} assumes that each client responds with a known probability. 
 
A handful of work exists on arbitrary client unavailability \cite{yan2020distributed,gu2021fast,wang2022,yang2022anarchic,cho2022towards}. 
Both~\cite{yan2020distributed} and~\cite{cho2022towards} focus on controllable environments, where every client sampled by the parameter server must respond accordingly.  
Non-controllable environments were investigated more recently \cite{gu2021fast,yan2020distributed,wang2022,yang2022anarchic} yet still imposes some structural requirements such as regularized participation \cite{wang2022}, bounded inactive periods \cite{yan2020distributed,yang2022anarchic,gu2021fast}, and asymptotic unbounded inactive periods \cite{gu2021fast}. It is easy to find patterns that violate the aforementioned assumptions. For example, a client may be inactive for a while and become active at some carefully chosen time in order to disturb the learning process.

\subsection{Byzantine-resilient distributed and federated learning.}
Byzantine attack is a canonical adversary model in distributed computing \cite{lynch1996distributed}. 
In general, %Byzantine attacks 
it includes two key components: (A.1) adversarial client selection, i.e., the compromised clients can be selected in the worst possible manner based on the knowledge of the system states, and (A.2) malicious value injection, i.e., the compromised clients inject arbitrary values into the system. 
Moreover, the subset of compromised client may vary over time \cite{bonomi2019approximate,lynch1996distributed,chen2016algorand}, and can be adaptively chosen by the system adversary. 
Tolerating Byzantine attacks in distributed and federated learning have received intensive attention recently \cite{feng2014distributed,sundaram2015consensus,su2016fault,blanchard2017machine,chen2017distributed,yin2018byzantine,xie2019zeno,ghosh2020communication,karimireddy2021learning,karimireddybyzantine22,farhadkhani2022byzantine,allouah2023fixing}. 
Our adversarial unavailability model can be viewed as a special case of Byzantine attacks with adversarial client selection [i.e.\,(A.1)] but no malicious value injection [i.e.,(A.2)]. Nevertheless, as we explain next, to the best of our knowledge, existing Byzantine-resilient results are not applicable to our problem. 
Both (A.1) and (A.2) that are adaptive to history information were considered in \cite{chen2017distributed,yin2018byzantine,su2019securing} yet under the simplified setting such as IID datasets, strongly-convex objectives, and one-step local updates.  
When the datasets are non-IID or unbalanced, unfortunately, the analysis therein breaks apart. 
Assuming the subset of Byzantine clients is pre-selected and fixed throughout training, a more recent line of work~\cite{farhadkhani2022byzantine,karimireddy2021learning,karimireddybyzantine22} focused on (A.2) only. IID local data is considered in \cite{farhadkhani2022byzantine,karimireddy2021learning}. 
Extending their convergence analysis to non-IID settings is challenging unless imposing strong assumptions such as common stationary points %(i.e.~$G=0$) 
or absolute bounded gradient dissimilarity.
%(i.e.~$B=0$). 
It is crucial to assume the subset of Byzantine clients is pre-selected and fixed in \cite{karimireddybyzantine22,allouah2023fixing}. When the system adversary can adaptively choose different subsets of clients, the mean of each Byzantine-free bucket in \cite{karimireddybyzantine22} is no longer unbiased
%\footnote{The method in \cite{karimireddybyzantine22} uniformly at random assigns data into multiple buckets. Its correctness requires, in expectation, the mean of the data in each Byzantine-free bucket is the same as the average of the non-corrupted data. Suppose that there are four data points $\{1, 2, 3, 4\}$, and at most one of them can be corrupted. Suppose that the four data points are uniformly at random partitioned into two groups each with two data points. For each partition, we choose the group with the larger mean and corrupt the smallest element. A simple enumeration shows that the mean of the Byzantine-free bucket is strictly smaller than the average of non-corrupted data.}.  
To tolerate time-varying subsets of Byzantine clients, the $(f,\kappa)$-robustness in \cite[Definition 2]{allouah2023fixing} requirement needs to be imposed on any client subset of proper size, which is hard to ensure unless the involved quantities  follow light-tailed distributions such as subgaussian or subexponential \cite{chen2017distributed,yin2018byzantine,su2019securing}.

\section{System model}
\label{sec: formulation}
% \noindent{\bf Learning objective.}
%\paragraph*{Learning objective.}
%A Federated Learning system consists of 
A parameter server and $M$ clients collaboratively minimize 
\begin{align}
\label{eq: global objective}
\min_{\theta\in \reals^d}F(\theta)=\sum_{i=1}^M w_i F_i(\theta), 
\end{align}
where $F_i(\theta)=\Expect_{z \sim \calD_i}[\ell_i(\theta;z)]$ is the local objective, $w_i$ is the weight, $\calD_i$ is the local data distribution, and $\ell_i(\theta; z)$ is a loss function. The local datasets can be unbalanced across clients. 
Let $n_i$ denote the number of data points generated from the unknown $\calD_i$ at every round by client $i$, 
and $N=\sum_{i=1}^M n_i$ be the total number of data points drawn at every round. 
Following the literature \cite{li2020federated,Li2020,yuan2022,karimireddy2020scaffold}
%\cite{kairouz2021advances}, 
we adopt the following bounded dissimilarity assumption.  
% Surprisingly, we find that the aforementioned two challenges can be tackled by imposing the following bounded dissimilarity condition on $F_i$'s. 
\begin{assumption}[Bounded dissimilarity]\label{ass:BG}

We say that $(w_i, F_i)$ satisfies the $(B,G)$-bounded dissimilarity condition for $B\ge 1$ and $G \ge 0$ if 
$\sum_{i \in [M]} w_i \norm{\nabla F_i(\theta)}^2 \le B^2 \norm{\nabla F(\theta)}^2 + G^2$.

% We say that $(w_i, F_i)$ satisfies the $(B,G)$-bounded dissimilarity condition for $B\ge 0$ and $G \ge 0$ if 
% $\sum_{i \in [M]} w_i \norm{\nabla F_i(\theta) - \nabla F(\theta)}^2 \le B^2 \norm{\nabla F(\theta)}^2 + G^2$.  
% \begin{align}
% \label{eq: bounded dissimilarity}
% \sum_{i \in [M]} w_i \norm{\nabla F_i(\theta) - \nabla F(\theta)}^2 \le B^2 \norm{\nabla F(\theta)}^2 + G^2.
% \end{align}
\end{assumption}
\noindent 
When the local datasets are IID (i.e., $\calD_i = \calD_j$ for any $i, j\in [M]$), % across clients, 
it holds that $B=1$ and $G=0$ as $F_i=F$ for all clients. 
When data is non-IID, %some significantly 
more stringent forms of bounded dissimilarity assumptions 
%special cases of Assumption \ref{ass:BG} 
are popular in the existing literature on  
%work on literature of 
arbitrary client unavailability and Byzantine resilience.
For example, \cite{yan2020distributed} and \cite{cho2022towards} assumed bounded gradients, i.e., $\sum_{i \in [M]} w_i \norm{\nabla F_i(\theta)}^2 \le G^2$. %, which satisfies Assumption \ref{ass:BG}. 
%with $B=0$); %and bounded $\norm{\nabla F(\theta)}$; 
Both \cite{wang2022} and \cite{yang2022anarchic} assumed $\sum_{i \in [M]} w_i \norm{\nabla F_i(\theta) - \nabla F(\theta)}^2 \le G^2$. %required $B\le 2$. 
% For example, the convergence of FedProx was characterized assuming $G=0$ \cite{li2020federated}, i.e., $F_i$ and $F$ share common stationary points. 
% The convergence of FedAvg was first shown in \cite{Li2020} assuming Assumption \ref{ass:BG} with $B=0$ and $\norm{\nabla F(\theta)}$ is bounded. 
Byzantine resilience  was shown in \cite{karimireddybyzantine22,allouah2023fixing} assuming either $G=0$ or $\sum_{i \in [M]} w_i \norm{\nabla F_i(\theta) - \nabla F(\theta)}^2 \le G^2$. \\
%$B=0$, and in \cite{allouah2023fixing} assuming $B=0$.   

\noindent{\bf Non-Static Client Unavailability.}
%\paragraph*{Adversarial Client Unavailability.} 
In each round $t$, first the parameter server uniformly at random selects a set $\tilde{\calS}_t \subset [M]$ of $K$ clients. % without replacement. 
Each client $ i \in \tilde{\calS}_t$
draws a fresh sample $z_{i,t}$ of size $n_i$ from the local dataset $\calD_i$. 
Let $p=K/M$. % denote the fraction of clients that are randomly selected by PS. \nbr{Added here. Not sure if this is the best place.}
Similar client selection was considered in \cite{mcmahan2017communication,kairouz2021advances,jhunjhunwala2022fedvarp}. 
Then the adversary adaptively chooses the set $\calS_t \subset \tilde{\calS}_t$ 
of participating clients.
The adversary's choice of $\calS_t$ may depend on the sets of local data points $\{z_{i,\tau}:
\tau \le t, i \in [M] \}$ drawn by all clients up to time $t$ and 
the sets of participating clients $\{\calS_\tau, \tau \le t-1\}$ chosen previously. We also allow the adversary to have total access to all system parameters including $n_i$ and local objective functions $F_i$. In other words, the adversary is fully aware of everything that happened up to time $t$. 
%his turn to take action. 
Such adaptive choice of ``faulty'' clients is standard 
in distributed computing literature \cite{lynch1996distributed,bonomi2019approximate,chen2016algorand}.

%For ease of exposition, we refer to the clients that are turned non-responsive by the adversary as dropouts. 
To further quantify the degrees of client unavailability, below we impose an upper bound on the total number of data points that are dropped by the adversary at every round $t.$  
\begin{assumption}[$\epsilon$-adversary dropout fraction]\label{ass:adversarial}
$$ \sum_{i\in \tilde\calS_t\setminus\calS_t} n_i
\le \epsilon \frac{KN}{M}, \quad \forall t.$$ 
\end{assumption}
\noindent Roughly speaking, since $\expect{\sum_{i \in \tilde{\calS}_t} n_i }= KN/M$, ~\prettyref{ass:adversarial} says that on average 
%comparing to the average number of data points that are randomly drawn in $\tilde{\calS}_t$,
at most $\epsilon$ fraction of the sampled data points can be further dropped out by the adversary. In the data balanced setting where $n_i=N/M$ for all $i$, 
\prettyref{ass:adversarial} simplifies to $|\tilde{\calS}_t \setminus \calS_t| \le \epsilon K$, that is, at most $\epsilon$-fraction of clients in $\tilde{\calS}_t$ are non-responsive.  

We remark that our adversary model covers random unavailability as a special case. 
With $\epsilon=0$, our adversary model reduces to the uniform-at-random client unavailability. % with uniform probability $K/M$.

\section{Algorithms and convergence guarantees}
\label{sec:convergence}
\noindent{\bf Algorithms.}
We analyze variants of FedAvg and FedProx in this section.
Formally, in each communication round $t$, each client $i \in \calS_t$ updates the model based on the local function $\ell_{i,t}(\theta) \triangleq \frac{1}{n_i}\sum_{j=1}^{n_i} \ell_i(\theta; z_{i,t}^{(j)})$, where $z_{i,t}=(z_{i,t}^{(1)},\dots,z_{i,t}^{(n_i)})$ is a fresh sample batch of size $n_i$:  
\begin{itemize}
\item FedAvg: First, set $\, \theta^{0}_{i,t} = \theta_t$; then perform $s$ local updates as $\theta^{\tau+1}_{i,t} =  \theta^{\tau}_{i,t} - \eta_t \nabla  \ell_{i,t}(\theta^{\tau}_{i,t})$\,
for $\tau=0, \cdots, s-1$; finally, set $\theta_{i,t+1} = \theta^{s}_{i,t}$.
\item FedProx\footnote{We assume the exact minimizer for ease of presentation. Our results can be readily extended when $\theta_{i,t+1}$ is an approximate solution, see e.g., \cite[Definition 2]{li2020federated} and \cite[Definition 3]{yuan2022}.}: 
\begin{align}
\theta_{i,t+1} \in \arg\min_{z} \tilde{\ell}_{i,t}(z) \triangleq \ell_{i,t}(z) + \frac{1}{2\eta_t} \norm{z-\theta_t}^2. \label{eq:FedProx_proximal}
\end{align}
\end{itemize}

%  \begin{itemize} 
%  \item FedAvg:
%  \begin{center}
%  \fbox{
%  \begin{minipage}{22.5em}
% Set $\, \theta^{0}_{i,t} = \theta_t$;\\ 
% Update $\, \theta^{\tau+1}_{i,t} =  \theta^{\tau}_{i,t} - \eta_t \nabla \ell_{i,t}(\theta^{\tau}_{i,t})$\,
% for $\tau=0, \cdots, s-1$; \\  
% Set $\, \theta_{i,t+1} = \theta^{s}_{i,t}$.
% \end{minipage}}
% \end{center}
 
% \item FedProx\footnote{We assume the exact minimizer for ease of presentation. Our results can be readily extended when $\theta_{i,t+1}$ is an approximate solution, e.g., \cite[Definition 2]{li2020federated} and \cite[Definition 3]{yuan2022}.}:
% \begin{equation}
% \label{eq:fedprox}
% \theta_{i,t+1} \in \arg\min_{z} \ell_{i,t}(z) + \frac{1}{2\eta_t} \norm{z-\theta_t}^2.
% \end{equation}
% \end{itemize}
Upon receiving the local model update $\theta_{i,t+1}$ from active clients in $\calS_t$,
%each participating client at round $t$, 
the parameter server computes
\begin{align}
\theta_{t+1}=\theta_t + \beta \sum_{i \in \calS_t} w_i \left( \theta_{i,t+1} - \theta_t \right), \label{eq:global_update}
\end{align}
where $\beta \ge 1$ is a parameter introduced to compensate for %ing for 
the lack of participation of the unavailable and non-sampled clients. 
Note that the standard FedAvg algorithm~\cite{mcmahan2017communication} updates 
$\theta_{t+1}= \theta_t + \sum_{i \in S_t} \frac{w_i}{\sum_{j \in S_t} w_j} \left( \theta_{i,t+1} - \theta_t \right)$ and hence can be viewed as a special instance of~\prettyref{eq:global_update} with varying $\beta=\frac{1}{\sum_{j \in S_t} w_j}$. However, this choice of $\beta$ may lead to unstable performance as we observe in numerical experiments.
%This is because when there exists a relatively large amount of adversarial dropout, $\sum_{j \in S_t} w_j$ may become too small. Numerically, we observe that instead choosing $\beta=M/K$ leads to more stable performance. 
% \begin{remark}
% The global aggregation rule \eqref{eq:global_update} is an affine combination of the received models $\sum_{i \in \calS_t} w_i \theta_{i,t+1}$ and the previous model $\theta_t$. 
% The factor $\beta$ compensates for the lack of participation from the unavailable clients. 
% A similar amplification idea has been used in~\cite{wang2022}. 
% However, they only apply the amplification every $P$ rounds for some large $P$ under the assumption that the unavailability follows benign stochastic patterns. 
% \cite{gu2021fast} compensates for unavailable clients by using the latest local updates. However, the information could be outdated or even nonexistent if the adversary keeps this client non-responsive for a long period.  
% \end{remark}
% \medskip

\begin{assumption}
We assume that
% Each local gradient function $\nabla \ell_{i,t}(\theta)$ is $L_i$-Lipschitz for all $t$, i.e., %that is,
\begin{equation}
\label{eq:L-Lipschitz}
    \norm{\nabla \ell_{i,t}(u)-\nabla \ell_{i,t}(v)}\le L_i \norm{u-v}.
\end{equation}
For FedProx, we additionally assume that
\begin{equation}
    \label{eq:fedprox-assumption}
    \lambda_{\min}(\nabla^2 \ell_{i,t}(\theta)) \ge -L_-,\quad \forall i, ~\text{and}~\forall t.
\end{equation}
\end{assumption}
% Eq.\,\eqref{eq:L-Lipschitz} %Such smoothness condition 
The conditions are commonly used in the convergence analysis of federated learning algorithms (see e.g.~\cite{li2020federated,yuan2022}).
Define $L\triangleq \max_{i\in[M]}L_i$. 
% Eq.~\eqref{eq:fedprox-assumption} is commonly adopted in the analysis of proximal gradient descent \cite{li2020federated}.
Eq.~\eqref{eq:fedprox-assumption} ensures that, using a sufficiently small $\eta_t$, the local proximal program~\prettyref{eq:FedProx_proximal} in FedProx is strongly convex, and hence the solution can be efficiently computed. 
Nevertheless, the local objective functions $\ell_{i,t}$ and $F_i$ may still be non-convex.
% not be strongly convex and the population objective functions $F_i$ and $F$
% may not be strongly convex either.

Let $\calF_t$ denote the filtration generated by the sequence $ \{z_{i,\tau},\calS_{\tau}:\tau\le t-1, i\in [M]\}.$ 
Then $\theta_t\in\calF_t$.
It follows that $\tilde \calS_t$ and $z_{i,t}$ for $i\in[M]$ are independent of $\calF_t$.
\begin{assumption}\label{ass:noisy-gradients}
We assume that
% We assume the regularity condition on $\nabla \ell_{i,t}(\theta_t)$:
% is an unbiased estimate of $\nabla F_i(\theta_t)$ conditioned on $\calF_t$:
\begin{align}
& \expect{\nabla \ell_{i,t}(\theta_t) \mid \calF_t}= F_i(\theta_t), \label{eq:sgd-mean}\\
& \expect{\norm{\nabla \ell_{i,t}(\theta_t) - F_i(\theta_t) }^2 \mid \calF_t} \le \sigma_i^2.\label{eq:sgd-variance}
\end{align}
Let $\sigma^2=\sum_{i=1}^M w_i \sigma_i^2$ denote the average noise level. 
\end{assumption}

% We consider both non-convex and strongly-convex global objectives $F$. 

\subsection{Non-convex functions}
\label{sec:non-convex}
% {\red Overall, sections 4.1 and 4.2 contain too many bullets points. }
% \nbr{update $\sigma^2/n$ to $\sigma^2$}
% We study the general non-convex setting in this subsection. 
% In addition to demonstrating that the error rate in \prettyref{thm:informal} can be achieved by simple variants of federated learning algorithms provided in \prettyref{sec:convergence}, we also show that the convergence speed is comparable to the classical centralized optimization algorithms to the set of approximate stationary points. 
%Let $F_{\min}\triangleq \inf_\theta F(\theta)$ denote the minimum value of the global objective function. %which is possibly infinite. 

\begin{theorem}
\label{thm:nonconvex}
Let $F_{\min}\triangleq \inf_\theta F(\theta) >-\infty$, Assumptions~\ref{ass:BG}--\ref{ass:noisy-gradients} hold, and $\sqrt{\epsilon} B \le 0.1$.
Let $R$ be the random time with
$
\Prob[R=k] = \eta_k/\sum_{t=0}^T \eta_t
$
for $k=0,\dots, T.$
For FedAvg, choose $\eta_t \le \frac{1}{10 \beta  sLB^2}$ for all $t$;
For FedProx, choose $\eta_t\le \frac{1}{10 \beta LB^2} \wedge \frac{1}{10 L_-}$ for all $t$. 
    % \begin{itemize}
    % \item FedAvg: under Assumptions \ref{ass:BG} and \ref{ass:adversarial}, choose $\eta_t \le \frac{1}{10 \beta  sLB^2}$ for all $t$.
    % \item FedProx: 
    % under the additional assumption in Eq.\eqref{eq:fedprox-assumption}, choose $\eta_t\le \frac{1}{10 \beta LB^2} \wedge \frac{1}{10 L_-}$ for all $t$. 
    % \end{itemize}
Then, there exists a universal constant $c$ such that
\begin{align}
\Expect \norm{\nabla F(\theta_R)}^2 \nonumber 
& \le \frac{3 \left( F(\theta_0)-F_{\min} \right) }{\beta p s \sum_{t=0}^T \eta_t} \\
& \quad + \pth{ 4 \epsilon +  \frac{ c \beta sL \sum_{t=0}^T \eta_t^2 }{ \sum_{t=0}^T \eta_t }}\pth{G+\sigma}^2, \label{eq:gradient-speed}
\end{align}
where $s=1$ for FedProx.
Consequently, let $\hat\theta=\theta_R$ with $\sum_{t=0}^T \eta_t \to \infty$ and $\frac{\sum_{t=0}^T \eta_t^2}{\sum_{t=0}^T \eta_t}\to 0$. Then,
\begin{align}
\label{eq:non-convex_rate}
\lim_{T\to\infty} \Expect\| \nabla F(\hat\theta)\|_2^2 \le 4 \epsilon \pth{G+\sigma}^2.   
\end{align}
\end{theorem}

Below we show the convergence rates under some concrete schemes of the learning rate based on~\prettyref{thm:nonconvex}. 
In the first case, $\eta_t$ depends on the eventual termination time $T$, which may be chosen based on a prescribed level of optimality;
in the second case, $\eta_t$ is independent of $T$, so the program can be executed indefinitely.
% until the desired optimality is attained. 
We only present the results for the FedAvg, and the results for the FedProx are entirely analogous yet with $s=1$. The proof is deferred to~\prettyref{app:pf-cor-rate}. 
% 
% We obtain the following convergence rates of FedAvg and FedProx 
% % for the FedAvg update rule 
% under two specific choices of the learning rates. 
Let $\Delta = F(\theta_0)-F_{\min}$.
Recall that $p=K/M$. 
\begin{corollary}
\label{cor:convergence-rate}
% Let $\Delta = F(\theta_0)-F_{\min}$.
% We have the following convergence rate for the FedAvg update rule under specific choices of the learning rates:
For FedAvg: Choose $\eta_t= \frac{1}{10 \beta  LB^2} \wedge \frac{1}{\beta  \sqrt{p T L} (G+\sigma)}$. Then
\begin{align*}
&\Expect \norm{\nabla F(\theta_R)}^2
\le 
\frac{30\Delta LB^2}{ p s T } +
\sqrt{\frac{L}{pT}}\left(3\Delta + c s \right) \pth{G+\sigma}
+ 4 \epsilon \pth{G+\sigma}^2.
%O \pth{ \epsilon \pth{G+\sigma}^2}  + O \pth{\Delta L B^2/(psT) + (\Delta+1) \pth{G+\sigma}s \sqrt{L/pT}}.
\end{align*}
%{\blue (LS: We'd better have $s$ show up. )}
%For FedProx: 
Choose $\eta_t = \frac{\eta_0}{\sqrt{t+1}}$ for all $t\ge 0$, where $ \eta_0=\frac{1}{10 \beta  LB^2} \wedge \frac{1}{\beta  \sqrt{pL}(G+\sigma)}$. Then
\begin{align*}
&\Expect \norm{\nabla F(\theta_R)}^2
\le  \frac{30\Delta L B^2}{ p s \sqrt{T} }
+ \sqrt{\frac{L}{pT}}\left(3\Delta + cs \log\left(e(T+1) \right)  \right) \pth{G+\sigma}
+ 4\epsilon \pth{G+\sigma}^2 .
%O\pth{ \epsilon \pth{G+\sigma}^2 }  + O \pth{\Delta L B^2/(ps\sqrt{T}) +  (\Delta+1) \pth{G+\sigma}s\sqrt{L\log^2 T/pT}}.
\end{align*}
% \begin{itemize}
% \item Suppose $\eta_t= \frac{1}{10 \beta  sLB^2} \wedge \frac{1}{\beta s \sqrt{p T L} (G+\sigma)}$. Then
% \begin{align*}
% &\Expect \norm{\nabla F(\theta_R)}^2
% \le O \pth{ \epsilon \pth{G+\sigma}^2}  + O \pth{ 
% \frac{ \Delta L B^2}{ pT } 
% + (\Delta+1) \pth{G+\sigma}\sqrt{\frac{L}{pT} }}.
% \end{align*}
% \item Suppose $\eta_t = \frac{\eta_0}{\sqrt{t+1}}$ for all $t\ge 0$, where $ \eta_0=\frac{1}{10 \beta  sLB^2} \wedge \frac{1}{\beta s \sqrt{pL}(G+\sigma)}$. Then
% \begin{align*}
% &\Expect \norm{\nabla F(\theta_R)}^2
% \le O\pth{ \epsilon \pth{G+\sigma}^2 }  + O \pth{ 
% \frac{ \Delta L B^2}{ p\sqrt{T} } 
% + (\Delta+1) \pth{G+\sigma}\sqrt{\frac{L\log^2 T}{pT} }}.
% \end{align*}
% % \item Suppose $\eta_t = ... \frac{1}{t+1}$. Then
% \end{itemize}
\end{corollary}

The rate $O(1/\sqrt{T})$ matches that of the standard stochastic gradient descent for optimizing non-convex functions \cite{ghadimi2013stochastic}. In fact, the rate $O(1/\sqrt{T})$ is the best possible for any first-order method that has only access to noisy gradients (see e.g.~\cite[Theorem 3]{arjevani2022lower}). %\nbr{to save space, we can keep just the first result in the corollary.} \ls{Let's keep both for now.}

\subsection{Strongly convex functions}
\label{sec:convex}
%\nbr{talk about convex, but not strongly-convex case?}
%We say 
We say $F$ is $\mu$-strongly convex %for some $\mu>0$ 
if
\begin{align}
\iprod{\nabla F(x)- \nabla F(y)}{x-y} \ge 
\mu \norm{x-y}^2,
\quad \forall x, y. \label{eq:def_strong_convexity}
\end{align}
% When $F$ is strongly convex, \prettyref{thm:nonconvex} implies that $\hat\theta$ converges to a region near the unique optimum $\theta^*=\argmin F(\theta)$. 
% The next result shows a faster convergence speed. 

It is worth noting that the convergence results here only require the global objective $F$ to be strongly convex, while we allow non-convex local objective $F_i$.
% ; each local objective $F_i$ can still be non-convex.  
This is slightly more relaxed compared with 
the typical assumptions that
% what is typically imposed in the federated learning literature. For instance, 
the local population functions $F_i$ at every client are $\mu$-strongly convex in~\cite{Li2020,karimireddy2020scaffold,gu2021fast}.

% \nbr{clarify $c,\gamma$? Try to clarify every parameters shown in theorem statement... $\theta$ notation abuse...}
\begin{theorem}\label{thm:strongly_convex}
Suppose that $F$ is $\mu$-strongly convex and $B \sqrt{\epsilon} < 0.1 \frac{\mu}{L}$, and Assumptions~\ref{ass:BG}--\ref{ass:noisy-gradients} hold. Choose $\eta_t = \alpha/(t+\gamma)$,
where  $\alpha, \gamma$ are constants such that 
$\alpha \ge 2/(\beta ps \mu)$, $\alpha/\gamma \le \frac{\mu}{20\beta L^2B^2}  $,
and further $\alpha/\gamma \le  \frac{1}{10L_-}$ for FedProx. 
% \begin{itemize}
% \item For FedAvg, and $\beta ps \theta \mu \ge 2 $ and $\eta_t \le \frac{\mu}{10\beta s L^2 B^2}$ for all $t$. 
% \item For FedProx, suppose that Assumptions~\ref{ass:BG}, \ref{ass:adversarial} and t, and choose $\eta_t = \theta/(t+\gamma)$ for constants $\theta, \gamma$ such that $\rho=c\beta p \theta \mu >1$ and $\eta_t  \le \frac{\mu}{10\beta L^2B^2} \wedge \frac{1}{10L_-}$ for all $t$.
% \end{itemize}
% \begin{itemize}
% \item FedAvg: under Assumptions~\ref{ass:BG} and \ref{ass:adversarial}, choose $\eta_t = \theta/(t+\gamma)$ for constants $\theta, \gamma$ such that 
%  $\rho=c\beta ps \theta \mu>1$ and $\eta_t \le \frac{\mu}{10\beta s L^2 B^2}$ for all $t$. 
% \item FedProx: under the additional assumption \eqref{eq:fedprox-assumption}, choose $\eta_t = \theta/(t+\gamma)$ for constants $\theta, \gamma$ such that $\rho=c\beta p \theta \mu >1$ and $\eta_t  \le \frac{\mu}{10\beta L^2B^2} \wedge \frac{1}{10L_-}$ for all $t$. 
% \end{itemize}
Then, there exist some universal constant $c$,
\begin{align*}
\Expect \norm{\theta_t-\theta^*}^2 
\le &  \left( 1+ \frac{t}{\gamma} \right)^{-1.1} \norm{\theta_0-\theta^*}^2  + \frac{4 \epsilon}{\mu^2} \pth{ G+ \sigma}^2 \\
& + 
\frac{c}{p\mu^2} \pth{ G+ \sigma}^2
\frac{1}{t+\gamma}.
\end{align*}
\end{theorem}
  Note that $\mu/L$ is an upper bound to the condition number of Hessian matrix $\nabla^2 F.$
    Thus, our standing condition  $B \sqrt{\epsilon} < 0.1 \frac{\mu}{L}$
    indicates that a larger fraction of adversarial dropouts
    can be tolerated when the population function $F$ is better conditioned.

\prettyref{thm:strongly_convex} shows that $\theta_t$ converges to $\theta^*$ at a rate of $O(1/t)$. 
    The $O(1/t)$ convergence rate is the best possible for any first-order method that has only access to noisy gradients \cite{nemirovski2009robust}, \cite[Theorem E.1]{gu2021fast}.
    When there is no adversarial dropout, our results reduce to the state-of-art convergence results for FedAvg or FedProx. For instance, an exponential decay term similar to $(1+t/\gamma)^{-1.1} \norm{\theta_0-\theta^*}$ and a linear decay term similar to $(\sigma^2 +G^2 )/t$ also appear in~\cite[Theorem V]{karimireddy2020scaffold}. The convergence results in \cite{Li2020} are similar but a bit weaker: there is a linear decay term similar to $(\sigma^2+ G^2) /t$, but the error bound decays only linearly rather than exponentially in the initial error as $\norm{\theta_0-\theta^*}/t$.
    
%Here are some further interpretations on \prettyref{thm:strongly_convex}:
%\begin{enumerate}%[label = (\arabic*)]
    % \item Our assumption of strongly convex is slightly more relaxed compared with what is typically imposed in the federated learning literature. For instance, the local population functions $F_i$ at every client are assumed to be $\mu$-strongly convex in~\cite{Li2020,karimireddy2020scaffold,gu2021fast}.
    %\item 
   
%\end{enumerate}

\begin{remark}[Convex objective functions]
% but not necessarily strongly-convex functions
\prettyref{thm:strongly_convex} establishes the convergence of $\theta_t$ to $\theta^*$ in the squared $L_2$ norm for strongly-convex functions $F$. It is tempting to ask whether we can establish the convergence of $F(\theta_t)$ to $F(\theta^*)$
for convex (but not necessarily strongly-convex) functions $F$, which is possible for standard stochastic gradient descent without adversarial dropouts (cf.~\cite[Theorem 2.1]{ghadimi2013stochastic}). If we assume, in addition, that $\norm{\theta_t -\theta^*}$ is always bounded by $C$, then by convexity of $F$  we can get that 
$$
F(\theta_t) - F(\theta^*) \le \iprod{\nabla F(\theta_t)}{\theta_t-\theta^*} 
\le C \norm{\nabla F(\theta_t)}, 
$$
which can be further bounded using the convergence of $\norm{\nabla F(\theta_t)}$ established in~\prettyref{thm:nonconvex}. It remains open to establish the convergence for function values for general convex functions without assuming the boundedness of $\norm{\theta_t -\theta^*}$. The main technical hurdle lies in the fact that the deviations caused by the objective inconsistency and the adversarial selections depend on $\norm{\theta_t -\theta^*}$, which can be potentially much larger than 
$\nabla F(\theta_t)$ or $F(\theta_t) - F(\theta^*)$, when $F$ is flat around the minimum point $x^*$. 
\end{remark}

\section{Proofs of the main convergence guarantees}
We provide proof of our main theorems and discuss the connections and differences from existing SGD analysis. 
The proof focuses on the FedAvg in the general non-convex functions setting.  
The cases for FedProx or strongly convex functions can be shown via % from 
similar arguments and deferred to appendices. 
%, and the differences will be sketched toward the end of this subsection. 

\subsection{Key challenges} Recall that each $F_i$ is $L_i$-smooth.
Hence, $F$ is $L$-smooth with $L=\max_i L_i$ and thus
%Following the standard analysis of smooth objective functions,
\begin{equation}
\label{eq:decrease-objective}
F(\theta_{t+1})
\le F(\theta_t) + \iprod{\nabla F(\theta_t)}{\theta_{t+1}-\theta_t} + \frac{L}{2}\norm{\theta_{t+1}-\theta_t}^2.
\end{equation}
The progress over one communication round $\theta_{t+1}-\theta_t$ per aggregation rule~\eqref{eq:global_update} is given by 
%\[
$
\theta_{t+1}-\theta_t = \beta\sum_{i\in\calS_t} w_i(\theta_{i,t+1}-\theta_t).
$
%\]
% 
The classical SGD iteration relies on the unbiasedness of $\theta_{i,t+1}-\theta_t$ for the direction of $-\nabla F(\theta_t)$.
The unbiasedness fails to hold due to the following two reasons:
\begin{itemize}
    \item {\bf Objective inconsistency}. % For the sake of communication efficiency and system heterogeneity, 
    Under FedAvg or FedProx, the clients either run multiple-step of local SGD or solve a subroutine with proximity regularization. The bias of local updates arises from objective inconsistency or client-drift as analyzed by \cite{wang2020tackling,karimireddy2020scaffold} for FedAvg. 
    %The objective consistance is also closely related to the client-drift studied in~\cite{karimireddy2020scaffold}. 
    Define
    \begin{align}
    B_t^{\sf obj} \triangleq \sum_{i\in\calS_t} w_i(\theta_{i,t+1}-\theta_t - (- s\eta_t \nabla \ell_{i,t}(\theta_t)) ). \label{eq:B_obj}
    \end{align}
    \item {\bf Selection bias}. Given the sampled clients $\tilde{S}_t$, the non-responsive clients $\tilde{\calS}_t\setminus\calS_t$ are selected by 
    %The participating clients have been selected 
    by the system adversary; consequently, the resulting participating clients $\calS_t$ no longer form %; hence,  and, as a result, may not be 
    a representative subset of all clients. What's more, the adversarial selection is time-varying and may even be \emph{correlated} with the data used to evaluate the local gradients. 
    %The impact is similar to existing analyses in the literature of robust statistics \cite{huberrobust,diakonikolas2019recent}. 
    Define
    \[
    B_t^{\sf sel} \triangleq \sum_{i\in \tilde\calS_t\setminus\calS_t}w_i \nabla \ell_{i,t}(\theta_t).
    \]
\end{itemize}
Our simple variants of FedAvg and FedProx can provably control the biases stemming from the two sources identified above. 
Existing bias reduction methodologies such as those proposed by \cite{wang2020tackling,karimireddy2020scaffold} only address the issue of objective inconsistency. 

%Therefore, the major challenge is to quantify the biases from the above two sources carefully. 
%and to prove that the estimation error matches our lower bound in \prettyref{sec: challenges and limits} within constant factors. 
%Next, we show the main ideas; the proof details are deferred to \prettyref{app:pf-ub}.

\subsection{Objective inconsistency}

We first introduce some convenient notations for the local update rules for FedAvg.
%\begin{compactitem}
%\item FedAvg: 
Let $\calG_{i,t}(\theta;\eta)$ denote the mapping of the gradient descent on client $i$ in round $t$ with the learning rate $\eta$:
\begin{equation}\label{eq:gradient}
\calG_{i,t}(\theta;\eta)\triangleq \theta - \eta \nabla \ell_{i,t}(\theta).
\end{equation}
% \ls{Write this in the form of for-loop?}
Then, the locally updated model after $s$ steps of gradient descent is
$
\theta_{i,t+1} = \calG_{i,t}^s(\theta_t,\eta_t),
$
where 
\[
\calG_{i,t}^s(\cdot;\eta)\triangleq \underbrace{\calG_{i,t}(\cdot;\eta)\circ \dots \circ \calG_{i,t}(\cdot;\eta)}_{s~\text{times}}. 
\]
% \item FedProx: let $\calP_{i,t}(\theta,\eta)$ denote proximal mapping defined as
% \begin{equation}
% \label{eq:proximal}
% \calP_{i,t}(\theta;\eta) \triangleq \arg\min_{z} \ell_{i,t}(z) + \frac{1}{2\eta} \norm{z-\theta}^2.
% \end{equation}
% Then $\theta_{i,t+1}\in \calP_{i,t}(\theta_t;\eta_t)$.
% \end{compactitem}
After collecting the updated model $\theta_{i,t+1}$ for $i\in\calS_t$, PS aggregates the local updates via \eqref{eq:global_update}.

%\subsubsection{Objective inconsistency.} 
% It follows from our aggregation rule~\eqref{eq:global_update} that, for every $t$,
% \begin{align}
% \theta_{t+1} - \theta_{t}
% & = \beta \sum_{i\in\calS_t} w_i (\calG_{i,t}^s(\theta_t;\eta_t)-\theta_t) \nonumber\\
% & = \beta \sum_{i\in\calS_t} w_i (\calG_{i,t}^s(\theta_t;\eta_t)-\theta_t+s\eta_t \nabla \ell_{i,t}(\theta_t))
% - \beta s \eta_t \sum_{i\in\calS_t}  w_i\nabla \ell_{i,t}(\theta_t). \label{eq:one-round}
% \end{align}
% On the right-hand side of~\eqref{eq:one-round}, the second term is similar to the usual stochastic gradient, where we need to deal with the existence of adversarial dropout.
Choosing a large $s$ accelerates the training process. 
However, it also increases the deviation from the stochastic gradient quantified in~\prettyref{eq:B_obj}, and possibly renders training process unstable as observed in  \cite{mcmahan2017communication}.
% \ls{What is the intuition of the power $s$? }
To quantify the stability of multiple local gradient steps, define
% \ls{What is "stability"?} 
\[
\kappa \triangleq \max_{i,t}\frac{(1+\eta_t L_i)^s - 1- s \eta_t L_i}{\binom{s}{2}(\eta_t L_i)^2}.
\]
% \ls{Need to cast intuition of the definition of $\kappa$. }
Intuitively, $\kappa$ characterizes the deviation of multiple local gradient descent from a single gradient descent with a larger step size, as shown in \prettyref{lmm:s-steps-diff}.
For the special case that $s=1$, we have $\kappa=0$;
for $s\ge 2$, it always holds that $\kappa \ge 1$.
% \ls{The following analysis does not cover the special case when $s=1$. }
Furthermore, $\kappa \le \frac{e^c-1-c}{c^2/2}$ if $\eta_t \le \frac{c}{sL}$.
We have the following lemma, which upper-bounds the deviation from the desired direction uniformly for all $\theta. $

\begin{lemma}\label{lmm:s-steps-diff} 
    For $s\ge 1$, we have
    \[
    \norm{\theta-\calG_{i,t}^s(\theta;\eta_t) -s\eta_t \nabla \ell_{i,t}(\theta)}
    \le \kappa \eta_t^2\binom{s}{2} L_i \norm{\nabla \ell_{i,t}(\theta)},\quad \forall\theta.
    \]  
\end{lemma}
With~\prettyref{lmm:s-steps-diff}, we can now bound the bias due to objective inconsistency as follows:
\begin{align*}
 \norm{B_t^{\sf obj}} & = \norm{\sum_{i\in\calS_t} w_i (\calG_{i,t}^s(\theta_t;\eta_t)-\theta_t+s\eta_t \nabla \ell_{i,t}(\theta_t))} \\
& \le 
\sum_{i\in\calS_t} w_i
\norm{\calG_{i,t}^s(\theta_t;\eta)-\theta_t+s\eta_t \nabla \ell_{i,t}(\theta_t)} \\
& \overset{(a)}{\le} 
\kappa \eta_t^2 L \binom{s}{2} \sum_{i\in\calS_t} w_i \norm{\nabla \ell_{i,t}(\theta_t) } \overset{(b)}{\le} 
\kappa \eta_t^2 L \binom{s}{2} \sum_{i\in \tilde\calS_t} w_i \norm{\nabla \ell_{i,t}(\theta_t)}, 
\end{align*}
where inequality (a) holds from~\prettyref{lmm:s-steps-diff} and inequality (b) is true because that $\calS_t\subseteq \tilde\calS_t$.   
%\nbr{Add transition from $\nabla \ell_{i,t}(\theta_t)$ to $\nabla F_i(\theta_t)$}

Note that 
\begin{align}
\expect{\sum_{i\in \tilde\calS_t} w_i \norm{\nabla \ell_{i,t}(\theta_t)  } \mid \calF_t }
& \le \expect{\sum_{i\in \tilde\calS_t} w_i \norm{\nabla F_i(\theta_t)} \mid \calF_t} + \expect{\sum_{i\in \tilde\calS_t} w_i \norm{\nabla \ell_{i,t}(\theta_t) - \nabla F_i(\theta_t)} \mid \calF_t }. \label{eq:obj_decomp}
\end{align}
It remains to bound the two terms in the RHS of ~\prettyref{eq:obj_decomp}
separately. First, we have
%$\expect{\sum_{i\in \tilde\calS_t} w_i \norm{ \nabla F_i(\theta_t) } \mid \calF_t }$. We have  
\begin{align}
\expect{\sum_{i\in \tilde\calS_t} w_i \norm{ \nabla F_i(\theta_t) } \mid \calF_t } 
& = \expect{\sum_{i=1}^M w_i \norm{ \nabla F_i(\theta_t) }\indc{i \in \tilde\calS_t}  \mid \calF_t }  \nonumber \\
& \overset{(a)}{=} p \sum_{i\in [M]} w_i \norm{ \nabla F_i(\theta_t) }\nonumber \\
&\overset{(b)}{\le} p \sqrt{ \sum_{i \in [M]} w_i \norm{\nabla F_i(\theta_t)}^2  } \nonumber\\
& \overset{(c)}{\le} p \sqrt{ B^2 \norm{\nabla F(\theta_t)}^2 + G^2} \nonumber\\\
&\le p\pth{ B \norm{\nabla F(\theta_t)} + G}, 
\label{eq: nonconvex: convergence: term II remains 1}
\end{align}
where equality (a) is true by the independent between $\tilde\calS_t$ and $\calF_t$;
inequality (b) holds by Jensen's inequality;
inequality (c) follows from~\prettyref{ass:BG}.  
Analogously, 
\begin{align}
\expect{\sum_{i\in \tilde\calS_t} w_i \norm{\nabla \ell_{i,t}(\theta_t) -\nabla F_i(\theta_t) } \mid \calF_t }
& = p  \sum_{i\in [M]} w_i \expect{\norm{\nabla \ell_{i,t}(\theta_t) -\nabla F_i(\theta_t) } \mid \calF_t }\nonumber \\
& \overset{(a)}{\le} p \sqrt{ \sum_{i\in [M]} w_i
\expect{ 
\norm{\nabla \ell_{i,t}(\theta_t) -\nabla F_i(\theta_t)}^2 \mid \calF_t }} \nonumber \\
& \overset{(b)}{\le} p \sqrt{ \sum_{i \in [M] } w_i \sigma_i^2} = p \sigma,
\label{eq: nonconvex: convergence: term II remains 2} 
\end{align}
where inequality (a) is true by  applying Jensen's inequality  twice, and inequality (b) follows from~\eqref{eq:sgd-variance}.
Combining \prettyref{eq:obj_decomp}, \eqref{eq: nonconvex: convergence: term II remains 1}, and \eqref{eq: nonconvex: convergence: term II remains 2}, we get 
\begin{align}
\expect{\sum_{i\in \tilde\calS_t} w_i \norm{\nabla \ell_{i,t}(\theta_t)  } \mid \calF_t }
% & \le \expect{\sum_{i\in \tilde\calS_t} w_i \norm{\nabla F_i(\theta_t)} \mid \calF_t} + \expect{\norm{\nabla \ell_{i,t}(\theta_t) - \nabla F_i(\theta_t)} \mid \calF_t } \nonumber\\
 \le p\left( B \norm{\nabla F(\theta_t)} + G +\sigma \right).  
\label{eq:noisy_gradient_bound}
\end{align}
Therefore,
\begin{align}
\expect{\norm{B_t^{\sf obj}}\mid \calF_t }
% &\le 
% p \norm{\nabla F(\theta_t)}
% \kappa \eta_t^2 L \binom{s}{2}
% \left( \sum_{i\in [M]} w_i 
%  \norm{\nabla F_i(\theta_t)} + \frac{\sigma}{\sqrt{n}} \right) \nonumber \\
 %& 
 \le p 
\kappa \eta_t^2 L \binom{s}{2}
\left( B \norm{\nabla F(\theta_t)} + G +\sigma \right). \label{eq:objective-inconsistency}
\end{align}

%We first upper-bound the deviation from the desired direction as
%\[
%\Norm{\theta_{i,t+1} - \theta_t  - (- s \eta_t \nabla \ell_{i,t}(\theta_t))}
%\lesssim s^2\eta_t^2 L \norm{\nabla \ell_{i,t}(\theta_t)}, 
%\]
%\emph{uniformly} for all possible $\theta_t,$
%provided that the local learning rate $\eta_t\lesssim \frac{1}{sL}$.
%The precise result is stated in \prettyref{lmm:s-steps-diff} in Appendix \ref{sec:lemma_proof}. 
%For each client, the stochastic variation of $\norm{\nabla \ell_{i,t}(\theta_t) - \nabla F_i(\theta_t)}$ follows from \eqref{eq:sgd-variance}.
%Then, %thanks to the bounded dissimilarity in 
%by~\prettyref{ass:BG}, the collective deviation is upper bounded by
%\begin{equation}
%\label{eq:objective-inconsistency}
%\expect{\sum_{i\in\tilde\calS_t} w_i s^2 \eta_t^2 L \norm{\nabla \ell_{i,t}(\theta_t)}  \mid \calF_t }
%\le ps^2 \eta_t^2 L \pth{B\Norm{\nabla F(\theta_t)} + G + \sigma}. 
%\end{equation}

% 
\subsection{Selection bias} 

Since $\calS_t$ is time-varying and is correlated with $\nabla \ell_{i,t}(\theta_t)$, we upper-bound the selection bias \emph{uniformly} via the analysis of extreme values. 
Two main ingredients in our analysis are: 1) the power of the adversary is limited as specified by \prettyref{ass:adversarial}; 2) the dissimilarity among the clients is bounded given by \prettyref{ass:BG}.

%Next, we control the gradients of clients belonging to $\tilde\calS_t \backslash \calS_t$. 
%are controlled in \prettyref{lmm:adversarial-attack} that we prove next.
% analyzed in the next lemma. 
% \begin{lemma}
%     \label{lmm:adversarial-attack}
%     Under Assumptions~\ref{ass:adversarial} -- \ref{ass:BG}, we have
%     \[
%     \expect{
%     \norm{\sum_{i\in \tilde\calS_t/\calS_t}w_i \nabla \ell_i(\theta_t)} \mid \calF_t
%     }
%     \le p \sqrt{\epsilon} \pth{ B \norm{\nabla F(\theta_t) } + G + \frac{\sigma}{\sqrt{N}} }.
%     \]
% \end{lemma}
\begin{lemma}
    \label{lmm:adversarial-attack}
    Under Assumptions~\ref{ass:BG} and \ref{ass:adversarial}, we have
    \begin{align}
    \expect{
    \big\|\sum_{i\in \tilde\calS_t \backslash \calS_t}w_i \nabla \ell_{i,t}(\theta_t) \big\|_2 \mid \calF_t
    }
    \le  p \sqrt{\epsilon} \pth{ B \norm{\nabla F(\theta_t) } + G + \sigma }.
    \label{eq:selection-bias}
    \end{align}
    % where $n\triangleq N/M=\frac{1}{M}\sum_{i=1}^M n_i$ denotes the average sample size per client. 
\end{lemma}

% It is worth noting that~\cite[Lemma 8]{karimireddy2020scaffold} obtained a bound comparable to~\prettyref{eq:objective-inconsistency}.
% The proof uses a similar idea based on the induction, but the exact steps are different and hence the final bound is also different. 
% {\blue (LS) different in what? need to elaborate on this. If possible, mention the technical difficulties of our bound compared with their bound. }\nbr{Let's just delete this. Although Lemma 8 shares a similar idea, the statement is still very different. They  bound something like $\sum_{i \in [M]} \sum_{\ell=1}^s \norm{\calG_{i,t}^{\ell}(\theta; \eta_t) - \theta}$. Also, they do not consider the adversarial selection.}

% Combining those two ingredients, we prove that %the deviation is at most 
% \begin{equation}
% \label{eq:selection-bias}
% % \norm{\sum_{i\in \tilde\calS_t/\calS_t}w_i \nabla \ell_{i,t}(\theta_t)}
% \expect{\norm{B_t^{\sf sel}} \mid \calF_t } 
% \lesssim 
% p \sqrt{\epsilon} \pth{ B \norm{\nabla F(\theta_t) } + G + \sigma }
% \end{equation}
% % {\blue (the notation $p$ is not introduced in the mainbody... JX and PK,please check the consistency between the mainbody result presentation and the appendix writing.)}
% uniformly over all possible dropout patterns on round $t$. The result is stated in \prettyref{lmm:adversarial-attack} (presented in Appendix \ref{sec:lemma_proof}). 

The above lemma gives an upper bound to $\expect{ \norm{B_t^{\sf sel}} \mid \calF_t}$. Surprisingly, $\epsilon$-fraction of maliciously chosen unavailable clients can contribute a deviation of $O(\sqrt{\epsilon})$.
The rate turns out to be optimal as the estimation error matches the fundamental limit in \prettyref{thm:nonconvex-lb-dropout-G}.

\subsection{Combining together}
Once we have tight upper bounds on the bias, the remaining steps are mostly similar to the standard SGD analysis. Specifically, 
\begin{align*}
\theta_{t+1}-\theta_t 
 = \beta\sum_{i\in\calS_t} w_i(\theta_{i,t+1}-\theta_t) 
 & = - \beta s \eta_t \sum_{i\in\calS_t} w_i \nabla \ell_{i,t}(\theta_t) + \beta B_t^{\sf obj} \\
 &= - \beta s \eta_t \sum_{i\in\tilde \calS_t} w_i \nabla \ell_{i,t}(\theta_t) + \beta B_t^{\sf obj} + \beta s \eta_t B_t^{\sf sel}, 
\end{align*}
where $\Expect[\Norm{B_t^{\sf obj}} | \calF_t ]$ and $\Expect[ \Norm{B_t^{\sf sel}} | \calF_t ]$ are upper bounded by~\eqref{eq:objective-inconsistency} and \eqref{eq:selection-bias}, respectively. 

For the first term, recall that $\tilde \calS_t$ is the set of the $K$ clients randomly sampled at round $t$. 
Since $\theta_t$ is adapted to $\calF_t$, we have 
\begin{align}
\label{eq:tildeS-mean}
\nonumber
\expect{ \sum_{i\in \tilde \calS_t}  w_i\nabla \ell_{i,t}(\theta_t) \mid \calF_t}
&=\expect{  \sum_{i=1}^M w_i \nabla \ell_{i,t}(\theta_t) \indc{i \in \tilde \calS_t} \mid \calF_t} \\
\nonumber
&\overset{(a)}{=} \sum_{i=1}^M w_i \expect{\nabla \ell_{i,t}(\theta_t)\mid \calF_t} \expect{\indc{i \in \tilde \calS_t} \mid \calF_t}\\
& \overset{(b)}{=}\sum_{i=1}^M w_i \nabla F_i(\theta_t) p 
\overset{(c)}{=} p\nabla F(\theta_t), 
\end{align}
where (a) holds because $\tilde \calS_t \indep z_{i,t} | \calF_t$;
(b) follows from the sampling fraction $\expect{\indc{i\in\tilde \calS_t} \mid \calF_t}=K/M=p$;
(c) applies \eqref{eq:sgd-mean}.
% are mutually independent, and they are independent of $\calF_t$, % \nbr{This crucial assumption is not mentioned in the main text anymore} 
% so that
% $\expect{\indc{i\in\tilde \calS_t} \mid \calF_t}=K/M=p$ and \eqref{eq:sgd-mean}.
% $$\expect{\nabla \ell_{i,t}(\theta_t) \mid \calF_t}=\frac{1}{n_i}\sum_{j=1}^{n_i} \expect{\nabla \ell_i(\theta_t;z_{i,t}^{(j)}) \mid \calF_t}= F_i(\theta_t).
% $$ 

Applying the above analysis in \eqref{eq:decrease-objective}, we obtain the progress in one communication round, which further yields upper bounds of $\Norm{\nabla F(\theta_t)}$ via suitable scheduling of learning rates and a telescoping sum over the iterations.  
%The proof details are provided in 
See details in \prettyref{app:pf-ub-nonconvex}.

% \paragraph{Extension to FedProx and strongly convex functions. }
% In the proof for FedProx update rule, we need to analyze a different type of objective inconsistency as stated in \prettyref{lmm:fedprox-diff} under the additional lower bound of $\lambda_{\min}(\nabla^2 \ell_{i,t}(\theta))$.  
% For $\mu$-strongly convex function $F$, we are able to track of the progress of the parameters 
% \[
% \Norm{\theta_{t+1}-\theta^*}^2 
% = \Norm{\theta_{t}-\theta^*}^2  + 2 \iprod{\theta_{t+1}-\theta_t}{\theta_t-\theta^*}+\norm{\theta_{t+1}-\theta_t}^2.
% \]
% The previous analysis yields the deviation of $\theta_{t+1}-\theta_t$ from the direction of $-\nabla F(\theta_t)$.
% The remaining analysis is similar using the standard fact that $\iprod{\theta_t-\theta^*}{\nabla F(\theta_t)}\ge \mu \norm{\theta_t -\theta^* }^2$.
% The proof details are deferred to \prettyref{app:pf-strongly-convex}.

\section{Minimax lower bounds}
\label{sec: challenges and limits}
 In this section, we prove a minimax lower bound on the estimation error rates. 
\begin{theorem}
\label{thm:nonconvex-lb-dropout-G}
Given any algorithm (including randomized and non-FL algorithm) and any time horizon $T$ (including $T=\infty$), there always exists a choice of $M$ functions $\{\ell_{i,t}(\theta): i \in [M], t \in [T]\}$ for which 
 \begin{itemize}
 \item 
%i)
each $\ell_{i,t}$ is $L$-smooth and $\mu$-strongly convex;
%ii) 
\item 
$F_i(\theta) \triangleq \expect{\ell_{i,t}(\theta)}$ satisfies the $(B,G)$ condition;
%iii) 
\item 
$\var(\nabla \ell_{i,t}(\theta)) \le \sigma^2_i$,
\end{itemize}
and a choice of $\calS_t \subset [M]$ for $t \in [T]$ with $|\calS_t| \ge (1-\epsilon) M$ %that depends only on $\{\ell_{i,t}(\theta): i \in [M] \}$ 
such that the output of the algorithm $\hat{\theta}$ given access to 
$\{\ell_{i,t}(\theta): i \in \calS_t, t \in [T]\}$ has an error at least
% \begin{itemize}
%  \item $F$ is non-convex: 
\begin{align}
\expect{\|\nabla F(\hat{\theta})\|^2} \ge \frac{\epsilon}{8(1-\epsilon)} \left( G^2 + \sigma^2\right),  \label{eq:lower_bound_0}  
\end{align}
when $F$ is non-convex;
% \item $F$ is $\mu$-strongly convex: 
\begin{align} \expect{\| \hat{\theta}-\theta^*\|_2^2}
\ge \frac{\epsilon}{8(1-\epsilon)\mu^2} \left( G^2 + \sigma^2\right) ,
\label{eq:lower_bound}
\end{align}
when $F$ is $\mu$-strongly convex.
% \end{itemize}
\end{theorem}

% \[
% \frac{\epsilon}{8(1-\epsilon)} \pth{G^2+\sigma^2} \le \inf_{\hat\theta}\sup_{\calA}\sup_{F_1, \cdots, F_M}
% \expect{\|\nabla F(\hat{\theta})\|^2}
% \le 4 \epsilon\pth{G^2+\sigma^2};
% \]
% \item $F$ is $\mu$-strongly convex:
% \[
% \frac{\epsilon}{8(1-\epsilon)\mu^2} \pth{G^2+\sigma^2} \le \inf_{\hat\theta}\sup_{\calA}\sup_{F_1, \cdots, F_M}
% \expect{\| \hat{\theta}-\theta^*\|_2^2}
% \le  \frac{4 \epsilon}{\mu} \pth{G^2+\sigma^2};
% \]

%\noindent {\bf Minimax optimality.} 
% Theorem \ref{thm:nonconvex-lb-dropout-G} says that any algorithm %even with access to the entire sequence of local functions subject to adversarial dropouts 
% suffers an estimation error that is at least on the order of $\epsilon \left( G^2 + \sigma^2 \right)$. 
We have shown in Theorems \ref{thm:nonconvex} and \ref{thm:strongly_convex} that the lower bounds in~\prettyref{eq:lower_bound_0} and~\prettyref{eq:lower_bound} can be attained (up to a constant factor) by simple variants of the standard FedAvg and FedProx with only access to noisy gradients and even when the adversary can adaptively choose $\calS_t$ based on all the history information. 
%Combining this result with~\prettyref{thm:nonconvex-lb-dropout-G}
%The two results together 
%implies that $\epsilon \left( G^2 + \sigma^2 \right)$ is minimax optimal. % rate. 

%\vskip \baselineskip
%\noindent{\bf Proof sketch and comparison of lower bounds.}
%We first briefly sketch the idea behind the proof of~\prettyref{eq:lower_bound}. 

%\vskip \baselineskip

%\noindent{\bf Proof sketch and comparison of lower bounds.}
\subsection{Proof sketch and comparison of lower bounds}
To deduce that the lower bounds are at least on the order of $\epsilon G^2$, it suffices to construct time-invariant $\calS_t$. 
We construct two  instances: one homogeneous instance where  $\ell_{i,t}=f$ for all $i, t$; and the other heterogeneous instance where $\ell_{i,t}=f$ for $i \in S$ and $\ell_{i,t}=g$ for $i \notin S$ for a fixed subset $S\subset[M]$ with $|S|=(1-\epsilon)M$. The functions $f$ and $g$ are properly chosen to be sufficiently distinct while satisfying the $(B,G)$-condition.
Importantly, if the adversary chooses $\calS_t=S$ for all $t$, then any algorithm with access to $\{\ell_{i,t}: i \in \calS_t\}$  cannot distinguish the two instances and hence cannot simultaneously optimize both instances.  %Note that here it suffices to consider a static choice of $\calS_t$. 
% A very similar construction of the two instances has been used in~\cite[Theorem III]{karimireddybyzantine22} to prove that the same lower bound $\epsilon G^2$ holds for a static Byzantine attack where the adversary can choose a pre-selected $\calS$ and corrupt $\{\ell_{i,t}, i \notin \calS\}$ by arbitrary errors. 

In contrast, to prove the lower bounds on the order of $\epsilon \sigma^2$, it is crucial to have %allow
$\calS_t$ %to 
be time-varying and chosen based on the realization of $\{\ell_{i,t}, i \in [M]\}$. %\nb{memoryless} 
In particular, we again construct two instances. The first instance is exactly the same as before. The second instance replaces $S$ 
% is also the same, except that $S$ is replaced 
by $S_t$ chosen uniformly and independently from all subsets of $[M]$ with size $(1-\epsilon) M$. 
%Note that under the second instance $\ell_{i,t}$ is equal to $f$ with probability $1-\epsilon$ and $g$ with probability $\epsilon.$ 
This time the functions $f$ and $g$ are chosen to be sufficiently distinct while the variance of $\nabla \ell_{i,t}$ is kept at most $\sigma^2$. 
%Now, if the adversary chooses $\calS_t=S_t$, then any algorithm with access to $\{\ell_{i,t}: i \in \calS_t\}$  cannot distinguish the two instances and hence cannot simultaneously optimize both instances. 

Note that the previous work~\cite{karimireddybyzantine22} assumes the adversary is static and chooses a prefixed $\calS$ to inject errors; thus the lower bound~\cite[Theorem III]{karimireddybyzantine22} does not contain the $\epsilon \sigma^2$ term as we do. For this reason, when $G=0$ they show that an FL algorithm with a robust aggregator of noisy gradients can approach the zero optimization error as $T \to \infty$. However, this is fundamentally impossible in our setting with non-static adversarial dropouts.

\begin{remark}[\bf The impact of dissimilarity parameter $B$]
\label{remark: impacts of B}
The lower bounds in \prettyref{thm:nonconvex-lb-dropout-G} do not depend on the other dissimilarity parameter $B.$ From the analysis of our algorithm, the parameter $B$ in \prettyref{ass:BG} instead influences the convergence speed and the requirement on the learning rate; the eventual precision is unaffected. 
\end{remark}

\begin{remark}[\bf Convergence rate in $T$]
\label{remark: convergence rate in T}
%While our lower bound \prettyref{thm:nonconvex-lb-dropout-G} holds for any $T$, it 
Our lower bounds do not capture the dependency on $T.$
It is known in the literature that
even in the centralized homogeneous setting, 
any algorithm
with access to $T$ queries of noisy gradients of $F$ with variance $\sigma^2$ has to suffer an estimation error $\expect{\|\nabla F(\hat{\theta})\|_2^2}\ge \Omega(\sigma/\sqrt{T}) $
for non-convex $F$ and $\expect{\|\hat\theta - \theta^*\|^2} \ge \Omega(\mu^2\sigma^2/T)$
for $\mu$-strongly convex $F$ in the worst case
(see e.g., \cite[Theorem 3]{arjevani2022lower} and~\cite[Theorem E.1]{gu2021fast}). 
\end{remark}

\section{Numerical Experiments}
\label{sec: experiment}
In this section, we use numerical experiments to corroborate our theories and analysis on both real-world and synthetic datasets. 
We evaluate the top-1 accuracy of the proposed variants of FedAvg and FedProx on real-world datasets CIFAR-10 \cite{krizhevsky2009learning}, Shakespeare \cite{mcmahan2017communication}, and synthetic dataset following \cite{shamir2014communication,li2020federated}.
% 
%All the results are averaged over \nbr{at least} two different random seeds.      nm       hnmnh m\ls{Mention this in the corresponding experiments only, not here.} 

\noindent {\bf Baselines.}
Each of the evaluations consists of two parts: 
\begin{enumerate}[(1)]
 \item Comparisons with the baselines FedAvg \cite{mcmahan2017communication}, FedProx \cite{li2020federated}, and MIFA \cite{gu2021fast}. 
 The baseline MIFA is chosen because it 
 % has high visibility and is observed to have superior numerical performance
 is designed
 against general client unavailability that does not have benign random patterns. 
The key idea of MIFA is that for unavailable clients, the parameter server uses the memorized latest updates from those clients for aggregation. 
For clients running FedProx, we use momentum SGD to solve the local program. 
For ease of presentation, henceforth we refer to FedAvg, MIFA, and our FedAvg variant as FedAvg-type algorithms because that the clients under those algorithms share the same form of local computations. 
\item Comparisons with the Byzantine-resilient algorithms centered clipping (cclip) \cite{karimireddy2021learning}, geometric median (GM) \cite{chen2017distributed}, and their bucketing variants \cite{karimireddybyzantine22}. 
Given $\{x_1, \cdots, x_M\}\subseteq \reals^d$,  cclip  \cite{karimireddy2021learning} and GM  \cite{chen2017distributed} aggregate those $M$ points as follows: 
\begin{itemize}
\item cclip: Given the clipping radius $\tau>0$ and iteration budget $L$, % and an initial guess $v_0$,  
$\text{agg}_{cclip, L}(x_1, \cdots, x_M)$ is iteratively obtained as %aggregates the $M$ points via pre-specified $L$ rounds as
\begin{align*}
  v_{{\ell}+1} = v_{\ell} + \frac{1}{M}\sum_{i=1}^M (x_i-v_{\ell}) \min\sth{1, ~ \frac{\tau}{\norm{x_i-v_{\ell}}}} ~~~ \text{for }\ell=0, \cdots, L-1.
\end{align*}
In our experiments, we choose $L=3$, the same as in \cite{karimireddy2021learning}, and an initial guess $v_0 = \bm{0}$. 
%\nbr{specify $v_0$?}
\item GM: $\text{agg}_{GM}(x_1, \cdots, x_M) \triangleq \arg\min_{v\in \reals^d} \sum_{i=1}^M \norm{v-x_i}$. 
%
% \mx{
% I also think the above should be 
% $\text{agg}_{GM}(x_1, \cdots, x_M) = \arg\min_{v\in \reals^d} \sum_{i=1}^M \norm{v-x_i}$.
% \eg in Eq.~6 in \cite{chen2017distributed}
% }
%

In our experiments, we use the smoothed Weiszfeld algorithm (Algorithm 2 in \cite{9721118}) with iteration budget $R=8$, which finds an approximate minimizer of $\min_{v\in \reals^d} \sum_{i=1}^M \norm{v-x_i}$.  
%$\text{agg}_{GM}(x_1, \cdots, x_M)$.

% 
%\nbr{I don't think this is true...}
%
\item Bucketing is a technique that aims to reduce the impact of data heterogeneity. 
It randomly partitions the $M$ points into $\lceil {M}/\pth{bs}\rceil$ buckets for some tuning parameter $bs$ that determines the bucket size. Then, the data points in each bucket are averaged to construct bucket means $\hat{x}_1, \cdots, \hat{x}_{\lceil {M}/\pth{bs}\rceil}$, which are fed into aggregators such as cclip and GM.
\end{itemize}
 
Following \cite[Section 5]{karimireddybyzantine22}, we use momentum for the Byzantine-resilient algorithms to reduce the variance of the stochastic gradients. That is, the $\{x_1, \cdots, x_M\}$ are the local momentum of the cumulative stochastic gradients.  
To account for partial client participation, for the inactive clients, the parameter server reuses their local momentum
% \nbr{one-step ($s=1$) momentum} 
 in the last active rounds, analogously to MIFA, for a fair comparison. 
Similar to \cite{karimireddybyzantine22}, we set the bucket size $bs=2$. %  %two buckets for the bucketing variants.        
\end{enumerate}

\noindent{\bf Setup and hyperparameters.}
For CIFAR-10 and synthetic datasets, we let the datasets be distributed over $M=100$ clients.
In each round, $K=10$ clients are sampled uniformly at random without replacement to build $\tilde{\calS}_t.$
The data partition and client population are a bit different for Shakespeare dataset; see Section \ref{subsec: NLP} for details. 

For our algorithms, the amplification factor $\beta$ is set as $M / K = 10$ throughout.
The learning rates and the proximal coefficient are tuned from grid searches, where 
the initial learning rates $\eta_0 \in \sth{10^{-4},\ldots,10^{1}},$
but the proximal coefficient $\mu_t\triangleq (1/\eta_t) \in \sth{10^{-3},\ldots,1}$
in Eq.~\eqref{eq:FedProx_proximal}. In our experiments, we choose $\mu_t=\mu_0=(1/{\eta_0})$. 
%$\nbr{\mu_0\triangleq \frac{1}{\eta_t}}\in \sth{10^{-3},\ldots,1}.$
Notably, we run SGD with momentum as Eq.~\eqref{eq: momentum SGD} on each client $i$ to solve the local program of FedProx and Byzantine-resilient algorithms: 
\begin{align}
\label{eq: momentum SGD}
\begin{cases}
m_{i,t}^{\tau+1} = \beta_0 m_{i,t}^{\tau} + (1 - \beta_0) \nabla \ell_{i,t} (\theta_{i,t}^{\tau}) \\
\theta_{i,t}^{\tau+1} =  \theta_{i,t}^{\tau} - \alpha_0 m_{i,t}^{\tau},
\end{cases}
\end{align}
where $\theta_{i,t}^{\tau}$ denotes the updated model after $\tau$ steps local computations in round $t$ on client $i$ with $\theta_{i,t} = \theta_t$ for all $i\in[M]$ and $t \ge 0$.
A common practice $\beta_0 = 0.9$ is adopted as the momentum coefficient 
and fixes a constant local learning rate $\alpha_0$ tuned from the same grid as $\eta_0$.
For a constant learning rate, we consider $\eta_t = \eta_0$ for all $t \ge 0$.
For a decaying learning rate, let $\eta_t = \eta_0 / \sqrt{t+1}$. 
% The momentum SGD on the clients runs $s$ steps with a constant local step size $\alpha$ and a momentum coefficient $\beta_0 = 0.9$.
% \nbr{reuse notations like $\theta_{i,t}^\tau$ as FedAvg?}

The hyper-parameter setups are summarized in~\prettyref{tab:setup}.
Additional details of the experiments and hardware environments are deferred to~\prettyref{app: exp}.

\begin{table}[!htb]
\centering
\resizebox{\textwidth}{!}{
\begin{tabular}{cccccccc}
\toprule
%%%%%%%%%%%%%%%%%%%%%%%%%%%%%%%%%%%%%%%%%%%
Dataset & 
Algorithms & 
Learning Rates & 
Initial Rates & 
Local Solver &  
Local Steps & 
Batch Size \\
\midrule
%%%%%%%%%%%%% CIFAR-10 %%%%%%%%%%%%%%%%%
\multirow{3}{*}{CIFAR-10} & 
FedAvg-type & 
 $\eta_0/\sqrt{t+1}$ & 
$\eta_0=0.1$ &  
SGD & 
$s=25$ &
\multirow{3}{*}{100}  \\
 &  
FedProx, FedProx variant & 
 $\alpha_0$ & 
$\alpha_0 = 0.03$, $\mu_0=0.1$ &  
SGD with momentum &  
$s=10$ & \\
& 
GM, cclip, and their bucketing variants &
 $\alpha_0$ & 
$\alpha_0 = 0.001$& 
SGD with momentum &
$s=1$ & \\
\midrule
%%%%%%%%%%%%% Shakespeare %%%%%%%%%%%%%%%%%
\multirow{2}{*}{Shakespeare} & FedAvg-type & 
 $\eta_0/\sqrt{t+1}$ & 
$\eta_0=1$ &  
SGD & 
$s=200$ &
\multirow{2}{*}{500}  \\
 % &  
% FedProx, FedProx variant & 
% - & 
% - &  
% -
% &  
% - & \\
& 
GM, cclip, and their bucketing variants &
 $\alpha_0$ & 
$\alpha_0 = 0.8$ & 
SGD with momentum &
$s=1$ & \\  
\midrule
%%%%%%%%%%%%% Synthetic %%%%%%%%%%%%%%%%%
\multirow{3}{*}{Synthetic} & FedAvg-type & 
 $\eta_0$ & 
$\eta_0=0.01$ &  
SGD & 
$s=25$ &
\multirow{3}{*}{Full batch}  \\
&  
FedProx, FedProx variant & 
 $\alpha_0$ & 
$\alpha_0 = 0.01$, $\mu_0=1$ &  
SGD with momentum &  
$s=10$ & \\
& 
GM, cclip, and their bucketing variants &
 $\alpha_0$ & 
$\alpha_0 = 0.01$ & 
SGD with momentum &
$s=1$ & \\
%%%%%%%%%%%%%%%%%%%%%%%%%%%%%%%%%%%%%%%%%%%
\bottomrule
\end{tabular}
}
\caption{Hyper-parameter setups}
\label{tab:setup}
\end{table}

\subsection{Adversarial client unavailability scheme} 
\label{subsec: adversarial client unavailability}

% \nbr{comment:: give motivation, context, and explain what you do...}
% {\red 
% \begin{itemize}
% \item First, explain why we come up with this scheme, something like "the adversary wants to adversarially drop out the clients with largest gradient change, etc. Since we do not know future gradients, the adversary use the simulated gradients.
% \item Explain how to simulate the gradients.
% \item Explain how we use simulated gradients to select the clients to drop out
% \end{itemize}} 
In this subsection, we describe our client unavailability scheme for the system adversary. 
Additional adversarial client unavailability schemes can be found in Appendix \ref{app: additional adversary}. 

Our adversarial client unavailability scheme entails the selection of specific clients from ${\tilde \calS}_t$ to dropout. 
%be non-responsive. 
%and rendering them non-responsive.
The high-level idea is to have the adversary drop the most valuable clients.
To assess the value of a client, 
%{\red we consider the difference between the gradients near the beginning of the training and near the round when the training is about to converge. }
we consider the difference of gradients between a pair of rounds $T_1$, $T_2$, which can be further tuned.  
%Any pair of round indices with sufficient separation will also work. 
In fact, our scheme can be readily extended to multiple pairs of such round indices.
%the end of the training process. 
Specifically, the adversary %'s strategy is to 
excludes the clients whose gradients change the most significantly between two chosen communication rounds. 
A key obstacle in exactly realizing this scheme is that the system adversary has no future information which is required to compute the aforementioned gradient changes.   
To get around this, we use auxiliary experiments and leverage the knowledge of the clients' local data distribution to obtain an approximation of the gradient changes. %, detailed as follows.  
% e adopt the following approximation via auxiliary experiments leveraging the knowledge of the clients' data distribution. 

% {\blue The high-level idea is to have the adversary drop the most useful clients.
% One way to capture the usefulness of a client is via the norm of its local gradients,}
% {\blue and the corresponding adversary} scheme aims to exclude the clients whose gradients have significant changes {\blue throughout the entire training}.
% {\blue A key obstacle in exactly realizing this scheme is that} 
% the system adversary has no {\blue future information}. % access to the future sample generated by the clients. 
% {\blue Hence, we adopt the following approximation. }
% \ls{Pengkun and Jiaming, I remembered I wrote down something like auxiliary experiment for each real experiments and some follow-up arguments. See our arXiv version, where the notion "shadow" experiments are used. It seems that those texts were deleted. Hence, I will leave it up to you guys to finish this part. } 
% \nbr{
%   I create a tag in overleaf history labeled "2023-10-06" for the version before the revision on the experiment sections. 
%   I also copied the contents to Draft/numerical\_backup.tex.
% }

Next we provide details of our approximation.  
%To identify a set of valuable clients without accessing the future gradients, 
Before running regular experiments, the adversary conducts independent runs of FedAvg and cclip with full client participation. 
The local datasets in these auxiliary experiments are independent copies of those in regular experiments, 
which both follow the same local distribution and thus
can be achieved through the use of distinct random seeds. 
%In these auxiliary experiments, local datasets are generated from clients' data distributions, achieved through the use of distinct random seeds.
%For each communication round $t$ and each client $i$, 
We denote the auxiliary local objective function as $\tilde\ell_{i,t}$, the auxiliary local model obtained from FedAvg after $\tau$ steps as $\tilde\theta_{i,t}^\tau$, and the auxiliary local momentum vector from cclip as $\tilde{m}_{i,t}$ for $i\in [M]$ and $t\ge 0$. 
For two carefully chosen rounds $T_1$ and $T_2$, we compute the gradient norm $g_{i,t}\triangleq \|\sum_{\tau=0}^{s-1}\nabla \tilde\ell_{i,t} (\tilde\theta_{i,t}^\tau)\|_2$, and the momentum norm $ g_{i,t}'\triangleq \| \tilde{m}_{i,t} \|_2$ for $t\in \{T_1,T_2\}$. 
% {\blue We adopt the heuristic of choosing $T_1$ near the beginning of the training and $T_2$ near the round when the training is about to converge.} 
%During the execution of these auxiliary experiments, the adversary records the status of clients at two carefully chosen rounds, denoted as $T_1$ and $T_2$. 
%{\blue We adopt the heuristic of choosing $T_1$ near the beginning of the training and $T_2$ near the round when the training is about to converge.}  
%
% \mx{Replaced $m_{i,t}$ by $\bm{m}_{i,t}$ to avoid conflicts with Eq.~\eqref{eq: momentum SGD}.}
%
We construct set $\calC$, termed as {\em candidate set of valuable clients}, as follows: 
%The candidate set of valuable clients, denoted as $\calC$, is constructed based on the status collected during the auxiliary experiments. 
\begin{enumerate}
\item $\calC_1 \subseteq [M]$ is a set of $K_1$ clients such that $|g_{i,T_2}-g_{i,T_1}| \ge |g_{i^{\prime},T_2}-g_{i^{\prime},T_1}|$ for $i\in \calC_1$ and 
%{\red $i^{\prime}\notin \calC_1$.}
$i'\in [M]\setminus \calC_1$. 
%with  a cardinality of $K_1$, consisting of those with the largest $|g_{i,T_2}-g_{i,T_1}|$.
\item $\calC_2\subseteq [M]\setminus \calC_1$ is a set of $K_2$ clients such that 
 $|g_{i,T_2}'-g_{i,T_1}'|\ge  |g_{i^{\prime},T_2}'-g_{i^{\prime},T_1}'|$ for 
 %{\red $i\in \calC_1$ and $i^{\prime}\notin \calC_1$. }
 $i\in \calC_2$ and $i'\in [M]\setminus\pth{\calC_1 \cup \calC_2}$. 
\item $\calC = \calC_1 \cup \calC_2$.
\end{enumerate}
In our implementation, we set $T_1=5$, $T_2=150$, $K_1=25$, and $K_2 = 10$. 

% Specifically, we define $\calC_1 \subseteq [M]$ as a subset of indices with a cardinality of $K_1$, consisting of those with the largest $|g_{i,T_2}-g_{i,T_1}|$.
% Similarly, we define $\calC_2\subseteq [M]\setminus \calC_1$ with a cardinality of $K_2$, comprising those with the largest $|g_{i,T_2}'-g_{i,T_1}'|$.
% The candidate set is then formed as $\calC = \calC_1 \cup \calC_2$.
% In our implementation, we set $T_1=5$, $T_2=150$, $K_1=25$, and $K_2 = 10$.

During the execution of the regular experiments, the adversary chooses $\calS_t$ by rendering clients in $\calC\cap \tilde{\calS}_t$ non-responsive subject to the constraint in~\prettyref{ass:adversarial}. 
In particular, in each communication round $t$, let $(i_{t,1},\ldots,i_{t,K})$ denote the client indices in $\tilde{\calS}_t$ after random permutation. 
Let $\calS_t$ be initialized as $\tilde{\calS}_t$, and we iteratively select the non-responsive clients.
For $k=1, \ldots, K$, if $i_{t,k}\in \calC$ and $\sum_{i\in (\tilde{\calS}_t \setminus {\calS_t})\cup\{i_{t,k}\}} n_i \le \epsilon K N / M,$ 
then set $\calS_t \gets \calS_t \setminus \{i_{t,k}\}$.  % \nbr{what does this mean.......fix notations???}
The procedure is terminated if $\abth{{\calS_t}}= 1$ to ensure that at least one client is responsive.

Clearly, the above scheme does not use future information from regular experiments.

\subsection{Experiments on CIFAR-10}
\label{sec: exp cifar10}
\begin{figure}[htb]
     \centering
    % \begin{subfigure}[b]{0.29\textwidth}
    % \includegraphics[width=\textwidth]{Numerical/rebuttal/rfig/dirichlet/dir30clients.pdf}
    % \end{subfigure}
    % \begin{subfigure}[b]{0.32\textwidth}
    \includegraphics[width=0.5\textwidth]{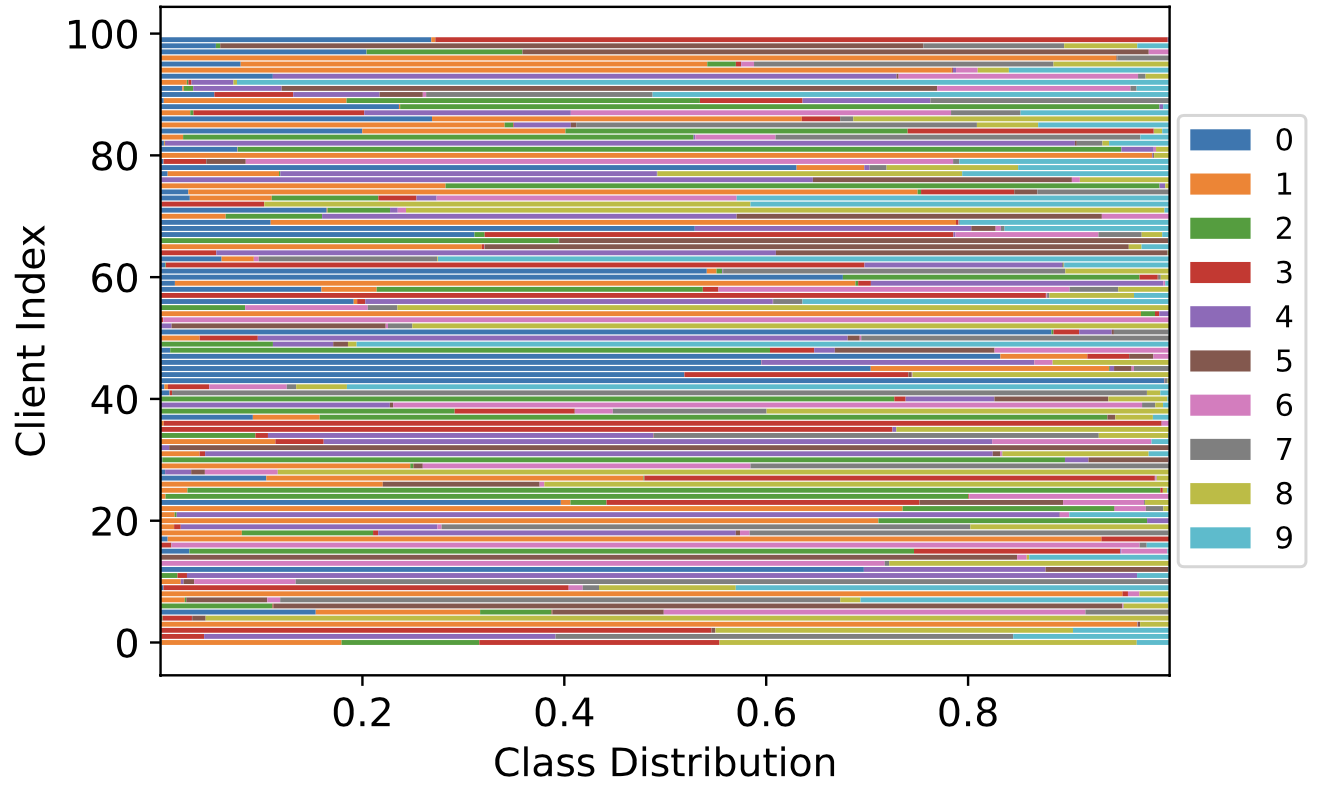}
    % \end{subfigure}
     \vskip -0.45\baselineskip 
    \caption{Populations generated from Dirichlet distribution $\pth{\alpha=0.1}$ with different number of clients. Each row corresponds to the empirical distribution of local data in terms of classes. The colors correspond to data with different class labels. }
    \label{fig: dirichlet distribution}
\end{figure}

Following \cite{49350},
the CIFAR-10 dataset is partitioned to build each client's local datasets according to the Dirichlet allocation with parameter $\alpha$; the smaller $\alpha$, the more non-IID of the local data. Each client keeps $500$ samples. 
%\underline{The local data volume is $n_i\equiv 100$ for $i\in [M]$ per round. }
In each round, client $i$ draws a batch of $n_i\equiv 100$ samples from its local dataset.  %$\calD_i$ for $i\in [M]$ per round.
% {\blue
% Client $i$ draws $n_i\equiv 100$ samples from its local data partition $\calD_i$ for $i\in [M]$ per round.
% The volume of a local dataset partition is $500$ at client $i \in [m]$.
% }
% \ls{XM to do: are these 100 data from the common data pool or from local datasets? Clarify this. }
% \mx{I am not sure if I can use local distribution $\calD_i$ to denote local dataset partition.}
% \nbr{each client holds 500 data?} 
% \nbr{Add details on how the dataset is partitioned. }
We choose $\alpha =0.1,$ which creates highly non-IID local datasets due to label skewness, where the volume of each label on the clients is shown in~\prettyref{fig: dirichlet distribution}.  
Dirichlet distribution is commonly adopted for characterizing non-IID distributions (see, e.g., \cite{wang2020tackling,cho2022towards,Wang2020Federated}). 
Notably, $\alpha=0.1$ was also chosen in \cite{wang2020tackling}, $\alpha=0.5$ was used in \cite{Wang2020Federated}, and $\alpha\in \{2, 0.3\}$ were considered in \cite{cho2022towards}. 
In the plot, we can readily see that the categorical distributions among clients are drastically different. For example, in Fig.~\ref{fig: dirichlet distribution}, client \# 1 has five classes of data, whereas client \# 100 has three classes. Moreover, the fractions of the classes vary significantly across clients.

%\nbr{Is it counted from local dataset?} \nbr{remove subfigure caption.}
We present the results with $\epsilon = 0.8$ in this subsection. 
The results with other choices of $\epsilon$ are similar and are reported in~\prettyref{app: exp}. 
We use LeNet-5 \cite{lecun1998gradient} with cross-entropy loss as the network model.
We plot the training and test performances every 15 communication rounds. 
% The full plot and other details are deferred to~\prettyref{app: exp}.

\begin{figure}[!htb]
\centering
\begin{subfigure}[b]{\textwidth}
    \includegraphics[width=\textwidth]{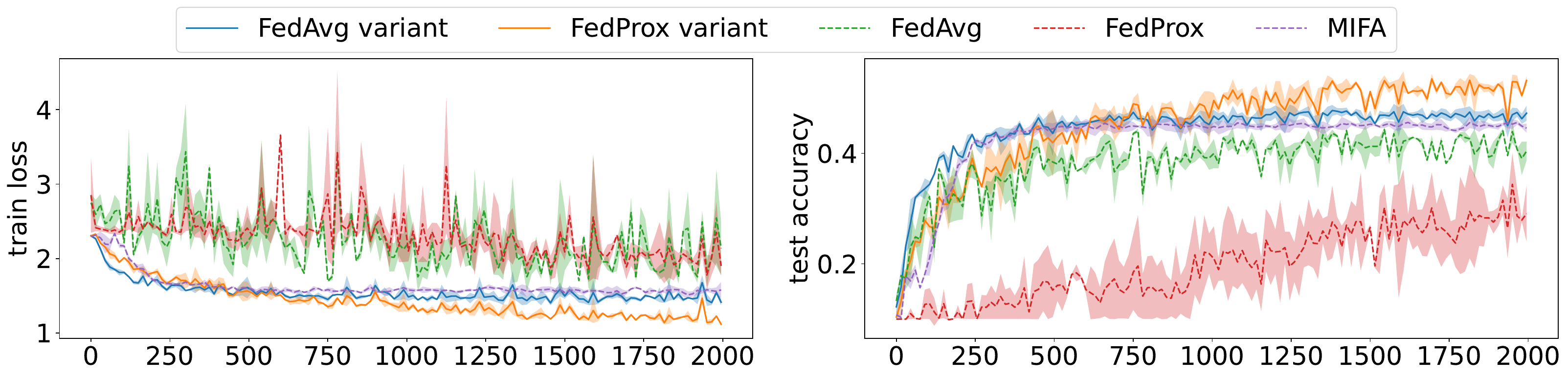}
    \caption{Baseline comparisons.}
    \label{fig:cifar10_0.1_baseline}
\end{subfigure}
\centering
\begin{subfigure}[b]{\textwidth}
    \includegraphics[width=\textwidth]{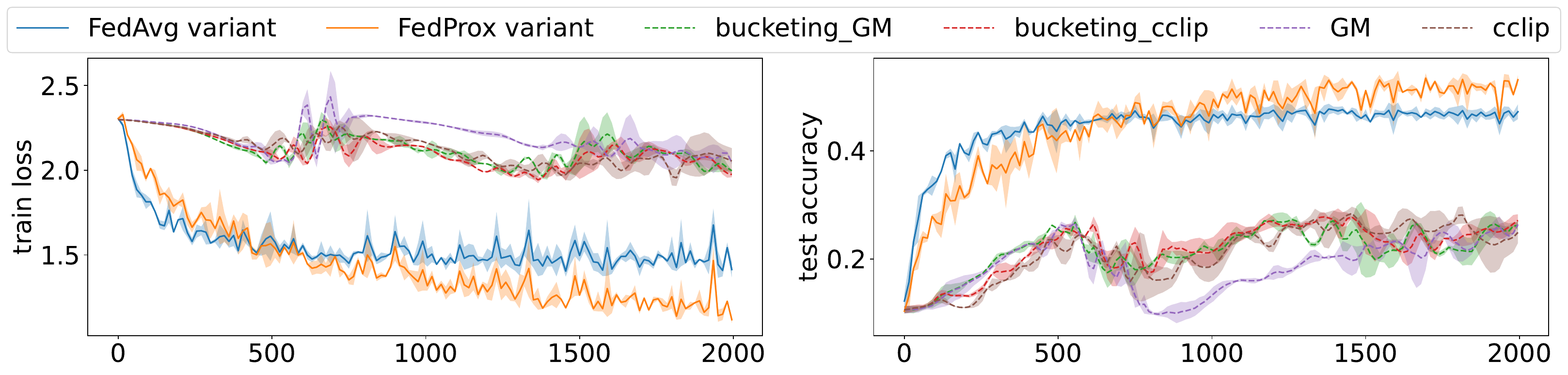}
    \caption{Byzantine comparisons.}
    \label{fig:cifar10_0.1_byzantine}
\end{subfigure}
\caption{CIFAR-10 results 
with Dirichlet parameter $\alpha=0.1$ and dropout fraction $\epsilon =0.8$ 
on adversarial client unavailability scheme in Section \ref{subsec: adversarial client unavailability}.
}
\label{fig:cifar10_0.1}
\end{figure}

We observe from~\prettyref{fig:cifar10_0.1} that the proposed variants of FedAvg and FedProx outperform all the other baselines. %algorithms. 
Compared with the FedAvg and FedProx, our variants progress smoother, 
and the improvements are quite prominent.
Notably, MIFA acts better than FedAvg but falls behind our variants.
% The second row of Fig.\,\ref{fig:cifar10_0.1_baseline} illustrates that the adversarial client unavailability scheme in  Section \ref{subsec: adversarial client unavailability}
% obeys Assumption \ref{ass:adversarial} and creates a highly adversarial environment,
% where only $2$ clients participate in the training per round.
% {\blue need to revise the words as the results in Appendix show comparable performance.} 
On the other hand, the Byzantine-resilient algorithms lag behind the proposed algorithms significantly.  
%We refer the interested readers to Appendix \ref{app: Byzantine literature} for 
Detailed discussions on existing Byzantine-resilient algorithms are presented in Section \ref{sec: related work}.

\subsection{Natural language processing (NLP) task: Shakespeare next-character prediction.}
\label{subsec: NLP}
\begin{figure}[!htb]
\centering
\begin{subfigure}[b]{0.42\textwidth}
\includegraphics[width=\textwidth]{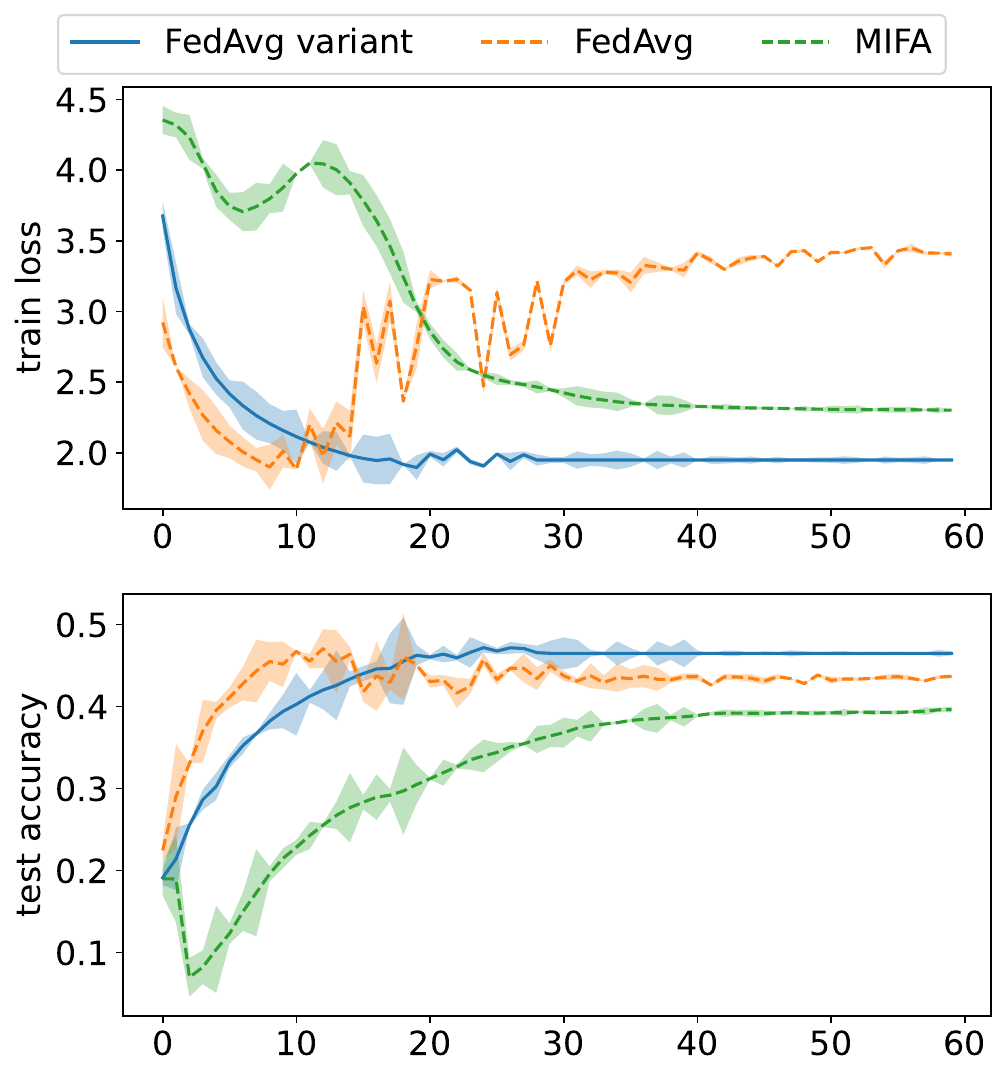}
\caption{Baseline comparisons.}
\label{fig: NLP baselines}
\end{subfigure}
\begin{subfigure}[b]{0.48\textwidth}
\includegraphics[width=\textwidth]{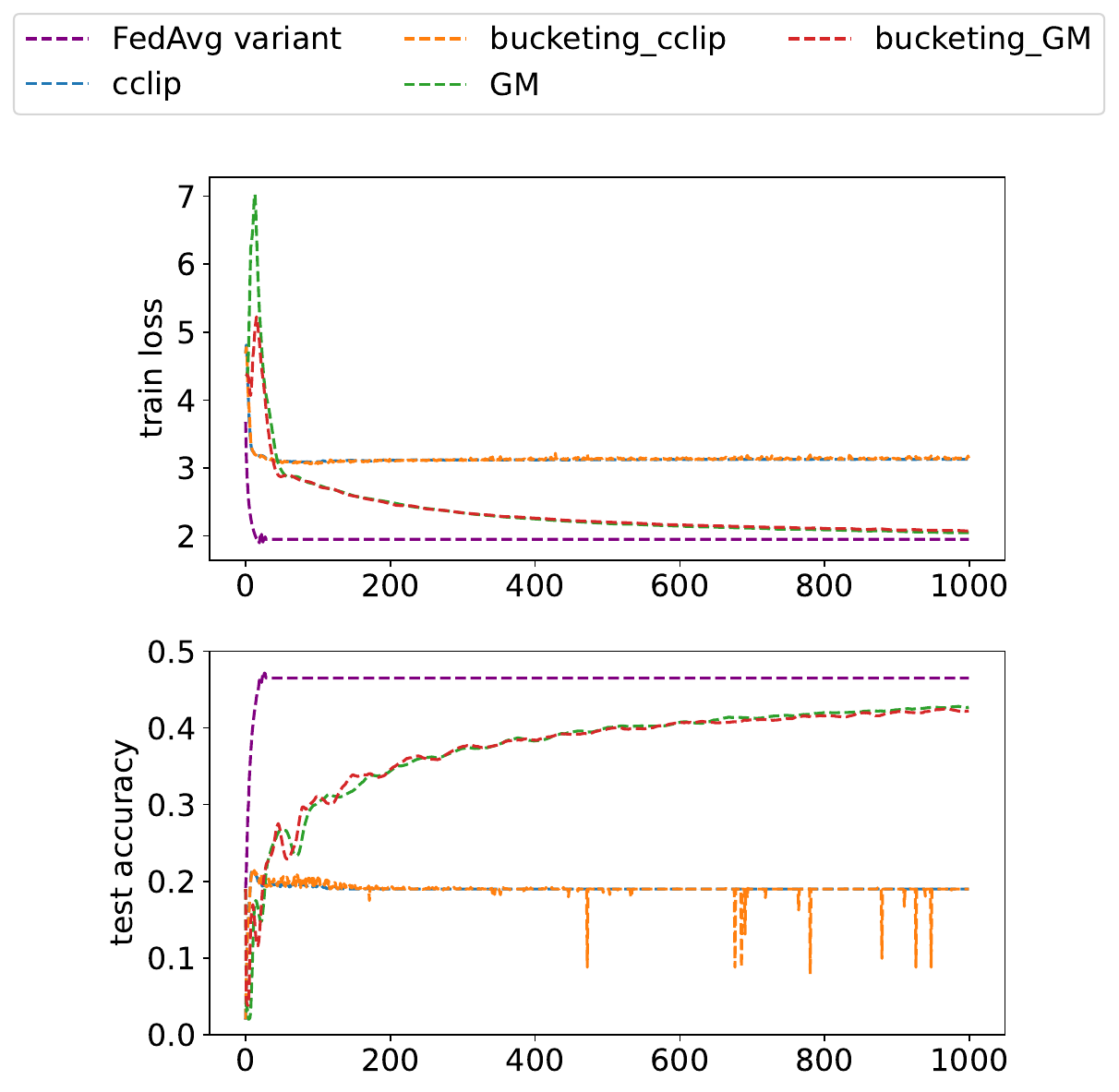}
\caption{Byzantine comparisons.}
\label{fig: NLP Byzantine}
\end{subfigure}
% \vskip -1pt
 \vskip -0.5\baselineskip  
\caption{
Natural language processing task with dropout fraction $\epsilon=0.7$ on a different adversarial client unavailability scheme,
where the adversary 
inspects each client's local gradient improvement
and removes clients of the greatest improvements subject to Assumption \ref{ass:adversarial}. 
Details can be found in Section \ref{subsec: NLP}.
}
\label{fig: NLP task}
\end{figure}

%\ls{Check and explain experiments on FedAvg -- why a straightline. More random seeds, figure types are not consistent with others. }
We also test the performance of FedAvg-type algorithms on Natural language processing (NLP) task. Our results are presented in Fig.\,\ref{fig: NLP task}. 
%Now, we are ready to illustrate our system setup. 
The Shakespeare dataset is built from {\em The Complete Works of William Shakespeare} \cite{mcmahan2017communication}, which contains 4,226,158 data instances. 
LEAF \cite{caldas2018leaf} (a federated learning benchmark) partitioned it into 660 groups. %clients, was adapted from the LEAF \cite{caldas2018leaf} (a federated learning benchmark). 
We sample around 18\% from the 660 groups to obtain the local datasets for $M = 125$ clients.  
The network model is an LSTM network with two layers, each of which has 256 neurons; this model takes in each character as an 8-dimensional embedding. In each round, $K = 20$ clients are sampled to form $\tilde{\calS}_t$, and each sampled client draws a batch of $n_i\equiv 500$ samples. % to participate in training. 
We choose the dropout threshold $\epsilon = 0.7$.

In addition to changing the learning task from computer vision to NLP, we also test a different dropout scheme:
\begin{itemize}
\item The adversary calculate the $l_2$ norms of the local gradient improvement $\|w_i\sum_{\tau = 0}^{s-1} \nabla \ell_{i,\tau}\|_2$ at the client $i \in \tilde{\calS}_t$ and sort the norms in descending order. 
\item The adversary inspects clients' data volume $n_i$ for $i\in \tilde{\calS}_t$ in the sorted order. If $n_i > \epsilon (KN) / M$, client $i$ will be admitted to the set ${\calS}_t$. Otherwise, the client's application will be denied until $\sum_{i \in \tilde{\calS}_t \setminus \calS_t} n_i \le \epsilon (KN) / M$. 
\end{itemize}

Intuitively, we discard the most "important" local gradient improvements in terms of $l_2$ norm.

In Fig.\,\ref{fig: NLP baselines}, our FedAvg variant progresses the most smoothly during training and obtains the best results.  
One can see that the convergence time in Fig.~\ref{fig: NLP baselines} (around 60 rounds) and Fig.~\ref{fig: NLP Byzantine} (around 1000 rounds) do not match each other. This is because GM and the bucketing version of GM converge very slowly. 
%=presumably attribute this to the fact that Byzantine-resilient algorithms do not perform local computations.
%In order to better showcase the benefits of the proposed variant, a dashed horizontal line in purple is added in Fig.~\ref{fig: NLP Byzantine} to present the final performance of FedAvg variant for reference.
In both Fig.\,\ref{fig: NLP baselines} and Fig.~\ref{fig: NLP Byzantine}, our FedAvg variant stands as the best.

\subsection{Experiments on synthetic datasets}
\label{sec:synthetic}
\begin{figure}[!htb]
\centering
\includegraphics[width=0.7\textwidth]{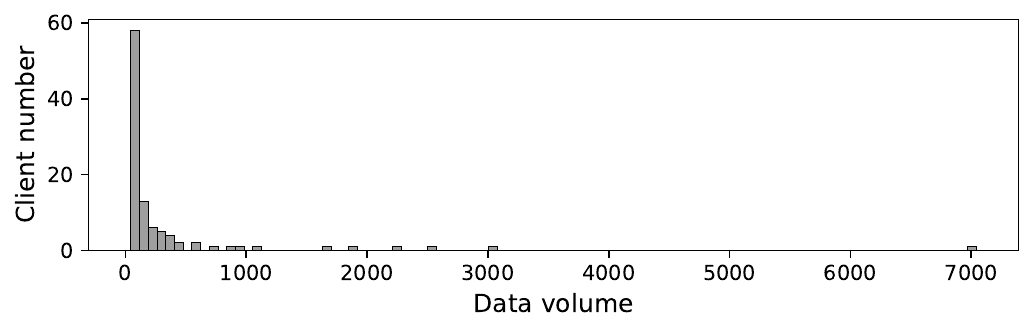}
\caption{Synthetic datasets: clients' local data volume histogram.}
\label{fig:power law}
\end{figure}

% \noindent{\bf Overview:}
We follow the setup of the synthetic experiments in \cite{shamir2014communication,li2020federated}. 
We generate the local dataset $(x_{ij},y_{ij})_{j=1}^{n_i}$ %$(X_i,Y_i)$ 
for each client $i$ according to the model $y_{ij}=\argmax \pth{\text{softmax}\pth{W_i x_{ij} + b_i}},$ where $x_{ij}\in\reals^{60},$ $W_i\in\reals^{10\times 60},$ $b_i\in\reals^{10}.$ 
% \nbr{fix notational typo....}
To generate heterogeneous clients, each element of $W_i$ and $b_i$ is independently drawn from $\calN\pth{u_i,1}$, where $u_i\sim \calN\pth{0,1}$. Moreover, $x_{ij}\sim \calN\pth{v_i,\Sigma},$ where the covariance matrix is diagonal with $\sum_{j,j}=j^{-1.2}.$
Each element of the mean vector $v_i$ is independently drawn from $\calN\pth{B_i,1},$ where $B_i\sim \calN\pth{0,1}.$
In contrast to the experiments on CIFAR-10, here we consider the setup with quantity skewness, where the local data volume $n_i$ follows a power law as shown in~\prettyref{fig:power law}.  
In the experiments, we run multinomial logistic regression with full batch gradient descent and cross-entropy loss.

\begin{figure}[htb]
\centering
\includegraphics[width=\textwidth]{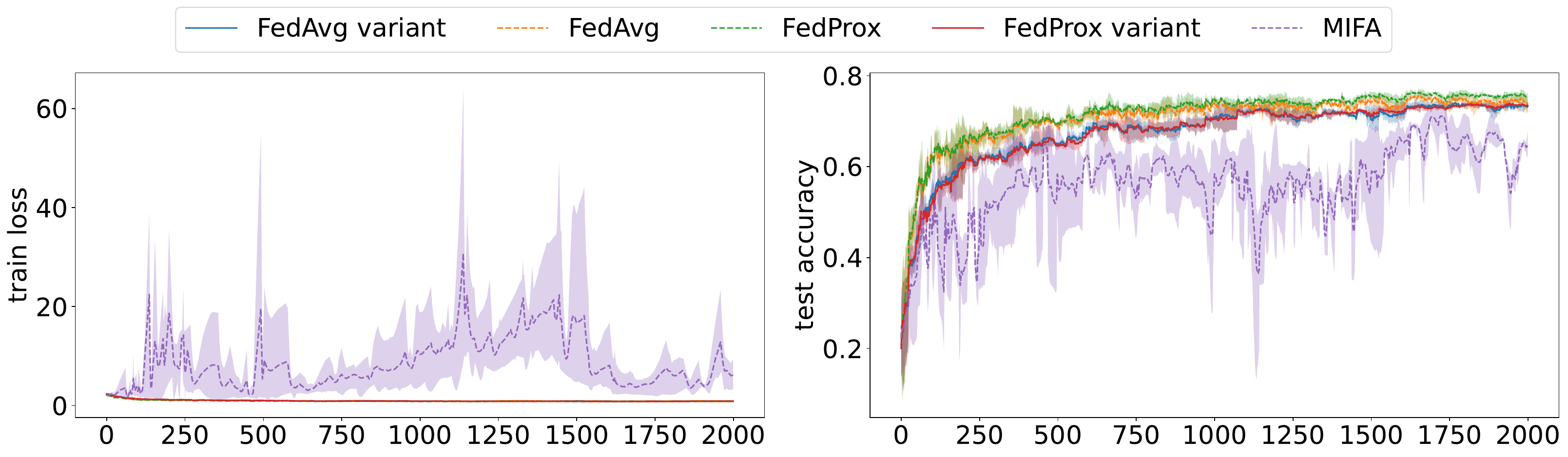}
\includegraphics[width=\textwidth]{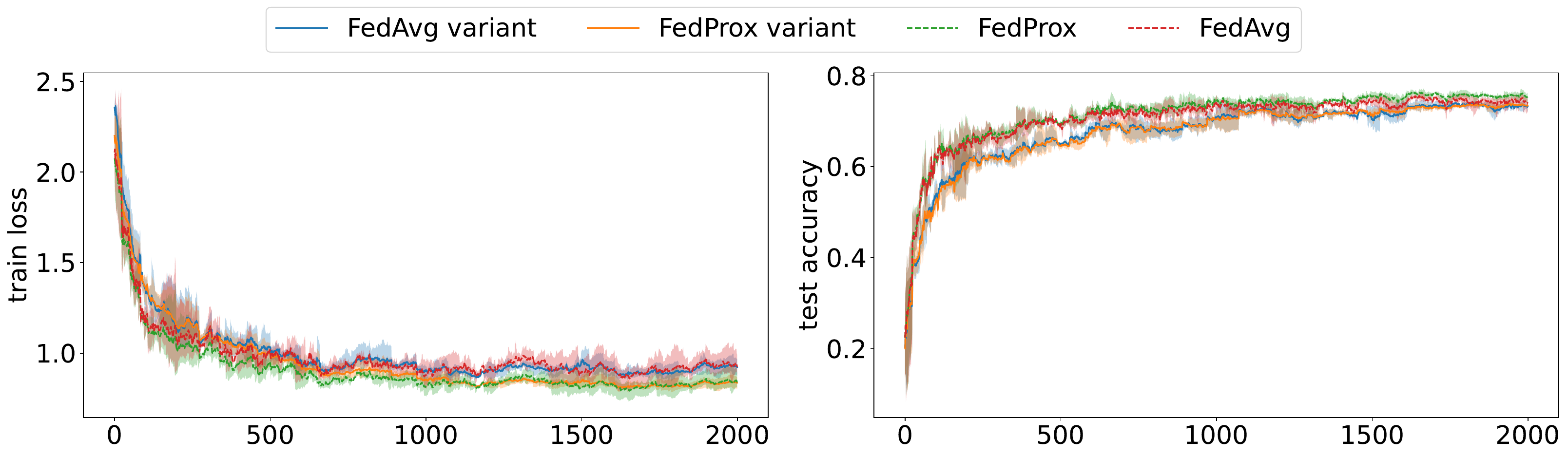}
\includegraphics[width=\textwidth,trim={0 0 0 1cm},clip]{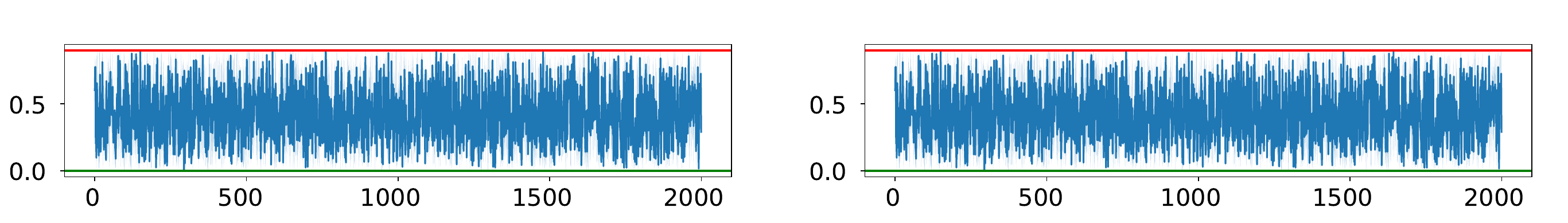}
% %
\begin{minipage}{0.49\textwidth}
\centering
\quad FedAvg variant dropout $\epsilon_t \in [0.00,0.90)$.
\end{minipage}
\begin{minipage}{0.49\textwidth}
\centering
\qquad FedProx variant dropout $\epsilon_t \in [0.00,0.90)$.
\end{minipage}
\vskip -1pt
\caption{ Synthetic datasets: comparisons with baselines with dropout fraction $\epsilon =0.9$ on adversarial client unavailability scheme in Section \ref{subsec: adversarial client unavailability}. 
}
% %
\label{fig:synthetic_11_baseline}
\end{figure}

\begin{figure}[htb]
\centering
\includegraphics[width=\textwidth]{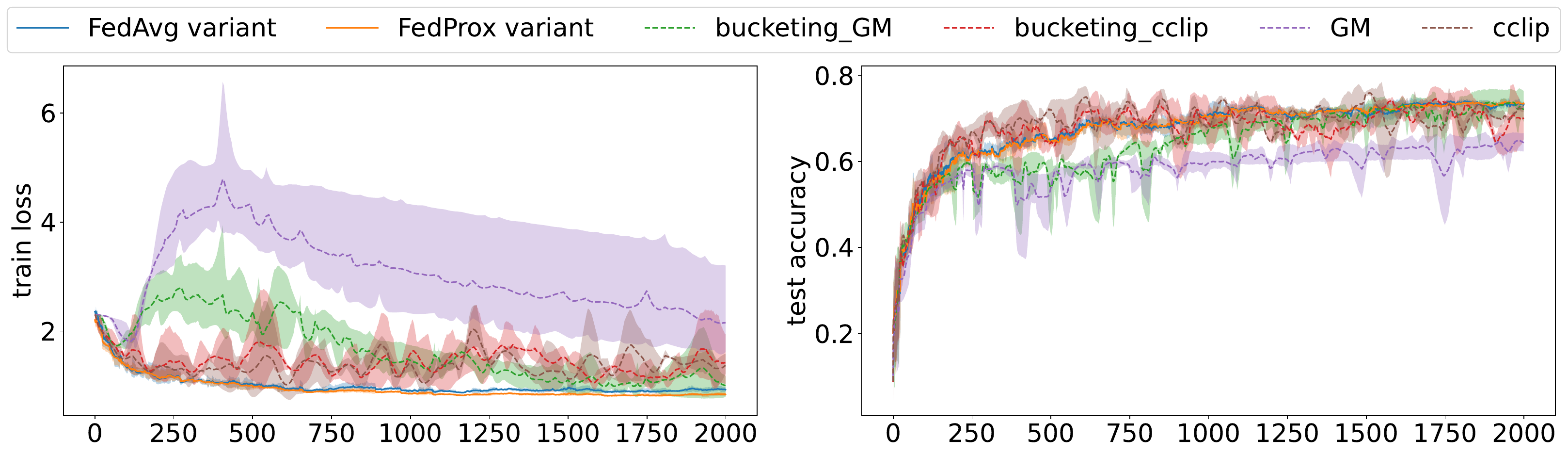}
\caption{Synthetic datasets: comparisons with Byzantine-resilient algorithms with dropout fraction $\epsilon =0.9$ on adversarial client unavailability scheme in Section \ref{subsec: adversarial client unavailability}.}
\label{fig:synthetic_11_byzantine}
\end{figure}

%\nbr{no plot with $\epsilon=0.8$?}
The first row of Fig.\,\ref{fig:synthetic_11_baseline} shows the comparisons with the baselines. 
MIFA's curve suffers from high fluctuations and does not converge till the end of training.
We plot in the second row the comparisons without MIFA. 
% , which are consistently noisy and do not abate throughout the training,
The performances of our variants are similar when compared with the FedAvg and FedProx.
% Although the test accuracies are similar, our variants achieve lower train loss when compared with the FedAvg and FedProx.
Due to the quantity skewness, the actual dropout fraction also fluctuates, and it is possible that no client is dropped in one round.
We plot in the third row $\epsilon_t\triangleq  (\sum_{i\in\tilde{\calS}_t\setminus \calS_t} n_i) /(K N / M)$.
Nevertheless, our unavailability scheme ensures that $\max_{t\in[T]} \epsilon_t \le \epsilon$.
% fulfilling the requirement in~\prettyref{ass:adversarial}.
Additional results with other choices of $\epsilon$ are similar and are reported in~\prettyref{app: exp}. 
% be observed from Fig.\,\ref{fig:synthetic_11_0.7_baseline} and \ref{fig:synthetic_11_0.7_byzantine}.
Similarly, we observe in Fig.\,\ref{fig:synthetic_11_byzantine} that the fluctuations of the Byzantine-resilient algorithms are also severe.

\clearpage
\newpage
\bibliographystyle{IEEEtran}
\bibliography{SM}

\newpage
\appendices

%\subsection{Proof of \prettyref{cor:convergence-rate}}
\section{Characterization of convergence speed in the non-convex case}
\label{app:pf-cor-rate}

% Below we show the convergence rates under some concrete schemes of the learning rate based on~\prettyref{thm:nonconvex}. 
% In the first case, $\eta_t$ depends on the eventual termination time $T$, which may be chosen based on a prescribed level of optimality;
% in the second case, $\eta_t$ is independent of $T$, so the program can be executed indefinitely.
% % until the desired optimality is attained. 
%Recall that we only present the results for the FedAvg, and the results for the FedProx are entirely analogous yet with $s=1$.  
% \begin{corollary}
% %\label{cor:convergence-rate}
% Let $\Delta = F(\theta_0)-F_{\min}$.
% We have the following convergence rate for the FedAvg update rule under specific choices of the learning rates:
% \begin{itemize}
% \item Suppose $\eta_t= \frac{1}{10 \beta  sLB^2} \wedge \frac{1}{\beta s \sqrt{p T L} (G+\sigma)}$. Then
% \begin{align*}
% &\Expect \norm{\nabla F(\theta_R)}^2
% \le O \pth{ \epsilon \pth{G+\sigma}^2}  + O \pth{ 
% \frac{ \Delta L B^2}{ pT } 
% + (\Delta+1) \pth{G+\sigma}\sqrt{\frac{L}{pT} }}.
% \end{align*}
% \item Suppose $\eta_t = \frac{\eta_0}{\sqrt{t+1}}$ for all $t\ge 0$, where $ \eta_0=\frac{1}{10 \beta  sLB^2} \wedge \frac{1}{\beta s \sqrt{pL}(G+\sigma)}$. Then
% \begin{align*}
% &\Expect \norm{\nabla F(\theta_R)}^2
% \le O\pth{ \epsilon \pth{G+\sigma}^2 }  + O \pth{ 
% \frac{ \Delta L B^2}{ p\sqrt{T} } 
% + (\Delta+1) \pth{G+\sigma}\sqrt{\frac{L\log^2 T}{pT} }}.
% \end{align*}
% % \item Suppose $\eta_t = ... \frac{1}{t+1}$. Then
% \end{itemize}
% \end{corollary}

In \prettyref{cor:convergence-rate}, 
the $\tilde{O}(1/\sqrt{T})$
convergence rate matches that of the standard stochastic gradient descent for optimizing nonconvex functions \cite{ghadimi2013stochastic}. In fact, the $O(1/\sqrt{T})$ convergence rate is the best possible for any first-order method that has only access to noisy gradients (see e.g.~\cite[Theorem 3]{arjevani2022lower}).
When there is no adversarial dropout, our results in~\prettyref{cor:convergence-rate} are comparable to the state-of-the-art convergence results for FedAvg or FedProx. Specifically, For FedAvg, decay terms similar to $\Delta B^2/T$ and $(G+\sigma)/\sqrt{T}$ also appear in~\cite[Thoerem V]{karimireddy2020scaffold}.
For FedProx, \cite[Theorem 1]{yuan2022} assumes $B=0$ and obtains a decay term similar to $(\Delta+G^2)/\sqrt{T}$.

\begin{proof}[Proof of~\prettyref{cor:convergence-rate}]
In the first case, $\eta_t=\eta_0$ is a constant for a given $T$. We obtain from \prettyref{thm:nonconvex} that
\[
\Expect\Norm{\nabla F(\theta_R)}^2
\le \frac{3\Delta}{\beta p s T \eta_0} 
+ \pth{4 \epsilon+ c \beta s L \eta_0}\pth{G+\sigma}^2.
\]
Plugging $\eta_0\le \frac{1}{\beta  \sqrt{p T L} (G+\sigma) }$ and $\frac{1}{\eta_0} \le 10\beta  L B^2 + \beta  \sqrt{p T L} (G+\sigma)$ yields that
\[
\Expect\Norm{\nabla F(\theta_R)}^2
\le \frac{30\Delta LB^2}{ p s T } +
\sqrt{\frac{L}{pT}}\left(3\Delta/s + c s \right) \pth{G+\sigma}
+ 4 \epsilon \pth{G+\sigma}^2.
\]

In the second case where $\eta_t=\eta_0/\sqrt{t+1}$, it follows from \prettyref{thm:nonconvex} that
\[
\Expect\Norm{\nabla F(\theta_R)}^2
\le \frac{3\Delta}{\beta p s \eta_0 \sum_{t=0}^T \frac{1}{\sqrt{t+1}}} 
+ \pth{4\epsilon+ c\beta s L \eta_0\frac{\sum_{t=0}^T \frac{1}{{t+1}}}{\sum_{t=0}^T \frac{1}{\sqrt{t+1}}}}\pth{G+\sigma}^2 .
\]
Note that $\sum_{t=0}^T \frac{1}{\sqrt{t+1}}
\ge \int_{1}^{T+2} \frac{1}{\sqrt{x}} dx
= 2\sqrt{T+2}-2 \ge \sqrt{T}$
%\le 1+ \int_{1}^{T+1} \frac{1}{\sqrt{x}} dx = 2 \sqrt{T+1}-1 \le 2\sqrt{T} $ 
and $\sum_{t=0}^T \frac{1}{{t+1}}\le 1+ \int_1^{T+1} \frac{1}{x} dx =\log(e(T+1))$.
Plugging $\eta_0\le \frac{1}{\beta  \sqrt{pL}(G+\sigma)}$ and $\frac{1}{\eta_0}\le {10 \beta  LB^2} + {\beta  \sqrt{pL}(G+\sigma)}$ yields that
\[
\Expect\Norm{\nabla F(\theta_R)}^2
\le \frac{30\Delta L B^2}{ p s \sqrt{T} }
+ \sqrt{\frac{L}{pT}}\left(3\Delta/s + cs \log(e(T+1))  \right) \pth{G+\sigma}
+ 4\epsilon \pth{G+\sigma}^2 .
\]
\end{proof}

\section{Proofs of upper bounds}
\label{app:pf-ub}

\subsection{Proof of~\prettyref{thm:nonconvex}}
\label{app:pf-ub-nonconvex}
We first prove the results for the FedAvg update rule. 
\subsubsection{Proofs of auxiliary lemmas} \label{sec:lemma_proof}

\begin{proof}[Proof of~\prettyref{lmm:s-steps-diff} ]
    By the definition of $\kappa$,
    $$
    \kappa \eta_t^2\binom{s}{2} L_i
    \ge \frac{(1+\eta_t L_i)^s - 1- s \eta_t L_i}{L_i}.
    $$
   Hence it suffices to show 
   % \ls{I cannot build up the connection immediately. }
    \begin{equation}
    \label{eq:s-step-diff-proof}
    \norm{\theta-\calG_{i,t}^s(\theta;\eta_t) -s\eta_t \nabla \ell_{i,t}(\theta)}
    \le \frac{(1+\eta_t L_i)^s - 1- s \eta_t L_i}{(\eta_t L_i)^2} \eta_t^2 L_i\norm{\nabla \ell_{i,t}(\theta)}.
    \end{equation}
    % \ls{Do we need to simplify the terms above?}
    We prove~\eqref{eq:s-step-diff-proof} holds for all $s\ge 1$ by induction.
    The base case $s=1$ follows from the definition of gradient mapping $\calG_{i,t}$ in~\eqref{eq:gradient}.
    Suppose~\eqref{eq:s-step-diff-proof} holds true for $s=1,\dots,n-1$, where $n\ge 2$.
    Next we prove~\eqref{eq:s-step-diff-proof} for $s = n$.
    By the telescoping sum
    \[
    \calG_{i,t}^n(\theta;\eta_t)-\theta
    =  \sum_{s=0}^{n-1}\calG_{i,t}^{s+1}(\theta;\eta_t)-\calG_{i,t}^{s}(\theta;\eta_t)
    = -\eta_t  \sum_{s=0}^{n-1} \nabla \ell_{i,t}(\calG_{i,t}^{s}(\theta;\eta_t)),
    \]
    we obtain
    \begin{align}
    \norm{\theta-\calG_{i,t}^n(\theta;\eta_t) -n\eta_t \nabla \ell_{i,t}(\theta)}
    & =\eta_t \norm{ \sum_{s=0}^{n-1}(\nabla \ell_{i,t}(\calG_{i,t}^{s}(\theta;\eta_t))-\nabla \ell_{i,t}(\theta)) }
    \le  \eta_t L_i \sum_{s=0}^{n-1} \norm{\calG_{i,t}^{s}(\theta;\eta_t)-\theta}\nonumber\\
    & \le \eta_t L_i \sum_{s=1}^{n-1}\pth{\norm{\theta-\calG_{i,t}^s(\theta;\eta_t) -s\eta_t \nabla \ell_{i,t}(\theta)} + s \eta_t \norm{\nabla \ell_{i,t}(\theta)}},\label{eq:n-step-ub}
    \end{align}
    where in the first inequality we used~\eqref{eq:L-Lipschitz}.
    Applying the induction hypothesis~\eqref{eq:s-step-diff-proof} for $s=1,\dots,n-1$, we get
    \[
    \sum_{s=1}^{n-1}\norm{\theta-\calG_{i,t}^s(\theta;\eta_t) -s\eta_t \nabla \ell_{i,t}(\theta)}
    \le \pth{(1+\eta_t L_i)^n-1-n\eta_t L_i -\binom{n}{2}(\eta_t L_i)^2}\frac{\norm{\nabla \ell_{i,t}(\theta)}}{\eta_t L_i^2}.
    \]
    Plugging into~\eqref{eq:n-step-ub}, we conclude~\eqref{eq:s-step-diff-proof} for $s=n$. The proof is completed. 
\end{proof}

\begin{proof}[Proof of \prettyref{lmm:adversarial-attack}]
% By the triangle inequality,
% \[
% \norm{
% \sum_{i\in \calS_t}w_i \nabla \ell_i(\theta_t) - p \nabla F(\theta_t)
% }
% \le 
% \norm{
% \sum_{i\in \tilde\calS_t}w_i \nabla \ell_i(\theta_t) - p \nabla F(\theta_t)
% }
% +  \norm{\sum_{i\in \tilde\calS_t/\calS_t}w_i \nabla \ell_i(\theta_t)}.
% \]
% For the first term, ...
For any $t\ge 1$, it holds that 
% we get
\begin{align}
    \norm{\sum_{ i\in \tilde\calS_t \backslash \calS_t}w_i \nabla \ell_{i,t}(\theta_t)}^2
    & = \sum_{i,j\in \tilde \calS_t \backslash\calS_t}w_iw_j\iprod{\nabla \ell_{i,t}(\theta_t)}{\nabla \ell_{j,t}(\theta_t)}\nonumber\\
    & \overset{(a)}{\le} \sum_{ i,j\in \tilde\calS_t \backslash \calS_t}w_iw_j \frac{\norm{\nabla \ell_{i,t}(\theta_t)}^2+\norm{\nabla \ell_{j,t}(\theta_t)}^2}{2}\nonumber\\
    & = \pth{\sum_{ i\in \tilde\calS_t \backslash \calS_t}w_i} \pth{\sum_{i\in \tilde\calS_t\setminus\calS_t}w_i\norm{\nabla \ell_{i,t}(\theta_t)}^2 } \nonumber\\
    & \le \pth{\sum_{ i\in \tilde\calS_t \backslash \calS_t}w_i} \left(  \sum_{i\in \tilde\calS_t}w_i\norm{\nabla \ell_{i,t}(\theta_t)}^2 \right).        
    % & \le \pth{\sum_{i\not\in\calS_t}w_i} \left( B^2 \norm{\nabla F(\theta)}^2 + G^2\right),
    \label{eq:droput-gradient}
\end{align}
where inequality (a) follows from the fact that $\iprod{u}{v}\le \frac{1}{2}(\norm{u}^2+\norm{v}^2)$ for any $u, v$ of the same dimension. Notably, Eq.\eqref{eq:droput-gradient} is the key step that enables us to control $\expect{\norm{\sum_{i\in \tilde\calS_t\setminus\calS_t}w_i \nabla \ell_{i,t}(\theta_t)}^2\mid \calF_t}$ in terms of $\epsilon$.
%
% \mx{Based on the statement highlighted in red, I think $\tilde\calS_t/\calS_t$ should be replaced by $\tilde\calS_t\setminus\calS_t$.}
%
By~\prettyref{ass:adversarial}, we have
\begin{equation}\label{eq: fedavg: weight bound}
\sum_{i\in \tilde\calS_t \backslash \calS_t}w_i
= \sum_{i\in \tilde\calS_t \backslash \calS_t}\frac{n_i}{N}
\le \frac{K\epsilon}{M} = p \epsilon.
\end{equation} 
 Hence, it suffices to show 
\[
\expect{ \sum_{i\in \tilde\calS_t}w_i\norm{\nabla \ell_{i,t}(\theta_t)}^2 \mid \calF_t}\le p \pth{B^2 \norm{\nabla F(\theta_t) }^2 + G^2 + \sigma^2},
\]
for which we have 
%To see this, 
\begin{align}
\expect{ \sum_{i\in \tilde\calS_t}w_i\norm{\nabla \ell_{i,t}(\theta_t)}^2 \mid \calF_t }
& = \expect{ \sum_{i=1}^M w_i\norm{\nabla \ell_{i,t}(\theta_t)}^2 \indc{i\in \tilde\calS_t} \mid \calF_t } \nonumber\\
& \overset{(a)}{=} \sum_{i=1}^M w_i \expect{\norm{\nabla \ell_{i,t}(\theta_t)}^2 \mid \calF_t} \expect{\indc{i\in \tilde\calS_t} \mid \calF_t } \nonumber\\
& \overset{(b)}{=} p \sum_{i=1}^M w_i \expect{\norm{\nabla \ell_{i,t}(\theta_t)}^2 \mid \calF_t} \nonumber\\
& \overset{(c)}{\le} p \sum_{i\in [M]} w_i \pth{ \norm{\nabla F_i(\theta_t)}^2 + \sigma_i^2 } \nonumber \\ 
% & = p \pth{ \sum_{i \in [M]} w_i \pth{ \norm{\nabla F_i(\theta_t)}^2 + \frac{\sigma^2}{n_i}  } }\\
& \overset{(d)}= p \pth{ \sum_{i \in [M]} w_i \norm{\nabla F_i(\theta_t)}^2 + \sigma^2  } \nonumber  \\
& \overset{(e)}{\le} p \pth{ B^2 \norm{\nabla F(\theta_t)}^2 + G^2 + \sigma^2 },\label{eq:noisy_gradient_bound_2}
\end{align}
where equality (a) is true because, as mentioned before, $\tilde \calS_t$ and $z_{i,t}$ are mutually independent; equality (b) holds because $\tilde\calS_t$ is independent of $\calF_t$; inequality (c) follows from the fact that $\theta_t$ is adapted to $\calF_t$ and \eqref{eq:sgd-variance}, so that
$$
\expect{\norm{\nabla \ell_{i,t}(\theta_t)}^2 \mid \calF_t}
=\norm{\nabla F_i(\theta_t)}^2 + 
\expect{\norm{\nabla \ell_{i,t}(\theta_t) - \nabla F_i(\theta_t) }^2 \mid \calF_t}
\le \norm{\nabla F_i(\theta_t)}^2 + \sigma_i^2;
$$
% and
% \begin{align}
% \expect{\norm{\nabla \ell_{i,t}(\theta_t) - F_i(\theta_t) }^2 \mid \calF_t}
% &=\expect{ \norm{ \frac{1}{n_i} \sum_{j=1}^{n_i} \ell_i(\theta_t, z_{i,t}^{(j)} )- F_i(\theta_t) }^2 \mid \calF_t} \nonumber \\
% & =\frac{1}{n_i^2}\sum_{j=1}^{n_i}
% \expect{ \norm{\ell_i(\theta_t, z_{i,t}^{(j)} )- F_i(\theta_t) }^2\mid \calF_t } \nonumber \\
% & \le \sigma_i^2; \label{eq:variance_bound}
% \end{align}
equality (d) uses the definition of $\sigma^2$; and inequality (e) follows from Assumption~\ref{ass:BG}.  
%where $n \triangleq N/M$ denotes the average number of data points drawn at each round per client. 
% \nb{Combining the above and some condition on $\sum_{i\in \tilde\calS_t/\calS_t}w_i$ to conclude the lower bound of 
The conclusion follows from \eqref{eq:droput-gradient}, \eqref{eq: fedavg: weight bound}, and \eqref{eq:noisy_gradient_bound_2}.
\end{proof}

\subsubsection{Bound on the decrease of the objective function}
% Equipped with the above lemmas, 
We are ready to analyze the decrease of the objective function values specified on the right-hand side of~\eqref{eq:decrease-objective} under our dropout model. % (random + adversarial). 

It follows from our aggregation rule~\eqref{eq:global_update} that, for every $t$,
\begin{align}
\theta_{t+1} - \theta_{t}
& = \beta \sum_{i\in\calS_t} w_i (\calG_{i,t}^s(\theta_t;\eta_t)-\theta_t) \nonumber\\
& = \beta \sum_{i\in\calS_t} w_i (\calG_{i,t}^s(\theta_t;\eta_t)-\theta_t+s\eta_t \nabla \ell_{i,t}(\theta_t))
- \beta s \eta_t \sum_{i\in\calS_t}  w_i\nabla \ell_{i,t}(\theta_t). \label{eq:one-round}
\end{align}

\paragraph{Bounding $\expect{ \iprod{\nabla F(\theta_t)}{\theta_{t+1}-\theta_t} \mid \calF_t}$:}
Plugging~\eqref{eq:one-round}, the second term on the right-hand side of~\eqref{eq:decrease-objective} % main drift term 
is decomposed into 
\begin{align}
\iprod{\nabla F(\theta_t)}{\theta_{t+1}-\theta_t}
& = \beta \cdot
\underbrace{
\iprod{\nabla F(\theta_t)}{\sum_{i\in\calS_t} w_i (\calG_{i,t}^s(\theta_t;\eta_t)-\theta_t+s\eta_t \nabla \ell_{i,t}(\theta_t))}
}_{\rm (I)} \nonumber \\
&\qquad  - \beta s \eta_t  \cdot
\underbrace{
\iprod{\nabla F(\theta_t)}{\sum_{i\in\calS_t}  w_i\nabla \ell_{i,t}(\theta_t)}
}_{\rm (II)}.
\label{eq:decomp}
% + \norm{\nabla F(\theta_t)}\norm{\sum_{i\in\calS_t} w_i (P_i^s(\theta_t)-\theta_t+s\eta \nabla F_i(\theta_t))}
\end{align}

On (I), we have 
\begin{align*}
 \mathrm{(I)} & \le \norm{\nabla F(\theta_t)}
 \norm{\sum_{i\in\calS_t} w_i (\calG_{i,t}^s(\theta_t;\eta_t)-\theta_t+s\eta_t \nabla \ell_{i,t}(\theta_t))} \\
 %& \le \norm{\nabla F(\theta_t)}
 %\sum_{i\in\calS_t} w_i
 %\norm{\calG_{i,t}^s(\theta_t;\eta)-\theta_t+s\eta_t \nabla \ell_{i,t}(\theta_t)} 
\end{align*}
and hence 
\begin{align}
\expect{\mathrm{(I)} \mid \calF_t }
% &\le 
% p \norm{\nabla F(\theta_t)}
% \kappa \eta_t^2 L \binom{s}{2}
% \left( \sum_{i\in [M]} w_i 
%  \norm{\nabla F_i(\theta_t)} + \frac{\sigma}{\sqrt{n}} \right) \nonumber \\
 %& 
 \le p \norm{\nabla F(\theta_t)}
\kappa \eta_t^2 L \binom{s}{2}
\left( B \norm{\nabla F(\theta_t)} + G +\sigma \right)\label{eq:drift-II},
\end{align}
where the last inequality holds by~\prettyref{eq:objective-inconsistency}.

For the lower bound of (II), we consider the difference of clients belonging to $\calS_t$ and $\tilde\calS_t$: 
\begin{align*}
\mathrm{(II)} &= \iprod{\nabla F(\theta_t)}{\sum_{i\in \tilde \calS_t}  w_i\nabla \ell_{i,t}(\theta_t)}
- \iprod{\nabla F(\theta_t)}{\sum_{ i\in\tilde \calS_t \backslash \calS_t}  w_i\nabla \ell_{i,t}(\theta_t)} \\
& \ge \iprod{\nabla F(\theta_t)}{\sum_{i\in \tilde \calS_t}  w_i\nabla \ell_{i,t}(\theta_t)}
- \norm{\nabla F(\theta_t) }
\norm{\sum_{ i\in \tilde \calS_t \backslash \calS_t}  w_i\nabla \ell_{i,t}(\theta_t)}.
\end{align*}
% Applying some concentration inequalities to show
% \[
% \norm{\sum_{i\in\tilde\calS_t}  w_i\nabla \ell_i(\theta) - p \nabla F(\theta)}
% \le C_1 \norm{\nabla F(\theta)} + C_2.
% \]
% \nb{missing step; add uniform sampling probability $p$ to setup. ref: FedProx (?) Should we renormalized the gradients from participating clients by $1/p$?}
Thus, 
\begin{align}
\expect{\mathrm{(II)} \mid \calF_t} & \ge \expect{\iprod{\nabla F(\theta_t)}{\sum_{i\in \tilde \calS_t}  w_i\nabla \ell_{i,t}(\theta_t)} - \norm{\nabla F(\theta_t)}\norm{\sum_{ i\in \tilde \calS_t \backslash \calS_t}  w_i\nabla \ell_{i,t}(\theta_t)} \mid \calF_t} \nonumber \\
& = \iprod{\nabla F(\theta_t)}{\expect{\sum_{i\in \tilde \calS_t}  w_i\nabla \ell_{i,t}(\theta_t) \mid \calF_t }}
- \norm{\nabla F(\theta_t)} \expect{\norm{\sum_{ i\in \tilde \calS_t \backslash \calS_t}  w_i\nabla \ell_{i,t}(\theta_t)}\mid \calF_t} \nonumber\\
& \overset{(a)}{\ge} p\norm{\nabla F(\theta_t) }^2 - p \norm{\nabla F(\theta_t) }
 \sqrt{\epsilon}
 \left( B \norm{\nabla F(\theta)} + G + \sigma \right),%\nonumber\\
 % & = p \pth{\norm{\nabla F(\theta_t) }^2 - \norm{\nabla F(\theta_t) }
 % \sqrt{\epsilon}
 % \left( B \norm{\nabla F(\theta)} + G + \frac{\sigma}{\sqrt{n}} \right)},  
\label{eq:drift-I}
\end{align} 
where equality (a) follows from~\eqref{eq:tildeS-mean}  and~\prettyref{lmm:adversarial-attack}. 
%\ls{Do we need further grouping of the terms? recheck this! }
% \begin{align}
% \expect{\mathrm{(I)} \mid \calF_t }
% % & \ge  p \norm{\nabla F(\theta_t) }^2
% % - p \norm{\nabla F(\theta_t) }
% % \sqrt{\epsilon}
% % \left( \sum_{i \in [M]} w_i 
% % \norm{\nabla F_i(\theta)}^2 + \frac{\sigma}{\sqrt{n}} \right)\\
% & \ge p \pth{ \norm{\nabla F(\theta_t) }^2
% -\norm{\nabla F(\theta_t) }
% \sqrt{\epsilon}
% \left( B \norm{\nabla F(\theta_t) } + G + \frac{\sigma}{\sqrt{n}} \right)
% }.\label{eq:drift-I}
% \end{align}

Combining~\eqref{eq:drift-I} and~\eqref{eq:drift-II} together and plugging it back to~\prettyref{eq:decomp}, we get that 
\begin{align}
\expect{ \iprod{\nabla F(\theta_t)}{\theta_{t+1}-\theta_t} \mid \calF_t}
& \le 
- \beta p s \eta_t \norm{\nabla F(\theta_t) }^2
+ \beta  p s \eta_t \norm{\nabla F(\theta_t) }
\sqrt{\epsilon}
\left( B \norm{\nabla F(\theta_t) } + G + \sigma \right) \nonumber\\
& \qquad 
+ \beta p \norm{\nabla F(\theta_t)}
\kappa \eta_t^2 L \binom{s}{2}
\left( B \norm{\nabla F(\theta_t)} + G +\sigma \right). \label{eq:main-drift}
\end{align}
% We have shown that 
% $$
% \norm{\sum_{i\notin\calS_t}  w_i\nabla F_i(\theta)} \le C_1 \nabla \norm{F(\theta)} + C_2. 
% $$
% and 
% $$
% \expect{\sum_{i\in [M]} w_i \norm{\nabla \ell_{i,t}(\theta_t) -\nabla F_i(\theta_t) } \mid \calF_t } \le \frac{\sigma}{\sqrt{n}}.
% $$
% It follows that 
% $$
% \expect{ (I) \mid \calF_t }
% \ge \norm{\nabla F(\theta_t) }^2
% - \norm{\nabla F(\theta_t)}
% \left( C_1 \norm{\nabla F(\theta_t)} + C_2 + \frac{\sigma}{\sqrt{n}}\right).
% $$

\paragraph{Bounding $\expect{\norm{\theta_{t+1}-\theta_t}^2 \mid \calF_t}$:} 
% 
%Similarly, plugging~\eqref{eq:one-round} into the third term on the right-hand side of~\eqref{eq:smooth-decrease}, %for the high-order term,
By~\eqref{eq:one-round}, we have  
\begin{align*}
\norm{\theta_{t+1}-\theta_t}
&= \norm{\beta \sum_{i\in\calS_t} w_i (\calG_{i,t}^s(\theta_t;\eta_t)-\theta_t+s\eta_t \nabla \ell_{i,t}(\theta_t))
- \beta s \eta_t \sum_{i\in\calS_t}  w_i\nabla \ell_{i,t}(\theta_t)} \\
&\overset{(a)}{\le} \beta s\eta_t \norm{\sum_{i\in\calS_t}  w_i\nabla \ell_{i,t}(\theta_t)}
+ \beta  \kappa \eta_t^2 L \binom{s}{2} \sum_{i\in\calS_t} w_i  \norm{\nabla \ell_{i,t}(\theta_t)} \\
& \le \beta s\eta_t \left( 1+ \kappa \eta_t L (s-1)/2 \right)\sum_{i\in\calS_t} w_i  \norm{\nabla \ell_{i,t}(\theta_t)} \\
& \le \beta s\eta_t \left( 1+ \kappa \eta_t L (s-1)/2 \right)
\sqrt{\sum_{i \in \tilde\calS_t} w_i }\sqrt{\sum_{i \in \tilde\calS_t} w_i  \norm{\nabla \ell_{i,t}(\theta_t)}^2}.
\end{align*}
where inequality (a) follows from Lemma \ref{lmm:s-steps-diff} and the fact that $L = \max_{i\in [M]}L_i$; the last inequality holds by the Cauchy-Schwarz inequality.   
Applying~\prettyref{eq:noisy_gradient_bound_2}
and using $\sum_{i \in \tilde\calS_t} w_i \le 1$, we deduce that
\begin{equation}
\label{eq:high-order}
\expect{\norm{\theta_{t+1}-\theta_t}^2 \mid \calF_t}
\le \beta^2 s^2\eta_t^2 \left( 1+ \kappa \eta_t L (s-1)/2 \right)^2
p \left( B^2 \norm{\nabla F(\theta)}^2 + G^2 + \sigma^2 \right). 
\end{equation}
% \nbr{I am confused. Did you forget a $p$ factor in the above?}

\paragraph{Bounding $\expect{F(\theta_{t+1}) \mid \calF_t }$:}
Combining~\eqref{eq:main-drift} and~\eqref{eq:high-order} together, we apply~\eqref{eq:decrease-objective} and get 
$$
\expect{F(\theta_{t+1}) \mid \calF_t }
\le F(\theta_t) - \bar{a}_t \norm{\nabla F(\theta_t)}^2 + b_t \norm{\nabla F(\theta_t)} + c_t,
$$
where 
\begin{align*}
\bar{a}_t & = \beta p s \eta_t \left( 1 - \sqrt{\epsilon}B - \frac{\kappa (s-1) 
\eta_t L}{2} B   - \beta \frac{s \eta_t L}{2} 
\left( 1+ \frac{\kappa (s-1) \eta_t L}{2}\right)^2 B^2 \right), \\
b_t &= \beta p s \eta_t \left( \sqrt{\epsilon} \pth{G+\sigma} + \frac{\kappa  (s-1) \eta_t L}{2}  \pth{ G+ \sigma}   \right), \\
c_t & = \beta^2 p s \eta_t  \frac{s \eta_t L}{2} \left( 1+ \frac{\kappa (s-1) \eta_t  L}{2}\right)^2 \pth{G^2 + \sigma^2}.
\end{align*}
By the conditions on $\eta_t$ and $\epsilon$, we have $\sqrt{\epsilon}B <c_0=0.1$ and $\eta_t s L \le \eta_t s L \beta B^2 \le 0.1$,  and thus $\kappa \le \frac{e^{0.1}-1-0.1}{(0.1)^2/2} \le 1.0342$. 
Then we get 
$$
\bar{a}_t \ge {C_0} \cdot \beta p s \eta_t \triangleq a_t,
$$
where $C_0 \ge 1-c_0-0.05\kappa -0.05(1+0.05\kappa)^2 \ge 0.7929 >0$
%${C_0}\approx 0.79$ 
is a constant. 
% \ls{use another notation -- to emphasize it is a constant?}
Hence, 
$$
\expect{F(\theta_{t+1}) \mid \calF_t }
\le F(\theta_t) - a_t \norm{\nabla F(\theta_t)}^2 + b_t \norm{\nabla F(\theta_t)} + c_t,
$$

%\subsubsection{Bound on the gradients by telescoping sum}

% we have $\kappa<1.1$, 
% and thus $a_t \ge p s \eta_t/4 $.

%\paragraph{Finishing the proof of~\prettyref{eq:telescope} for FedAvg:}
\paragraph{Finishing the proof for FedAvg:}

Note that 
$$
\norm{\nabla F(\theta_t)} \le a
\norm{\nabla F(\theta_t)}^2 + \frac{1}{4a}, \quad \forall a \ge 0.
$$
Picking $a= \frac{a_t}{2b_t}$, we get 
$$
\expect{F(\theta_{t+1})}\le \expect{F(\theta_t)} - \frac{a_t}{2} \Expect{\norm{\nabla F(\theta_t)}^2} + \frac{b_t^2}{2a_t} + c_t. 
$$
By telescoping sum, we get 
$$
\expect{F(\theta_{T+1})} \le F(\theta_0)
- \frac{1}{2} \sum_{t=0}^T a_t
\Expect\norm{\nabla F(\theta_t)}^2 +\sum_{t=0}^T \left( \frac{b_t^2}{2a_t} + c_t\right) .
$$
Rearranging the terms, we deduce that 
\begin{align}
\frac{\sum_{t=0}^T \eta_t \Expect\norm{\nabla F(\theta_t)}^2 }{\sum_{t=0}^T \eta_t} =\frac{\sum_{t=0}^T a_t \Expect\norm{\nabla F(\theta_t)}^2 }{\sum_{t=0}^T a_t}
\le  \frac{2 \left( F(\theta_0)-F_{\min} \right) }{\sum_{t=0}^T a_t} +
\frac{\sum_{t=0}^T \left( \frac{b_t^2}{a_t} + 2c_t\right)}{\sum_{t=0}^T a_t}. \label{eq:telescope}
\end{align}
Since $a_t= {C_0}  \beta ps\eta_t$, we have 
% \nbr{maybe briefly explain what is $C_1, C_2, c'?$}
\begin{align*}
\frac{b_t^2}{a_t} + 2c_t
& \le \frac{\beta}{{C_0}} (ps\eta_t) 2(\epsilon + (C_1 s \eta_t L)^2 )  \pth{ G+ \sigma}^2 + \beta^2 (ps\eta_t) (C_2 s \eta_t L) \pth{ G+ \sigma}^2 \\
& \le \frac{2\epsilon}{{C_0}} \pth{ G+ \sigma}^2 \beta ps\eta_t
 + C_3\beta^2 p  s^2 \eta_t^2 L \pth{ G+ \sigma}^2.
\end{align*}
where $C_1=\frac{\kappa}{2}$, $C_2= (1+\frac{0.1\kappa}{2})^2$, and $C_3= C_2+\frac{0.2C_1^2}{{C_0}\beta}$.
Hence, 
\[
\frac{\sum_{t=0}^T \left( \frac{b_t^2}{a_t} + 2c_t\right)}{\sum_{t=0}^T a_t}
\le \pth{\frac{2\epsilon}{C_0^2}+ \beta \frac{ C' sL \sum_{t=0}^T  \eta_t^2 }{\sum_{t=0}^T  \eta_t}}\pth{ G+ \sigma}^2
\]
for some universal constant $C',$ where $2/C_0^2 \le 2/0.7929^2 <3.2.$
Plugging  the last displayed equation back to~\prettyref{eq:telescope}
completes the proof of~\prettyref{thm:nonconvex} for FedAvg.
% Here, we can have different choices of step sizes $\eta_t$,
% under which we will have different convergence results. 

% \nb{I couldn't get a factor of $p^2$ for the higher-order terms. In fact,
% \[
% \Expect\pth{\sum_{i\in\tilde\calS_t} w_i  \norm{\nabla \ell_i(\theta_t)} \mid \calF_t}^2
% \ge \expect{\sum_{i\in[M]} w_i^2 \norm{\nabla \ell_i(\theta_t)}^2 \indc{i\in\tilde\calS_t} \mid \calF_t }
% = p \sum_{i\in[M]} w_i^2 \pth{\norm{\nabla F_i(\theta_t)}^2 + \frac{\sigma^2}{n_i} }
% \]
% Finally, the sampling variance kicks in as we anticipated for a while. 
% Even for the usual SGD (population analysis with $s=1$ and no adversarial dropout), we need to upper bound
% \[
% \expect{\norm{\sum_{i\in\tilde\calS_t}  w_i\nabla F_i(\theta_t)}^2}
% % \expect{\norm{\sum_{i\in\tilde\calS_t}  w_i\nabla \ell_i(\theta_t)}^2 \mid \calF_t}
% = p^2 \norm{\nabla F(\theta_t)}^2 
% + \Expect\norm{ \sum_{i\in [M]} w_i \nabla F_i(\theta_t)(\indc{i\in\tilde\calS_t}-p) }^2. 
% \]
% Check literature to see what assumption can be imposed here. 
% }
% \nb{
% Some refs:
% \cite{ghadimi2013stochastic} doesn't consider the sampling variance but a general noisy gradient satisfying their Assumption A1.
% \cite{li2020federated} the final error bound has some terms proportional to $\frac{1}{\sqrt{K}}$. I think that's roughly equivalent to taking $\beta=\frac{1}{p}$ in our \eqref{eq:affine-fedavg}.
% }

\subsubsection{Proof for FedProx}

For FedProx, let $\calP_{i,t}(\theta,\eta)$ denote proximal mapping defined as
 \begin{equation}
 \label{eq:proximal}
 \calP_{i,t}(\theta;\eta) \triangleq \arg\min_{z} \ell_{i,t}(z) + \frac{1}{2\eta} \norm{z-\theta}^2.
 \end{equation}
 Then $\theta_{i,t+1}\in \calP_{i,t}(\theta_t;\eta_t)$.

The proof for the FedProx is similar, where we need an analogous result to \prettyref{lmm:s-steps-diff} under the additional condition  that $\lambda_{\min}(\nabla^2 \ell_{i,t}(\theta)) \ge -L_-$ for all $i$. This condition is standard in analyzing the FedProx algorithm \cite{Li2020}. Note that the following result and proof can be readily extended to the case where the proximal problem \eqref{eq:proximal} at each local step is solved inexactly (see e.g.~\cite[Lemma 5]{yuan2022}).   
\begin{lemma}\label{lmm:fedprox-diff}
    Suppose $\lambda_{\min}(\nabla^2 \ell_{i,t}(\theta)) \ge -L_-$. Then for $\eta< \frac{1}{L_-}$, we have
    \[
    \norm{\theta-\calP_{i,t}(\theta;\eta) -\eta \nabla \ell_{i,t}(\theta)}
    \le \frac{\eta^2}{1-\eta L_-}L_i\norm{\nabla \ell_{i,t}(\theta)},
    \quad \forall\theta.
    \]
\end{lemma}
%\nb{
% Do we want to include $\gamma$-inexact minimizer? The argument at this lemma's level has been done in \cite{li2020federated}.
%Remark: extending to $\gamma$-inexact minimizer.
%}
\begin{proof}
    For $\eta<\frac{1}{L_-}$, the local objective function $\tilde{\ell}_{i,t}(z)\triangleq \ell_{i,t}(z) + \frac{1}{2\eta} \norm{z-\theta}^2$ as given in \eqref{eq:proximal} is $(1/\eta-L_-)$-strongly convex in view of the definition~\prettyref{eq:def_strong_convexity}.
    Hence, the optimization program \eqref{eq:proximal} has a unique minimizer. 
    By the first-order optimality condition,
    \[
    \theta-\calP_{i,t}(\theta,\eta) = \eta \nabla \ell_{i,t}( \calP_{i,t}(\theta,\eta) ).
    \]
    By the $(\frac{1}{\eta}-L_-)$-strong convexity of the objective function $\tilde{\ell}_{i,t}(z)$, we have
    \[
    \norm{\calP_{i,t}(\theta,\eta) -\theta} 
    \le \frac{1}{\frac{1}{\eta}-L_-}\norm{
    \nabla \tilde{\ell}_{i,t}(\calP_{i,t}(\theta,\eta))-\nabla \tilde{\ell}_{i,t}(\theta)}
    =\frac{1}{\frac{1}{\eta}-L_-}\norm{ \nabla \ell_{i,t}(\theta)},
    \]
    where we used the facts that $\nabla \tilde{\ell}_{i,t}(\calP_{i,t}(\theta,\eta))=0$
    and $\nabla \tilde{\ell}_{i,t}(\theta)=\nabla \ell_{i,t}(\theta)$.
    Therefore, by the smoothness of $\ell_{i,t}$, we have 
%     applying \prettyref{ass:smooth} yields 
    \begin{align*}        
    \norm{\theta- \calP_{i,t}(\theta,\eta) -\eta \nabla \ell_{i,t}(\theta)}
    & = \eta \norm{\nabla \ell_{i,t}( \calP_{i,t}(\theta,\eta) )-\nabla \ell_{i,t}(\theta)}
    \le \eta L_i \norm{ \calP_{i,t}(\theta,\eta) -\theta} \\
    & \le \frac{\eta^2 L_i}{1-\eta L_-}\norm{\nabla \ell_{i,t}(\theta)}.
    \qedhere
    \end{align*}
\end{proof}

The remaining analysis is similar to the FedAvg update rule.
In particular, following the previous arguments, we upper bound the right-hand side of \eqref{eq:decrease-objective} by
\begin{align}
\expect{ \iprod{\nabla F(\theta_t)}{\theta_{t+1}-\theta_t} \mid \calF_t}
& \le 
- \beta p \eta_t \norm{\nabla F(\theta_t) }^2
+ \beta p \eta_t \norm{\nabla F(\theta_t) }
\sqrt{\epsilon}
\left( B \norm{\nabla F(\theta_t) } + G + \sigma \right) \nonumber\\
& \qquad + \beta p \norm{\nabla F(\theta_t)}
 \frac{\eta_t^2}{1-\eta_t L_-} L 
\left( B \norm{\nabla F(\theta_t)} + G +\sigma \right), \label{eq:main-drift-fedprox}\\
\expect{\norm{\theta_{t+1}-\theta_t}^2 \mid \calF_t}
& \le \beta^2 \eta_t^2 \left( 1+  \frac{\eta_t L}{1-\eta L_-} \right)^2
p \left( B^2 \norm{\nabla F(\theta)}^2 + G^2 + \sigma^2 \right). 
\end{align}
Then, we get
$$
\expect{F(\theta_{t+1}) \mid \calF_t }
\le F(\theta_t) - \bar{a}_t' \norm{\nabla F(\theta_t)}^2 + b_t' \norm{\nabla F(\theta_t)} + c_t',
$$
where 
\begin{align*}
\bar{a}_t' & = \beta p \eta_t \left( 1 - \sqrt{\epsilon}B - \frac{\eta_t L}{1-\eta_t L_-} B   - \beta \frac{\eta_t L}{2} 
\left( 1+ \frac{\eta_t L}{1-\eta_t L_-}\right)^2 B^2 \right), \\
b_t' & = \beta p \eta_t \left( \sqrt{\epsilon} \pth{G+\sigma} + \frac{\eta_t L}{1-\eta_t L_-}  \pth{ G+ \sigma}   \right), \\
c_t' & = \beta^2 p \eta_t  \frac{\eta_t L}{2} \left( 1+ \frac{\eta_t L}{1-\eta_t L_-}\right)^2 \pth{G^2 + \sigma^2}.
\end{align*}
By the conditions on $\eta_t$ and $\epsilon$, we have $\sqrt{\epsilon}B\le c_0=0.1$, $\eta_t L \le \eta_t \beta L B^2 \le 0.1$, and $\eta_t L_-\le 0.1$, and thus $\bar{a}_t' \ge C_0'\cdot \beta p s \eta_t \triangleq a_t'$, where $C_0' \ge 1- c_0 - 0.1/0.9 - 0.05\times (1+0.1/0.9)^2 \ge 0.727$.
Following a similar analysis, we get
\begin{align*}
\frac{\sum_{t=0}^T \eta_t \norm{\nabla F(\theta_t)}^2 }{\sum_{t=0}^T \eta_t} =
\frac{\sum_{t=0}^T a_t' \norm{\nabla F(\theta_t)}^2 }{\sum_{t=0}^T a_t'}
& \le  \frac{2 \left( F(\theta_0)-F_{\min} \right) }{\sum_{t=0}^T a_t'} +
\frac{\sum_{t=0}^T \left( \frac{b_t'^2}{a_t'} + 2c_t'\right)}{\sum_{t=0}^T a_t'} \\
& \le \frac{2 \left( F(\theta_0)-F_{\min} \right) }{\sum_{t=0}^T a_t'} 
+ \pth{\frac{2\epsilon}{C_0'^2} + \beta \frac{cL \sum_{t=0}^T  \eta_t^2 }{\sum_{t=0}^T  \eta_t} } \pth{ G+ \sigma}^2,
\end{align*}
where $2/C_0'^2 \le 2/0.727^2 \le 3.8.$

\subsection{Proof of \prettyref{thm:strongly_convex}}
\label{app:pf-strongly-convex}

\begin{proof}[Proof of \prettyref{thm:strongly_convex}]
We first consider FedAvg. The proof is in the same spirit as that in the non-convex case. 
The major difference is that in the strongly convex setting, we study the iterates of $\Delta_t\triangleq \norm{\theta_t-\theta^*}$, for which we have %In particular, 
\begin{align}
\Delta_{t+1}^2 &=  \norm{\theta_{t+1}-\theta_t+\theta_t-\theta^*}^2 \nonumber \\
& = \Delta_t^2 +  2\iprod{\theta_{t+1}-\theta_t}{\theta_t-\theta^*}+  \norm{\theta_{t+1}-\theta_t}^2 \label{eq:Delta_evol}.
\end{align}

Let us first analyze the drift term $\iprod{\theta_{t+1}-\theta_t}{\theta_t-\theta^*}$. 
Plugging~\eqref{eq:one-round}, we get that 
\begin{align*}
\iprod{\theta_{t+1}-\theta_t}{\theta_t-\theta^*}
& =  \beta 
\underbrace{
\iprod{\theta_t-\theta^*}{\sum_{i\in\calS_t} w_i (\calG_{i,t}^s(\theta_t;\eta_t)-\theta_t+s\eta_t \nabla \ell_{i,t}(\theta_t))}
}_{\rm (I)} \\
&\qquad -\beta s \eta_t 
\underbrace{
\iprod{\theta_t-\theta^*}{\sum_{i\in\calS_t}  w_i\nabla \ell_{i,t}(\theta_t)}
}_{\rm (II)} .
% + \norm{\nabla F(\theta_t)}\norm{\sum_{i\in\calS_t} w_i (P_i^s(\theta_t)-\theta_t+s\eta \nabla F_i(\theta_t))}
\end{align*}

For term $\rm (I)$, analogous to~\prettyref{eq:drift-II}, applying~\prettyref{lmm:s-steps-diff} and~\prettyref{eq:noisy_gradient_bound}, we get
\begin{align}
\expect{\mathrm{(I)} \mid \calF_t }
 & \le p \norm{\theta_t-\theta^*}
\kappa \eta_t^2 L \binom{s}{2}
\left( B \norm{\nabla F(\theta_t)} + G +\sigma \right) \nonumber \\
& \le p \norm{\theta_t-\theta^*}
\kappa \eta_t^2 L \binom{s}{2}
\left( B L \norm{\theta_t-\theta^*} + G +\sigma \right),
\label{eq:term_I_convex}
\end{align}
where the last inequality follows from $L$-smoothness so that
$$
\norm{\nabla F(\theta_t) } \le L \norm{\theta_t-\theta^*}
$$

For term $\rm (II)$, 
\begin{align*}
\mathrm{(II)} &= \iprod{\theta_t-\theta^*}{\sum_{i\in \tilde \calS_t}  w_i\nabla \ell_{i,t}(\theta_t)}
- \iprod{\theta_t-\theta^*}{\sum_{i\in\tilde \calS_t\setminus \calS_t}  w_i\nabla \ell_{i,t}(\theta_t)} \\
& \ge \iprod{\theta_t-\theta^*}{\sum_{i\in \tilde \calS_t}  w_i\nabla \ell_{i,t}(\theta_t)}
- \norm{\theta_t-\theta^* }
\norm{\sum_{i\in \tilde \calS_t \setminus \calS_t}  w_i\nabla \ell_{i,t}(\theta_t)}
\end{align*}
% Applying some concentration inequalities to show
% \[
% \norm{\sum_{i\in\tilde\calS_t}  w_i\nabla \ell_{i,t}(\theta) - p \nabla F(\theta)}
% \le C_1 \norm{\nabla F(\theta)} + C_2.
% \]
% \nb{missing step; add uniform sampling probability $p$ to setup. ref: FedProx (?) Should we renormalized the gradients from participating clients by $1/p$?}
Then, applying~\eqref{eq:tildeS-mean} and~\prettyref{lmm:adversarial-attack} yields
\begin{align}
\expect{\mathrm{(II)} \mid \calF_t }
% & \ge  p \norm{\nabla F(\theta_t) }^2
% - p \norm{\nabla F(\theta_t) }
% \sqrt{\epsilon}
% \left( \sum_{i \in [M]} w_i 
% \norm{\nabla F_i(\theta)}^2 + \sigma \right)\\
& \ge p \pth{ \iprod{\theta_t-\theta^*}{\nabla F(\theta_t)}
-\norm{\theta_t-\theta^*}
\sqrt{\epsilon}
\left( B \norm{\nabla F(\theta_t) } + G + \sigma \right)
}.
\end{align}
%\nbr{

By $\mu$-strong convexity, we have
$$
\iprod{\theta_t-\theta^*}{\nabla F(\theta_t)}\ge \mu \norm{\theta_t -\theta^* }^2.
$$
By $L$-smoothness, we have
$$
\norm{\nabla F(\theta_t) } \le L \norm{\theta_t-\theta^*}
$$
% \nbr{If we assume only convexity, we can only get 
% $$
% \iprod{\theta_t-\theta^*}{\nabla F(\theta_t)}\ge \frac{1}{L}
% \norm{\nabla F(\theta_t) }^2
% $$
% or 
% $$
% \iprod{\theta_t-\theta^*}{\nabla F(\theta_t)} \ge F(\theta_t)-F(\theta^*).
% $$
% However, $\norm{\theta_t-\theta^*}$ can still be very large even if
% $\nabla F(\theta_t)$ or 
% $F(\theta_t)-F(\theta^*)$ is very small. For instance, near the optimum, $F$ is very flat. 
% In view of above, let us proceed by assuming the $\mu$-strong convexity. }
It follows that 
\begin{align}
\expect{\mathrm{(II)} \mid \calF_t }
% & \ge  p \norm{\nabla F(\theta_t) }^2
% - p \norm{\nabla F(\theta_t) }
% \sqrt{\epsilon}
% \left( \sum_{i \in [M]} w_i 
% \norm{\nabla F_i(\theta)}^2 + \sigma \right)\\
& \ge p \pth{ \mu \norm{\theta_t-\theta^*}^2
-\norm{\theta_t-\theta^*}
\sqrt{\epsilon}
\left( BL \norm{\theta_t-\theta^*} + G + \sigma \right)
}.
\end{align}

Combining the last displayed equation with~\prettyref{eq:term_I_convex} yields that
\begin{align}
& \expect{ \iprod{\theta_{t+1}-\theta_t}{\theta_{t}-\theta^\ast} \mid \calF_t} \nonumber \\
& \le 
-  \beta p s \eta_t \mu \norm{\theta_t - \theta^* }^2
+  \beta p s \eta_t \norm{\theta_t - \theta^* }
\sqrt{\epsilon}
\left( BL \norm{\theta_t - \theta^* } + G + \sigma \right) \nonumber\\
& \qquad 
+ \beta p \norm{\theta_t - \theta^*}
\kappa \eta_t^2 L \binom{s}{2}
\left( BL \norm{\theta_t - \theta^*} + G +\sigma \right). 
\end{align}

For the quadratic term, analogous to  \prettyref{eq:high-order}, we have
\begin{align*}
\expect{\norm{\theta_{t+1}-\theta_t}^2 \mid \calF_t}
&\le \beta^2 s^2\eta_t^2 \left( 1+ \kappa \eta_t L (s-1)/2 \right)^2
p \left( B^2 \norm{\nabla F(\theta)}^2 + G^2 + \sigma^2 \right) \\
& \le \beta^2 s^2\eta_t^2 \left( 1+ \kappa \eta_t L (s-1)/2 \right)^2
p \left( B^2 L^2 \norm{\theta_t-\theta^*}^2 + G^2 + \sigma^2 \right).
\end{align*}

Combining the last two displayed equations and plugging them into~\prettyref{eq:Delta_evol} yields that 
\begin{align*}
   \expect{ \Delta_{t+1}^2 \mid \calF_t} & \le \Delta_t^2  -  2\beta p s \eta_t \mu \Delta_t^2
+   2 \beta p s \eta_t \Delta_t
\sqrt{\epsilon}
\left( BL \Delta_t + G + \sigma \right) \nonumber\\
& \qquad 
+  2 \beta p \Delta_t
\kappa \eta_t^2 L \binom{s}{2}
\left( BL \Delta_t + G +\sigma \right)\\
& \qquad + \beta^2 s^2\eta_t^2 \left( 1+ \kappa \eta_t L (s-1)/2 \right)^2
p \left( B^2 L^2 \Delta_t^2 + G^2 + \sigma^2 \right).
\end{align*}
In particular,
$$
\expect{\Delta_{t+1}^2 \mid \calF_t} \le (1-a_t)\Delta_t^2 + b_t \Delta_t + c_t,
$$
where 
\begin{align*}
    a_t &= 2\beta ps \eta_t  \left(
    \mu - \sqrt{\epsilon} BL - \eta_t \kappa \frac{s-1}{2} B
    L^2
    - \frac{1}{2} \beta s \eta_t \left( 1+ \kappa \eta_t L (s-1)/2 \right)^2 B^2 L^2  \right) \\
    b_t &= 2\beta p s \eta_t \left( \sqrt{\epsilon} \pth{G+\sigma} + \frac{\kappa  (s-1) \eta_t L}{2}  \pth{ G+ \sigma}   \right), \\
c_t & = \beta^2 p s^2   \eta^2_t  \left( 1+ \frac{\kappa (s-1) \eta_t  L}{2}\right)^2 \pth{G^2 + \sigma^2}.
\end{align*}
Using the fact that 
$
\Delta_t \le \frac{a_t}{2b_t}\Delta_t^2 + \frac{b_t}{2a_t},
$
and taking expectation over $\calF_t$
 we get that 
 $$
\expect{\Delta_{t+1}^2} \le \left(1- \frac{a_t}{2} \right) \expect{\Delta_t^2} + \frac{b_t^2}{2a_t}+ c_t,
 $$
Unrolling the recursion, we deduce that
\begin{align*}
\expect{\Delta_t^2} &\le \prod_{\tau=0}^{t-1}\left( 1- \frac{a_\tau}{2}\right) \Delta_0
+\sum_{\tau=0}^{t-1}
\left(\frac{b_\tau^2}{2a_\tau}+ c_\tau\right)
\prod_{\tau'=\tau+1}^{t-1}
\left( 1- \frac{a_{\tau'}}{2}\right).
% & \le \exp\left( -\frac{1}{2} \sum_{\tau=0}^{t-1} a_\tau \right)\Delta_0 + 
% \sum_{\tau=0}^{t-1}
% \left(\frac{b_\tau^2}{2a_\tau}+ c_\tau\right) \exp\left(
% -\frac{1}{2} \sum_{\tau'=\tau+1}^{t-1} a_{\tau'} \right)
\end{align*}
Choose $\eta_t= \frac{\theta}{t+\gamma}$ for some constants $\theta$
 and $\gamma$ such that 
 $\beta ps \mu \theta \ge 2$ and $\theta/\gamma \le \mu/(20\beta s L^2 B^2)$.
% where 
% $$
% \theta = \frac{3}{\beta ps \mu}, \quad  \gamma = \frac{20\beta s L^2 B^2 \theta}{\mu}.
% $$
Then
$\eta_t  \le \frac{\theta}{\gamma} \le \frac{\mu}{20\beta s L^2 B^2}$.
Therefore, $\eta_t \beta s L^2 B^2 \le 0.1 \mu$ and $\eta_t sL \le 0.1$. Furthermore, 
$\kappa \le \frac{e^{0.1}-1-0.1}{(0.1)^2/2}$. By 
the standing assumption, $\sqrt{\epsilon}B \le 0.1 \mu/L$ and hence
$
a_t \ge 2 C_0 \beta ps \eta_t \mu.
$
for some constant $C_0 \ge 1-0.1-0.05\kappa -0.05(1+0.05\kappa)^2 \ge  0.7929.$
Let $\rho =C_0\beta ps\theta \mu $. By our choice of $\theta$, $\rho>1.5.$ Then it follows that 
$$
\prod_{\tau'=\tau}^{t-1}\left( 1- \frac{a_{\tau'}}{2}\right)
\le \exp\left( -\frac{1}{2} \sum_{\tau'=\tau}^{t-1} a_{\tau'} \right)
\le \exp\left( -\rho \sum_{\tau'=\tau}^{t-1}
\frac{1}{\tau'+\gamma}
\right)
\le \exp\left( -\rho \log \left( \frac{t+\gamma}{\tau+\gamma}\right)
\right),
$$
where the last inequality holds because
$\sum_{\tau'=\tau}^{t-1}
\frac{1}{\tau'+\gamma} \ge \int_{\tau+\gamma}^{t+\gamma} \frac{1}{x} \diff x=\log \frac{t+\gamma}{\tau+\gamma}.$
% Similarly,
% $$
% \prod_{\tau'=\tau}^{t-1}
% \left( 1- \frac{a_{\tau'}}{2}\right)
% \le \exp\left( -\rho \log \left( \frac{t+\gamma}{\tau+\gamma}\right) \right).
% $$
% \nbr{some $t+\gamma+1$ issue.}
Moreover,  $a_t/2 \le  \beta p s \eta_t \mu \le \frac{\mu^2 p}{20 L^2 B^2} \le 0.05$, so that $1-a_t/2 \ge 0.95$. Hence,
\begin{align*}
\frac{1}{1-a_t/2}
\left(\frac{b_t^2}{2a_t} + c_t\right)
& \le \frac{\beta (2ps\eta_t)^2 2(\epsilon + (C_1 s L\eta_t)^2 )  }{ 0.95 \cdot 4C_0 ps\eta_t \mu}\pth{ G+ \sigma}^2 + C_2 \beta^2 ps^2 \eta_t^2 \pth{ G+ \sigma}^2
%\beta^2 (ps\eta_t) (C_2 s L\eta_t) \pth{ G+ \sigma}^2
\\
& \le \frac{2}{0.95 C_0} \epsilon \pth{ G+ \sigma}^2 \beta ps\eta_t/\mu
 + C'\beta^2 p  s^2  \eta_t^2 \pth{ G+ \sigma}^2,
\end{align*}
where $C_1, C_2, C'$ are some universal constants and the last inequality uses the fact that $\eta_t \beta s L^2 \le 0.1 \mu$.
Therefore,
\begin{align*}
\left(\frac{b_\tau^2}{2a_\tau}+ c_\tau\right)
\prod_{\tau'=\tau+1}^{t-1}
\left( 1- \frac{a_{\tau'}}{2}\right)
\le
\pth{ G+ \sigma}^2  
 \left(\frac{2\epsilon \beta ps\theta}{0.95 C_0 \mu(\tau+\gamma)}
 + \frac{C'  \beta^2 p  s^2  \theta^2}{(\tau+\gamma)^2} \right)
 \left(\frac{t+\gamma}{\tau+\gamma}\right)^{-\rho}.
\end{align*}
It follows that 
\begin{align*}
& \sum_{\tau=0}^{t-1} \left(\frac{b_\tau^2}{2a_\tau}+ c_\tau\right)
\prod_{\tau'=\tau+1}^{t-1}
\left( 1- \frac{a_{\tau'}}{2}\right)\\
& \le 
\frac{2}{0.95 C_0} \epsilon \pth{ G+ \sigma}^2 \beta ps\theta \mu^{-1} (t+\gamma)^{-\rho}
\sum_{\tau=0}^{t-1} (\tau+\gamma)^{\rho-1} \\
& \qquad + C' \pth{ G+ \sigma}^2 \beta^2 ps^2  \theta^2 (t+\gamma)^{-\rho}
\sum_{\tau=0}^{t-1} (\tau+\gamma)^{\rho-2} \\
& \le \frac{2}{0.95 C_0} \epsilon \pth{ G+ \sigma}^2 \beta ps\theta/(\rho \mu) +
C' \pth{ G+ \sigma}^2 \frac{\beta^2 ps^2  \theta^2 }{\rho-1}
\frac{1}{t+\gamma}.
\end{align*} 
In conclusion, we get that 
$$
\expect{\Delta_t^2}
\le \left( 1+ t/\gamma\right)^{-\rho} \Delta_0 + \frac{2\epsilon}{0.95 C_0^2\mu^2} \pth{ G+ \sigma}^2+C'  \pth{ G+ \sigma}^2  \frac{\rho^2 }{(\rho-1)p\mu^2} 
\frac{1}{t+\gamma},
$$
where $2/(0.95 C_0^2) \le 2/(0.95 \times 0.7929^2) < 4. $

The proof for FedProx is completely analogous. 
The only difference is that we need to apply~\prettyref{lmm:fedprox-diff} instead of~\prettyref{lmm:s-steps-diff} to bound the perturbation of the local proximal step. In the end, we get that 
$$
\expect{\Delta_{t+1}^2 \mid \calF_t} \le (1-a'_t)\Delta_t^2 + b'_t \Delta_t + c'_t,
$$
where
\begin{align*}
a_t' & = 2 \beta p \eta_t \left( \mu - \sqrt{\epsilon}BL- \frac{\eta_t L }{1-\eta_t L_-} BL   - \beta \frac{\eta_t }{2} 
\left( 1+ \frac{\eta_t L}{1-\eta_t L_-}\right)^2 B^2L^2 \right), \\
b_t' & =2  \beta p \eta_t \left( \sqrt{\epsilon} \pth{G+\sigma} + \frac{\eta_t L}{1-\eta_t L_-}  \pth{ G+ \sigma}   \right), \\
c_t' & = \beta^2 p \eta_t^2 \left( 1+ \frac{\eta_t L}{1-\eta_t L_-}\right)^2 \pth{G^2 + \sigma^2}.
\end{align*}
Following the same steps as above, we can further deduce that 
$$
\expect{\Delta_t^2}
\le \left( 1+ t/\gamma\right)^{-\rho} \Delta_0 + \frac{2\epsilon}{0.95 C_0^2\mu^2} \pth{ G+ \sigma}^2+C' \pth{ G+ \sigma}^2\frac{\rho^2 }{(\rho-1) p \mu^2}
\frac{1}{t+\gamma},
$$
where $\rho =C_0\beta p\theta \mu>1.1 $ and $C_0 \ge 0.727$ so that
$2/(0.95C_0^2) < 4.$
\end{proof}

\section{Experiments}
\label{app: exp}
In this section, we provide the detailed setups and additional experiments.
% \footnote{Our code can be accessed at 
% \url{https://anonymous.4open.science/r/FL_AdvDropout}.
% Alternative #2:
% Our code will be made public upon publication.
% }.
% \nbr{Just want to make sure providing the Github link does not violate the anonymization requirement. Please double-check the submission guidelines if you're not sure.}

\paragraph*{\bf Hardware environments:}
we run all the experiments on a cluster with 8 Tesla V100 Volta GPUs and 80 Intel Xeon E5 CPUs. The codes are built upon PyTorch 1.13.1 \cite{paszke2019pytorch}.

\paragraph*{\bf Implementation details:}
For the numerical calculation of GM, we adopt the smoothed Weiszfeld algorithm from \cite{9721118}.
To adapt to clients with quantify skewness, the iterative centered clipping rule is modified to
\begin{align}
\bm{v}_{l+1} = \bm{v}_l + \sum_{i=1}^M w_i \pth{\theta_i - \bm{v}_l} \min \pth{1,\frac{\tau_l}{\norm{\theta_i - \bm{v}_l}}},
\label{eq: imbalanced cclip}
\end{align}
where the choice of $\tau_l = \frac{10}{1-\beta_0}$ follows from \cite{karimireddybyzantine22}, and
$\beta_0$ is the momentum coefficient.
Likewise,
Line 9 in Algorithm 1 of \cite{gu2021fast} is modified to
\begin{align}
\theta_{t+1} = \theta_t - \eta_t \sum_{i=1}^M w_i G^i.
\label{eq: mifa updates}
\end{align}
%\nbr{Also need to polish the plots, e.g., font sizes}

\subsection{CIFAR-10}
\begin{figure}[b]
\includegraphics[width=\textwidth]{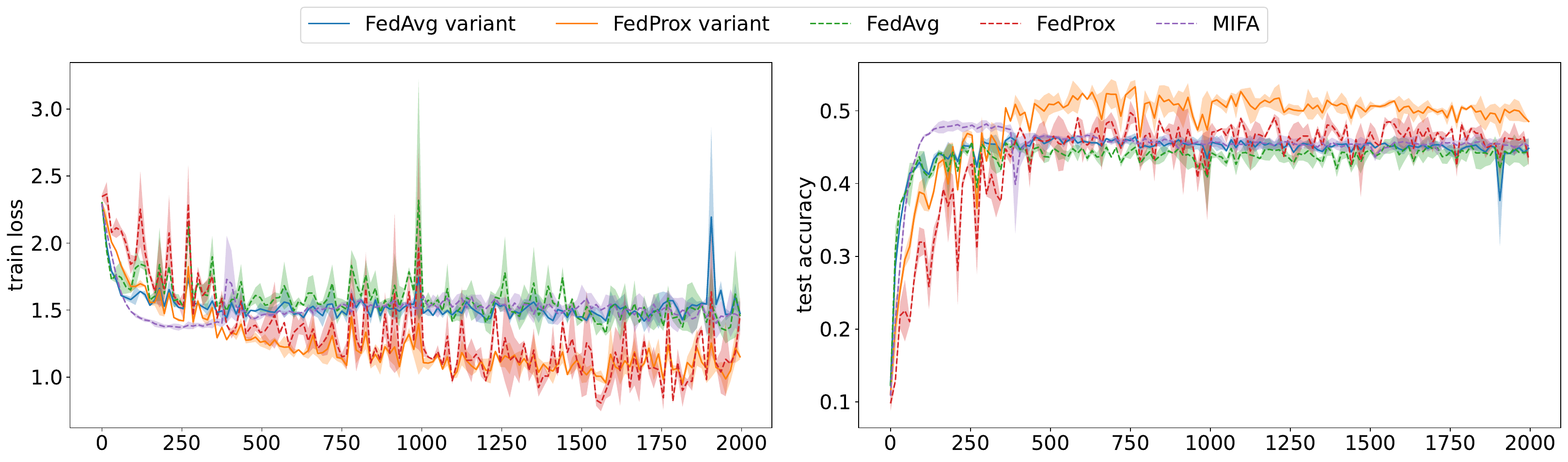}
\caption{CIFAR-10 results with Dirichlet parameter $\alpha=0.1$ and dropout fraction $\epsilon =0.2$ baseline comparisons on adversarial client unavailability scheme in Section \ref{subsec: adversarial client unavailability}.}
\label{fig:cifar10_0.1_0.3_baseline}
\end{figure}

\begin{figure}[b]
\centering
    \includegraphics[width=\textwidth]{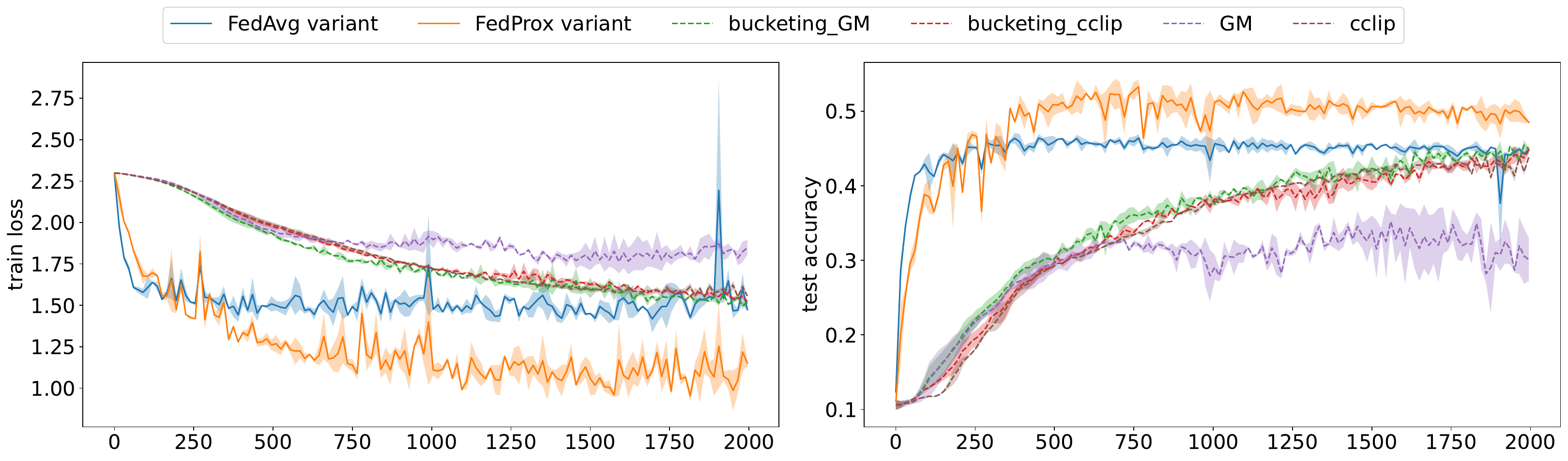}
\caption{CIFAR-10 results with Dirichlet parameter $\alpha=0.1$ and dropout fraction $\epsilon =0.2$ Byzantine comparisons on adversarial client unavailability scheme in Section \ref{subsec: adversarial client unavailability}.}
\label{fig:cifar10_0.1_0.3_byzantine}
\end{figure}

We present the additional experiments on CIFAR-10 dataset under light unavailability scheme with $\epsilon=0.2$. 
The results are shown in Fig.\,\ref{fig:cifar10_0.1_0.3_baseline} and \ref{fig:cifar10_0.1_0.3_byzantine}.
Generally, the observations resemble those from the case of $\epsilon = 0.8.$
Our variants' trajectories are less fluctuating than the naive FedAvg and FedProx algorithms.
In the baseline comparisons, all the FedAvg-type algorithms achieve comparable performance,
while our FedProx variant achieves the best test accuracy.
%Meanwhile, 
The Byzantine-resilient algorithms, except the naive GM, reach similar results as well, as a total of eight clients participate in the training per round.
This is in sharp contrast to the highly adversarial case of $\epsilon = 0.8,$ where only two clients are responsive in each round.

\subsubsection{Prolonged training from 2000 rounds to 4000 rounds of CIFAR-10.}

Since the bucketing-GM, bucketing-cclip, GM, and cclip do not appear to have converged in Fig.\,\ref{fig:cifar10_0.1_0.3_byzantine}, we further increase the training horizon from 2000 communication rounds to 4000 rounds in Fig.\ref{fig: prolong 1}. There are indeed improvements in training for bucketing-cclip and bucketing-GM with respect to train loss and test accuracy; however, the FedProx variant still achieves the best performance. The final average loss and accuracy of our variants are plotted as horizontal lines for a neat presentation. It is worth noting that the Byzantine algorithms we adopt in the numerical evaluations are not the original algorithms proposed by the authors since the original ones do not apply to our system setup. 
% Intuitively, they are the enhanced algorithms with memory of momentum information from the past as compared with the original ones. 
% {\red 
% Thus, the analysis of the Byzantine benchmarks in their corresponding papers is not necessarily applicable here.
% }
% In addition, they are not equipped with local computations, and the convergence is quite slow.

\subsection{Synthetic datasets}
\begin{figure}[htb]
\centering
\includegraphics[width=\textwidth]{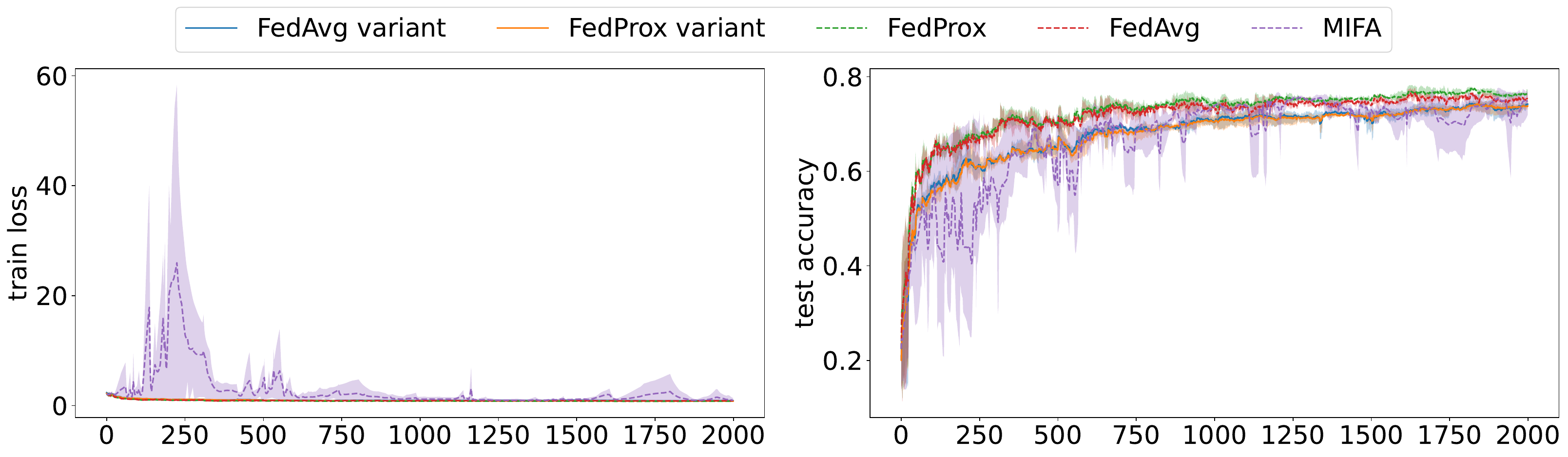}
\includegraphics[width=\textwidth]{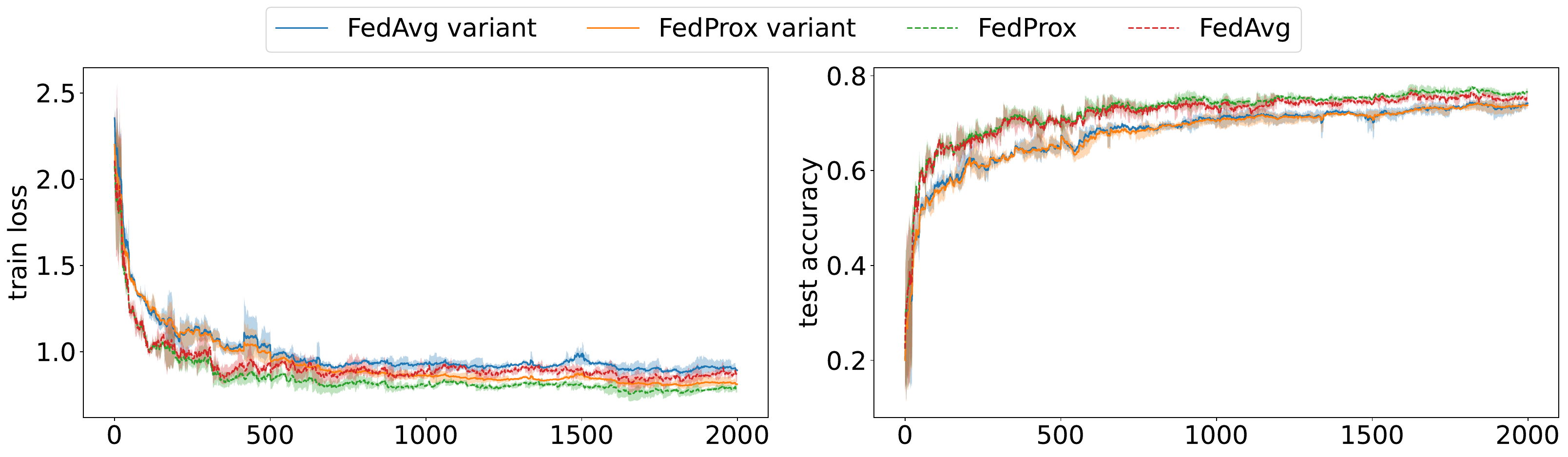}
\includegraphics[width=\textwidth,trim={0 0 0 1cm},clip]{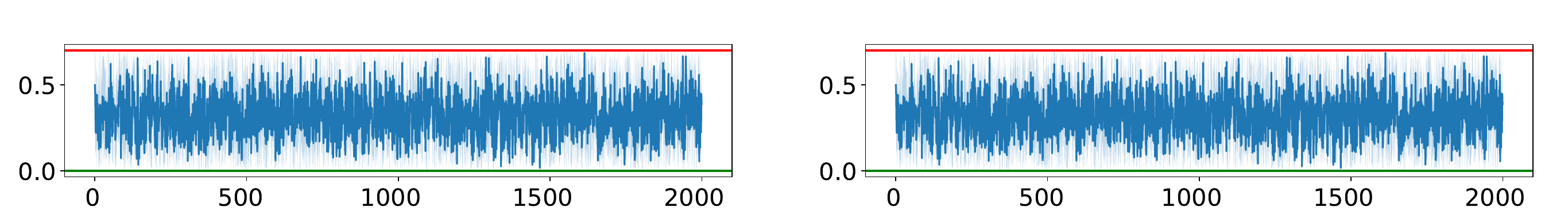}
% %
{
\begin{minipage}{0.49\textwidth}
\centering
\quad \small FedAvg variant dropout fraction $\epsilon_t \in [0,0.70)$.
\end{minipage}
\begin{minipage}{0.49\textwidth}
\centering
\qquad \small FedProx variant dropout fraction $\epsilon_t \in [0,0.70)$.
\end{minipage}
}
% %
\vskip -1pt
\caption{\ Synthetic datasets: baseline comparisons with dropout fraction $\epsilon =0.7$
on adversarial client unavailability scheme in Section \ref{subsec: adversarial client unavailability}.
}
\label{fig:synthetic_11_0.7_baseline}
\end{figure}

\begin{figure}[htb]
\centering
\includegraphics[width=\textwidth]{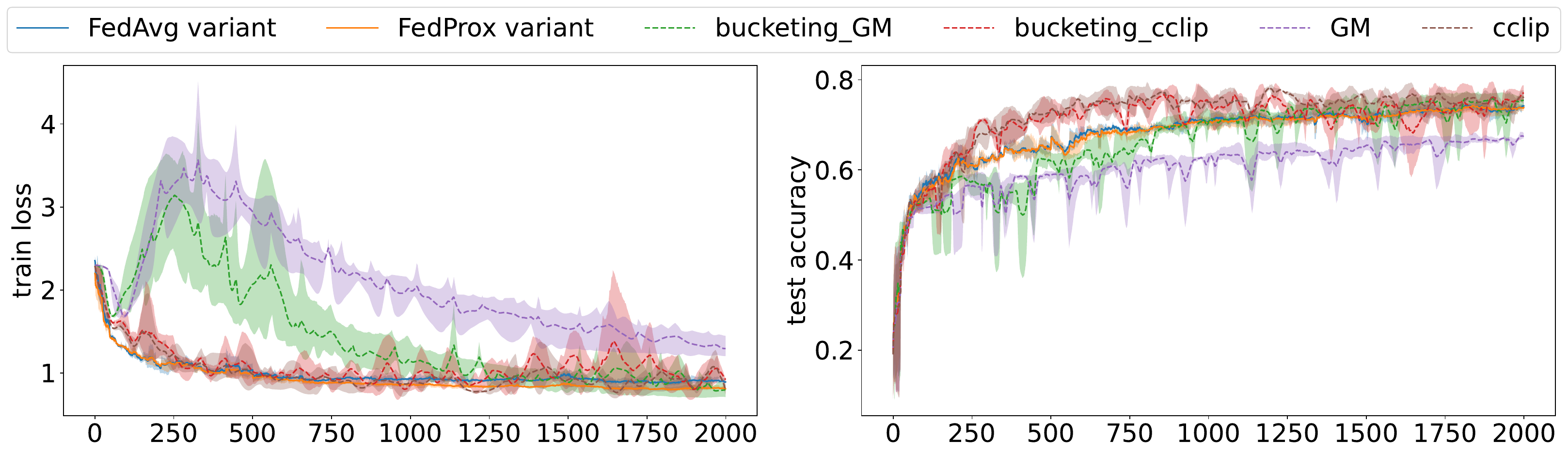}
\caption{Synthetic datasets: Byzantine comparisons with dropout fraction $\epsilon =0.7$
on adversarial client unavailability scheme in Section \ref{subsec: adversarial client unavailability}.
}
\label{fig:synthetic_11_0.7_byzantine}
\end{figure}

We present the additional experiments on the synthetic datasets with $\epsilon=0.7$. 
The results are shown in Fig.~\ref{fig:synthetic_11_0.7_baseline} and~\ref{fig:synthetic_11_0.7_byzantine}.
Again, the trends are similar to~\prettyref{sec:synthetic}. 
In all the experiments, the proposed variants progress smoothly and achieve similar or better performances compared with other baselines.
More importantly, the performance does not demand additional memory, unlike MIFA or the Byzantine-resilient algorithms.


\subsubsection{Prolonged training from 2000 rounds to 4000 rounds for synthetic datasets.}

\begin{figure}[!htb]
    \centering
    \begin{subfigure}[b]{0.46\textwidth}
    \includegraphics[width=\textwidth]{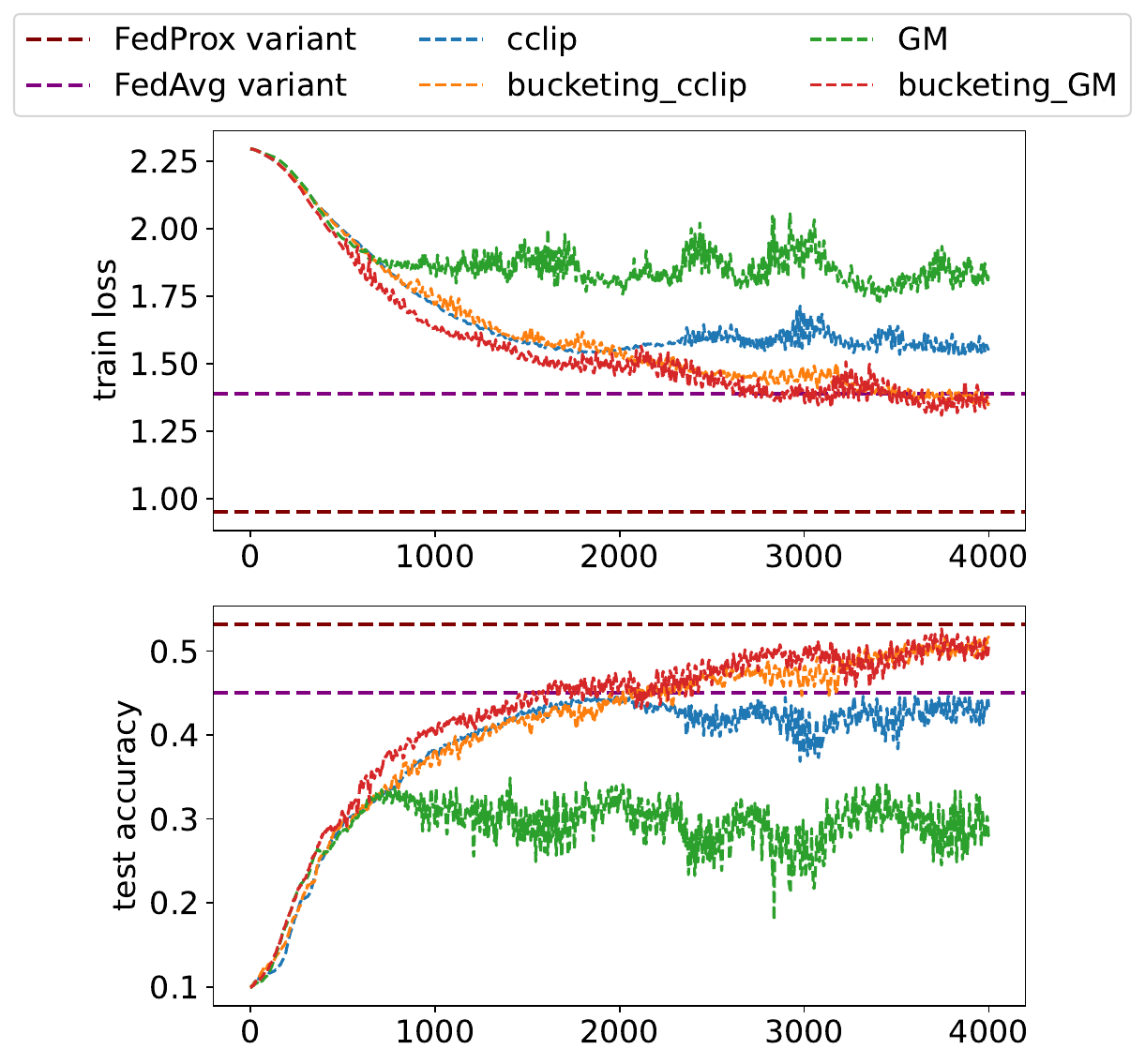}
    \caption{The results of Fig.~\ref{fig:cifar10_0.1_0.3_byzantine} after prolonged training.}
    \label{fig: prolong 1}
    \end{subfigure}
    \begin{subfigure}[b]{0.48\textwidth}
    \includegraphics[width=\textwidth]{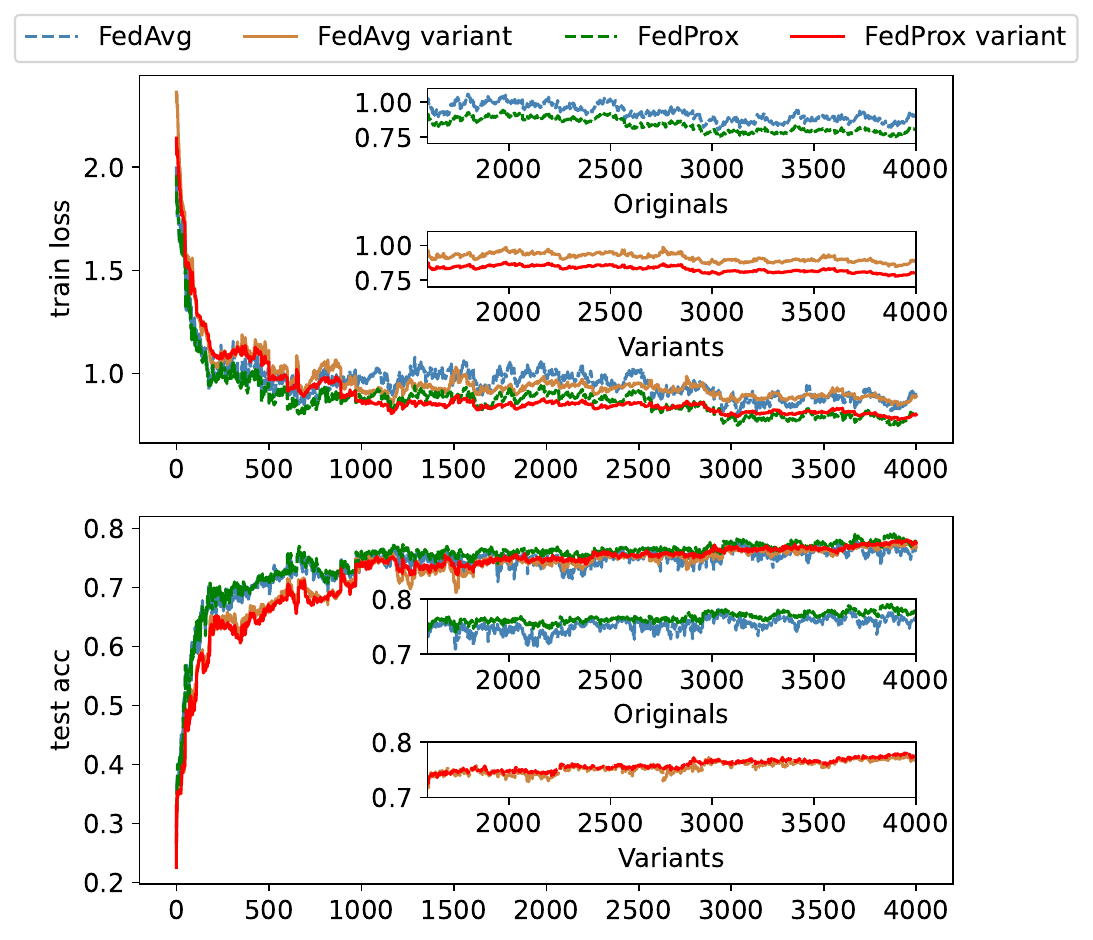}
    \caption{The results of Fig.~\ref{fig:cifar10_0.1_0.3_byzantine} after prolonged training.}
     \label{fig: prolong 2}
    \end{subfigure}
    % \vskip -1pt
     \vskip -0.45\baselineskip 
    \caption{The plots of Fig.~\ref{fig:cifar10_0.1_0.3_byzantine} after extending the training time from 2000 rounds to 4000 rounds.}
\end{figure}

We also increased the training horizon from 2000 communication rounds to 4000 rounds on Synthetic (1,1) data in Fig.\,\ref{fig: prolong 2}. 
All the algorithms achieve comparable test accuracy, while the algorithms of the same types (FedAvg-type and FedProx-type) reach alike train losses. We group the numerical results of the original algorithms (FedAvg and FedProx) and the proposed variants into two zoom-in boxes. The scrutiny reveals that the trajectories of the proposed variants are more smooth when compared with the original ones. This matches our anticipation. It is worth noting that Eq.\,\eqref{eq:global_update} of our submission is equivalent to 
$\theta_{t+1} = (1- \beta \sum_{i\in {\calS}_t} w_i)\theta_t + \beta \sum_{i\in {\calS}_t} w_i \theta_{i,t+1}$. 
By choosing $\beta$ so that $\beta \sum_{i \in {\calS}_t} w_i <1$, we are smoothing the trajectory of $\theta_t$.

\subsection{Additional attacks: Round-robin adversary and random responsive probability}
\label{app: additional adversary}
\begin{figure}[!htb]
\centering
\includegraphics[width=0.9\textwidth]{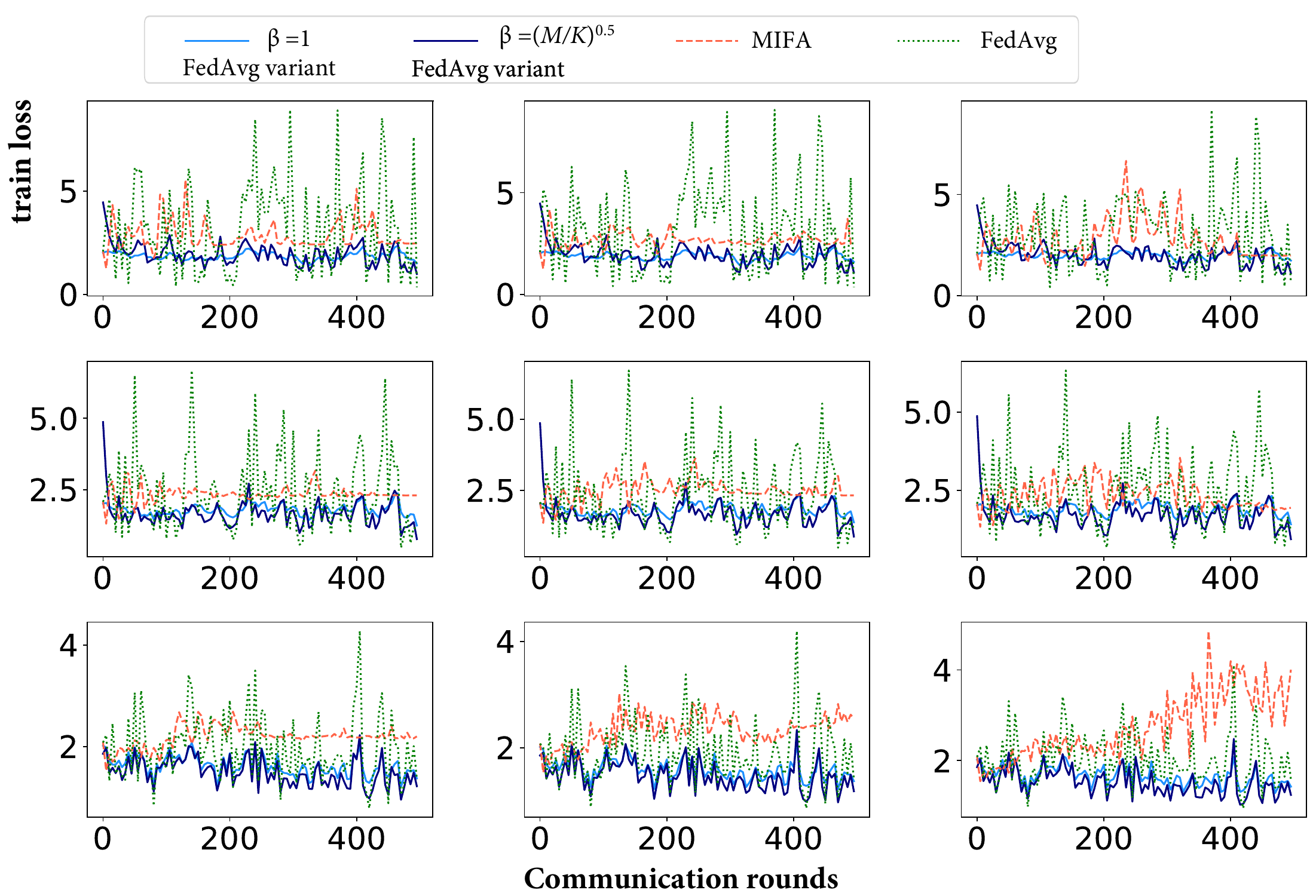}
\caption{Train loss of round-robin adversaries on {\em CIFAR-10} dataset.} \label{fig: round-robin baseline}
\end{figure}

\begin{figure}[!htb]
\centering
\includegraphics[width=0.9\textwidth]{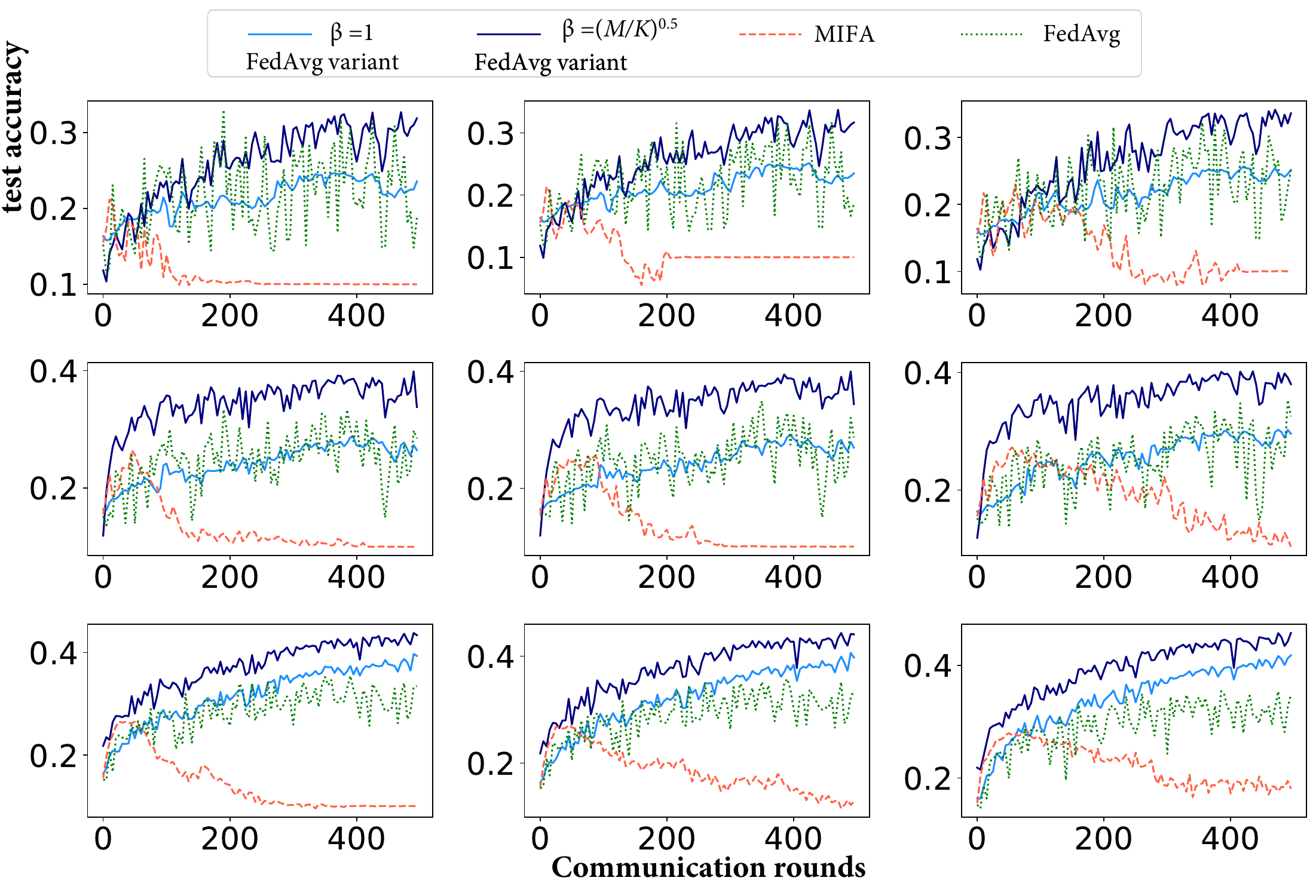}
\caption{Test accuracy on {\em CIFAR-10} dataset.}
\label{fig: round-robin baseline test}
\end{figure}

%We thank the reviewer z1S5's proposition to try our algorithm in the random unavailability settings. 
We also test the variants of FedAvg and FedProx against another dropout scheme, which we refer to as {\it round-robin attack.} Our results can be found in Fig.\,\ref{fig: round-robin baseline} and Fig.\,\ref{fig: round-robin baseline test}. The adversary randomly partitions the clients into $r$ equal-sized groups and assigns each group with a label in $\{0,\dots,r-1\}$. During $t$-th communication round for $100(i-1)<t\le 100i$, all clients from the group with the label $i~(\mathrm{mod}~r)$ will be unresponsive, i.e., the groups become unavailable in a round-robin fashion every 100 rounds. In every round, there are $(1/r)$-fraction of unresponsive clients on average. 

%This scheme captures a scenario where a client that is unavailable at time $t$ is likely to stay inactive at time $t+1$, whereas it will come back in the long run.
%For the system setup, 

We considered the sort-and-partition scheme to generate non-IID datasets: 
A total of 100 clients each holding image samples of 2 classes from {\it CIFAR-10} dataset. This follows \cite{li2020federated} and yields highly non-IID data. Specifically,  as we have 10 data classes in {\it CIFAR-10} datasets, we partition the 100 clients into 5 groups with equal sizes. We assign each group an $\text{ID}\in[5]$. The clients in the group $\text{ID}$ will be assigned the images from the classes $\sth{2\text{ID}-1,2\text{ID}}$. In each communication round, a client draws a batch of 10 samples from the local dataset.

In each communication round, $|\tilde{\calS_t}|=K\in\{5,10,30\}$ clients are randomly sampled. Clients will be dropped by the adversary according to the schemes we discussed. In all figures, the plots from top to bottom are with $K= 5, 10, 30$, respectively. The plots from left to right are of $r=4, 5, 10$, respectively.

It is observed from Fig. 4 that the FedAvg variant with $\beta = \sqrt{M/K}$ beats all the other algorithms. The increase in $K$ helps to smooth out the curves. Empirically, the increase in $\beta$ from 1 to $\sqrt{M/K}$ introduces acceleration in training.

\end{document}